\documentclass[12pt]{book}
\usepackage{geometry}
\usepackage{amsthm}
\usepackage{xparse,hyperref,bookmark}
\usepackage{mathrsfs,amsmath,amsfonts,amssymb}
\usepackage{amssymb}
\usepackage{amsmath}
\usepackage{aliascnt}
\usepackage{appendix}
\usepackage{graphicx}
\usepackage[english]{babel}
\usepackage[framemethod=TikZ]{mdframed}
\usepackage[utf8]{inputenc}
\usepackage[T1]{fontenc}
\usepackage{mathtools}
\usepackage[style=alphabetic,natbib=true, maxbibnames=99]{biblatex}
\usepackage{color}
\usepackage{bbm}
\usepackage{subcaption}
\usepackage{booktabs}

\newcommand\ID{\text{ID}}
\newcommand\EE{\mathbb{E}}
\newcommand\NN{\mathbb{N}}
\newcommand\PP{\mathbb{P}}

\newcommand\RR{\mathbb{R}}

\newcommand\SSS{\mathbb{S}}
\newcommand\matS{\mathcal{S}}
\newcommand\matC{\mathcal{C}}
\newcommand\PA{\mathrm{Pa}}
\newcommand\Ind{\mathbbm{1}}
\renewcommand{\d}[1]{\ensuremath{\operatorname{d}\!{#1}}}

\newcommand{\matr}[1]{\mathbf{#1}}

\DeclareMathOperator*{\argmin}{arg\,min}
\DeclareMathOperator*{\argmax}{arg\,max}
\newcommand{\pc}{\%}
\DeclareMathOperator{\codim} {codim}
\newcommand\Supp{\text{Supp}}

\newcommand{\test}{\text{test}}
\newcommand{\train}{\text{train}}
\newcommand{\Ptest}{\PP^{\test}}
\newcommand{\Ptrain}{\PP^{\train}}
\newcommand{\envs}{\mathcal{E}}
\newcommand{\Etrain}{\mathcal{E}_{\train}}
\newcommand{\Etr}{\Etrain}
\newcommand{\Eall}{\mathcal{E}_{\text{all}}}
\newcommand{\OOD}{\text{OOD}}
\newcommand{\hH}{\hat H}
\newcommand{\hh}{{\hat h}}

\newcommand{\matY}{\mathcal{Y}}
\newcommand{\matH}{\mathcal{H}}
\newcommand{\matX}{\mathcal{X}}
\newcommand{\matA}{\mathcal{A}}
\newcommand{\fancyp}{\mathscr{P}}
\newcommand{\hmatH}{\hat{\matH}}
\newcommand{\hmatY}{\hat{\matY}}
\newcommand{\Tmax}{{H}}

\theoremstyle{plain}
  \newtheorem{theo}{Theorem}[chapter]

  \newaliascnt{coro}{theo}
  
  \aliascntresetthe{coro}

  \newaliascnt{prop}{theo}
  \newtheorem{prop}[prop]{Proposition}
  \aliascntresetthe{prop}

  \newaliascnt{lemma}{theo}
  
  \aliascntresetthe{lemma}

  \newaliascnt{obs}{theo}
  
  \aliascntresetthe{obs}

  \newaliascnt{obsdef}{theo}
  \newtheorem{obsdef}[obsdef]{Observation / Definition}
  \aliascntresetthe{obsdef}

\theoremstyle{definition}
  \newtheorem{example}{Example}[chapter]

  \newaliascnt{defi}{theo}
  \newtheorem{defi}[defi]{Definition}
  \aliascntresetthe{defi}

  \newaliascnt{assu}{theo}
  \newtheorem{assu}[assu]{Assumption}
  \aliascntresetthe{assu}

\usepackage{fancyhdr}
\newenvironment{abstract}%
{\cleardoublepage \vfill\begin{center}%
\bfseries Acknowledgements \end{center}}%
  {\vfill\null}

\author{Martin Arjovsky}

\graphicspath{{./figures/}}

\begin{document}
\thispagestyle{empty}

\begin {center}

\includegraphics[scale=2]{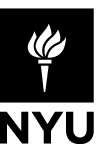}

\medskip
\textbf{NEW YORK UNIVERSITY}

\smallskip

\textbf{Courant Institute of Mathematical Sciences}

\vspace{3.5cm}

\textbf{\large PhD Thesis}

\vspace{1.5cm}

\textbf{\large Out of Distribution Generalization in Machine Learning}

\vspace{1.5cm}

\textbf{Martin Arjovsky}

\end {center}

\vspace{1.5cm}

\noindent \textbf{Advisor: L\'eon Bottou} \\

\noindent \textbf{Thesis Committee:
  Yoshua Bengio,
  Charles Blundell,
  L\'eon Bottou,
  Joan Bruna,
  Kyunghyun Cho,
  Christina Heinze-Deml,
  Yann LeCun.
  } \\

\vspace{3cm}

\rightline{Defense Date: 17 December 2019}

\clearpage
\begin{flushright}
\null \vspace {\stretch {1}}
  A mi pap\'a Jorge.
\vspace {\stretch {2}}\null
\end{flushright}

\begin{abstract}
These acknowledgements are surely incomplete and I apologize in advance to all those that have been ignored by these brief words.
An uncountable number of people helped me through this stage and I could have never done any of this alone.

To my father, to whom this thesis is dedicated.
From an early age you inspired in me the desire to solve problems, to learn, and to push the boundaries.

To my mother and my brother, for the unconditional support, the love, and the eternal companionship.

To L\'eon Bottou. Being advised by you is a privilige.
Working with you is a constant learning experience.
Your vision for what is and what will be important, enveloped with a deep curiosity and a genuine drive
  for understanding, helped me take shape as a researcher, and I hope to keep learning from you
  in the many years to come. None of this thesis would have existed without you.

To Yoshua Bengio. Few people have had a strong impact in my life as you did.
After sending you a short email with a half-baked idea, you decided to take me under your wing
  in Montr\'eal, and I can honestly say those months were some of the happiest I've ever been.
The constant feeling of learning every day, being surrounded by amazing people, and at the same
  time being humbled and inspired by your everyday knowledge made waking up every day in the freezing
  weather an experience of sheer joy. Finally, it is not just your role as an academic mentor,
  but your human character that make working with you so rewarding:
  your commitment to education, to your students, and
  to society are a constant source of inspiration.

To David Lopez-Paz. I consider you family. Thank you for so many happy moments, for the techno, for calling me out when I need it,
for teaching me about causality, for the infinite conversations, for everything. Saying all that I'm thankful for in our friendship would take
more space than this thesis does. If it wasn't for you, this thesis would also not have existed.
I am so looking forward to many more years of friendship.

To Pablo Sprechmann. When you're new to a city, finding someone that speaks not only the same language but the exact same accent, and from someone as welcoming as Pablo, makes you feel blessed. For the many many lunches, conversations, the clear advice, and your great company, thank you for making me feel at home.

To Cinjon Resnick. Thank you for your friendship and the many discussions on all aspects of life. Every time I talk to you I leave feeling motivated and ambitious. Thank you for so much love.

To Marco Vanotti. Thank you for so much laughter, for the interminable talks about life, work, what's right and what's wrong, and all the stupid corners of life. Thank you for allowing me to bounce all my stupid ideas with you. Thank you for so much joy.

To Ishmael Belghazi and Marcin Mocsulzki. For bouncing off crazy ideas, for complaining together about life, for sharing happiness and realizing how lucky we are to have this amazing life. For the balcony chats, for the many beers, the technical discussions, the wondering about the future, thank you for everything.

To Amar Shah. Thank you for making research fun. The many shared laughs, crashing into the corporate parties, thank you for introducing me into this research world with a smile.

To Arthur Szlam and Soumith Chintala. You guys basically taught me how to do research. From day one when I was interning as an unbearable undergrad, despite my resistance you tried to teach me how to make arguments bulletproof. What problems we should care about, and how to develop experiments to precisely prove our points. How state of the art is not what matters; that what does is identifying the core problem, validating it's importance, solving it, and testing the solution; with targeted experiments all the way through, not just at the end. To this day I don't know if I ever met a more brilliant person than Arthur, or a better engineer than Soumith, and I don't know if I ever will. A quote from Arthur I received on my first year as a PhD student and I will always remember is as follows: "Never make a claim that doesn't follow from an experiment or a theorem". The surprising importance of this phrase, given the amount of times we tend to break it in the false name of intuition,
  will stay with me for a very long time.

To Youngdock Choi. For the many life and math discussions. During my PhD you helped me keep the love for pure math alive. 
  I'm sure you are going to be a great mathematician, regardless of whatever endeavour you end up doing.

To Ishaan Gulrajani, for being one of the best researchers I've had the priviliged of working with. I look forward to seeing you become a star soon enough.

To all my coauthors, you know who you are, I couldn't have done any of this alone.

To my dear friends Mar\'ia Arone, Juan Luis Arag\'on, Agustín Valente, Facundo Orellana, Clementina Flores and Agostina Baiardino.

To Carlos Diuk, who taught me what Machine Learning is, and why it's awesome.

To Federico Felguer, who inspired my love for mathematics.

To Pablo Groisman and Juan Pablo Pinasco, who taught me most of the math I knew until my PhD. Perhaps more importantly, who made me understand that pure mathematics can be an incredibly powerful tool, that can alter the way we think about most objects in the applied world, and solve otherwise unreachable problems. You made me understand what research in math really looks like, and made it be a part of my life that I so desperately adore.

To Luis Scoccola and Nacho Darago. For so many inspiring conversations, for giving me a window into the insanely large side of math that I know nothing about.
For making university and learning math so much more inspiring and fun.

To Alex Lamb, for being the funniest guy in ML, a great friend, and a great researcher.

  To Federico Lebr\'on, for showing me that unreachable things can be reachable, when I was nothing but starting to study. Thanks for the far reaching support and advice.

  To Pablo Zadunaisky, for teaching us how to properly write math in the Differential Geometry class. This has been one of the most useful things I've learnt in the last 5 years.

To Anil Kokaram and Yao-Chung Lin, for giving me my first research project, helping me every step of the way, and teaching me how awesome applied research
can be.

To my thesis committee.
  To Kyunghyun Cho, for uncountable discussions and providing an unachievable example of what I want to become as a junior professor.
  To Christina Heinze-Deml, for the friendship, for not letting me get away with half-baked arguments, and for beautiful work that inspired a large part of this thesis.
  To Yann LeCun, for leading this amazing lab I'm proud to be part of, and for helping me whenever I needed.
  To Joan Bruna, for letting me bounce ideas with you whenever I wanted, and overall being a ridiculously nice person.
  To Charles Blundell, for making my stay at DeepMind so much fun, for all the great advice, for giving me freedom to do whatever research I wanted, and at the same time being incredibly helpful and making me feel part of your amazing research group.

  To all my amazing professors, both in Courant and in UBA. I consider myself extremely lucky to learn from such ridiculously enthusiastic and smart people. To the people that made all the online courses I took, without this I wouldn't have been able to learn machine learning.

  To the many people that helped me grow in this journey. To Christina Heinze-Deml, Alex Nowak, Alfredo Canziani, Cijo Jose, Manuel Gim\'enez, Avital Oliver, Mike Phillips, Axel Sirota, Tom Sercu, Pablo Zivic, Oded Regev, Caglar Gulcehre, Aaron Courville, Gabriel Synnaeve, Francesco Visin, Yarin Gal, Uri Shalit, Emily Denton, Ferenc Huszar, Jake Zhao, Aditya Rahmesh, Faruk Ahmed, C\'esar Laurent, Keith Ross, Jelena Luketina, Damian Furman, Ignacio Niesz, Mart\'in Mongi Badia, Juan Doldan, Maxi Bustos and all the rest of GARG, Andrés Campero, Jessica Forde, Facundo Sapienza, Amy Zhang, Agustín Somacal, Pedro Ortega, Felix Hill, Leopoldo Taravilse, Erin Grant, Ari Pakman, Stanisław Jastrzębski, Joseph Paul Cohen, Mathieu Germain, Federico Carnevale, Will Whitney, Maxime Oquab, Anna Klimovskaia, François Charton, Alex Pritzel, Aahlad Manas, Elman Mansimov, Zbigniew Wojna, and Matias Domingues.

To the Balinese, for welcoming me into their magical place.

  To the hundreds of coffee shops where I did so much studying and work. Here are a few of them for the interested reader: Caravane Café, La Bicicleta, Del Viento, Storyville Queen Anne, The Mill, City of Saints, Blue Bottle, SEY, Cafenation Amsterdam, Ona Café, Fortunata, Leia’s, Everyman Espresso. Augmenting this list is left for future work.

\end{abstract}

\tableofcontents

\begin{center}
  \clearpage
\null \vspace {\stretch {1}}
  \textit{``Pensar es olvidar diferencias, es generalizar, abstraer.''} \\Jorge Luis Borges, \textit{Funes el memorioso}
  \vspace {\stretch {2}}\null
\end{center}


\chapter{Introduction}
Why are we here? Most of us are here for a simple reason: we want to solve AI.
Then what does artificial intelligence have to do with the title of this book?
One way AI is modernly defined is as the study and design of \emph{rational agents} \cite{russell-ai-2009}.
In this sense, we describe a system as intelligent when it acts maximizing some expected notion of performance.
The subfield of machine learning deals with the subset of problems and algorithms where experience is available to the agent (usually in some form of data) that can be leveraged
to improve this notion of performance \cite{mohri-fml-2012}.
Most often, performance is measured by how the agent acts in new and unseen situations that do not form part of its training experience.
For example, an agent might be trained to perform translation from English to French, and its training experience consist of a large collection of translated UN documents.
However, at evaluation time, it may be tested on new UN documents different from those it has seen.
Naturally, there is a gap in performance between how the agent behaves in the training experiences it has seen, and in the new situations it is evaluated on.
The ability of an agent to \emph{generalize} is measured by how small this gap in performance is.

The previous paragraph hopefully has explained what generalization is in the context of machine learning, and thus in the bigger context of AI. However, what are these
``Out of Distribution'' words remaining in the title? As mentioned, generalization consists in reducing the gap in performance of an agent acting in seen training
situations and the same agent acting in unknown test ones.
However, there are many different types of unknowns. The kind of generalization commonly addressed by statistical learning is \emph{in-distribution}:
when the data generating process from the training examples is indistinguishable from that of the test examples. In the case described above, where the translation agent is trained
on UN documents, and it is tested on \emph{new} UN documents, this is an in-distribution generalization problem.
If the translator is trained on a large collection of UN documents, but tested on user requested translations that are statistically different in any way
(such as using certain words with different frequencies, more coloquial constructions, larger or shorter sentences), then this is not an in-distribution
generalization problem.
By definition, generalization problems that are not in-distribution, are called \emph{out of distribution} (OOD) generalization problems, which are the topic of this book.

While in-distribution generalization has a massive importance, a large part of the intelligence we 
commonly attribute to humans has to do with being able to solve complex tasks in new environments or contexts \cite{lake-people-2017}.
These environments are typically different from the human's previous experience, but the human is still able to leverage its knowledge to adapt and perform well in the new setting.
Furthermore, humans are \emph{general} purpose machines: we are not able to solve a single task, but a wide range of them.
In stark contrast, current AI systems are \emph{narrow}: when they are trained to perform one or a few tasks, this is all they can do, and generalization to new tasks is in general not possible \cite{marcus2018, welling2019}.
The generality of humans is partly due to them being able to understand and make use of past experience in the new task \cite{lake-people-2017}.
For these reasons, out of distribution generalization is often discussed as being likely to have an important role in the development of general purpose intelligent machines \cite{lake-people-2017, marcus2018, welling2019}.

Now, not all the readers of this book will want to solve AI. Some (among which the author is from time to time included) will want to solve a concrete problem of practical or industrial interest.
The role of out of distribution generalization in these issues then needs to be clarified so that these readers can assess whether
this document is of any interest to them.
Given a problem, to answer the question of how necessary the study of OOD is to it, we have to ask two things (with a few obvious details like
infrastructure aside): what kind of data is available, and what do we want to generalize to?
If the data we have is of the same data generating process than the one at test, then OOD generalization is not a priori the best thing to attack the problem with.
An example instance of this is when training and test data are different UN documents, or when the task involves decision making for customers,
  and the data at training time comes from the same type of customers that will be seen at test.
However, in most real life problems, while data is abundant, it is not necessarily the right kind of data.
If we are building a machine learning tool for users, and the users' preferences vary over time, then this becomes an out
of distribution generalization problem, in which we would still like to leverage the past users' data.
Furthermore, in most cases we want to build \emph{robust} systems that perform well across many different settings. An example of this
is building a self-driving car, where our training data might come from certain cities, and we would like to build a car that generalizes
to many more cities.
Often at testing time the data is perhaps of lower quality, such as images of lower resolution or noisier, or even provided by a
malicious user that purposely wants to fool a sensitive machine learning system. 
Finally, many times we have access to massive datasets obtained by scrapping the web, such as the popular image classification datasets, 
and we want our models to generalize well to data provided by the particular users of our applications.
All of these are out of distribution generalization problems of industrial interest.

The goal of this work is simple. We want to review, contextualize, and clarify what the current knowledge in out of distribution generalization is.
As a consequence, a large part of this work will be devoted to understanding the (sometimes subtle) differences and similarities between different methods and assumptions, commonly presented in an isolated fashion.
A focus will be put in ideas that are relevant to sensory nonlinear problems such as those related to AI, or modern large scale machine learning applications.
Furthermore, we will devote special attention to studying the shortcomings of the different approaches, and what might be important next steps.

Out of distribution generalization is a problem unsolvable without assumptions on the task at hand.
If the test data is arbitrary or unrelated to the training data, then generalization is obviously futile.
Therefore, all the methods discussed make some assumption on the data.
Our goal as surveyors then is to clarify what these assumptions are, how they relate to each other, and where they fall short.
Our goal as researchers is to find assumptions that are 1) approximately satisfied in real problems of interest, and
2) that if they are approximately satisfied, we can obtain generalization improvements with a suitable algorithm.
Assumptions that are too strong will break property 1 because they won't be approximately realized, and assumptions that are too weak will 
break property 2 because they won't have meaningful practical implications.

Hence, we embark on a quest for assumptions!

\section{Outline}
This book is split into 4 chapters, without counting this tiny introduction:
\begin{itemize}
  \item In \autoref{chap:rob}, we first discuss how to quantify out of distribution generalization.
    Then, arming ourselves with several examples, we examine the relationship between out of distribution generalization and several methods commonly employed
    to tackle different out of distribution tasks. Particular emphasis will be put in highlighting the assumptions underlying these methods, and showing when they do and they
    do not work. Along this line, an important take-off is that there is no general out of distribution machine. Some assumptions are validated on some tasks,
    and completely false in others, thus we will try to highlight when a particular method may be reasonable or not according to the problem at hand.

    This chapter will also serve as a concise literature review on existing approaches for out of distribution generalization.
  \item In \autoref{chap:CIG}, we will focus on a particular class of out of distribution tasks. In these predictive tasks, as in many real problems, the difficulty in 
    generalizing out of
    distribution will lie in figuring out which correlations in the data are spurious and unreliable,
      and which ones represent the phenomenon of interest. These tasks will often require
    predictors to be robust to strong changes in context, and to be able to distinguish between cause and effect.

    This chapter is an extended version of \cite{arjovsky-ea-irm}, and as with the previous chapter, all this work has been done in collaboration
    with David Lopez-Paz, L\'eon Bottou, and Ishaan Gulrajani. In this thesis, we include a few more theoretical insights and results,
    particularly dealing with nonlinear predictors and sample complexity in terms of the needed number of training distributions. Furthermore, many things
    are rewritten to maximize clarity of exposition in a longer format, with added explanations and details making the theory more palatable.
  \item In \autoref{chap:cases}, we discuss for different applied fields the type of out of distribution tasks that appear in practice, and how these fields have dealt with those issues in the past. This chapter is not an applications chapter of the ideas in \autoref{chap:CIG} (I wish it were).
    These are all nontrivial tasks that will likely require much research to be attacked properly, and as such we will not focus on potential solutions. 
Instead, we will try to focus on the particular \emph{structure} that is available in each domain, and what shape the different OOD problems in these fields take.

    This chapter came about from a reading group on out of distribution generalization
    that we ran together with Cinjon Resnick and Will Whitney at NYU. I am particularly indebted to Will Whitney for introducing us to the works
    in the robotics section.

  \item In \autoref{chap:forward}, we lay the ground for new research areas in the context of out of distribution generalization and AI.
    In this chapter we will be concerned with agents interacting with the world in either an exploration or reinforcement learning setting, and how they could
    benefit from out of distribution generalization. 
\end{itemize}
As an important note, while in my PhD I have also worked on other topics \cite{arjovsky-ea-princ, arjovsky-ea-wgan, gulrajani-ea-wgangp, bottou-ea-geometry}, these are not going to be covered in this thesis. The reasons are multiple, but primarily because this is the work that I find most promising.

\section{Definitions and Terminology}
\label{sec:definitions}
While out of distribution generalization is relevant in many aspects of machine learning, we will instantiate our exposition in the supervised learning setup, with minor important exceptions.
This is in order to ground the exposition in concrete examples, and to avoid confusing notation that attempts to cover all areas of machine learning.
A large proportion of the ideas discussed translate verbatim to certain situations in self-supervised learning and reinforcement learning, and the relevant parallels will be provided.
Situations specific to reinforcement learning and exploration will be discussed in sections \ref{sec:robotics}, \ref{sec:rl} and \ref{sec:exploration}, and unsupervised learning will be discussed in more detail in \autoref{sec:outlook}.

We now introduce our notation:
\begin{itemize}
  \item Random variables such as inputs and outputs will be denoted by capital letters $X, Y$, and specific values will be denoted by undercaps letters $x, y$.
  \item Unless otherwise stated all vectors are column vectors.
    If $x \in \RR^d$, we will do a slight abuse of notation by denoting its corresponding row vectors as $x^T \in \RR^{1 \times d}$. Therefore, if $a, b \in \RR^d$, its inner
    product will often be written as $a^T b$.
  \item The input space will be denoted by $\matX \subseteq \RR^d$, and the space of the labels by $\matY$. The space of predicted labels will be
    denoted as $\hat \matY$. While often $\hat \matY = \matY$, this is not always the case. For instance, in binary classification $\matY = \{0, 1\}$,
    and $\hat \matY = [0, 1]$ if denoting the estimated probability of the true label being 1, or even $\hat \matY = \RR$ if $\hat Y$ contains pre-sigmoid logits.
  \item A loss function is any function that measures the difference between a predicted label and a true label. In that sense, a loss is a function
    $\ell: \hat \matY \times \matY \rightarrow \RR_+$.
    A \emph{hypothesis} or \emph{predictor} is any function $f: \matX \rightarrow \hat \matY$.
    In general, our goal will be to estimate from data a hypothesis $f$ such that $\ell(f(X), Y)$ is small in expectation under different possible test distributions.
  \item We will often have families of probability distributions $\{\PP^e\}_{e \in \envs}$ over $\matX \times \matY$. The indices $e \in \envs$ are called \emph{environments}.
    Given an environment $e$, we will denote the random variables with distribution $\PP^e$ as $(X^e, Y^e) \sim \PP^e$. When taking expectations over $(X^e, Y^e)$, it is
    then understood that the expectation is taking under the probability distribution $\PP^e$. Namely, we will use the symbol $\EE_{X^e, Y^e}$ to mean $\EE_{(X^e, Y^e) \sim \PP^e}$.
    When the distribution of the expectation is obviously $\PP^e$, such as if only terms $Y^e, X^e$ appear, we often omit all subscripts and simply use $\EE$.
  \item The \emph{risk} or expected loss for a given environment $e$ and hypothesis $f$ will be denoted as $R^e(f) := \EE_{X^e, Y^e}[ \ell(f(X^e), Y^e)]$.
  \item A \emph{featurizer} or \emph{representation} is any function $\Phi: \matX \rightarrow \hmatH$. The induced random variable $\Phi(X^e)$ will be denoted $\hat H^e$.
    A function $w: \hmatH \rightarrow \hmatY$ will be called \emph{classifier} (even if the problem is not classification but, say, regression),
    to avoid confusion with predictors or hypothesis. 
  \item We will use the term \emph{correlation} vaguely as a general way to refer to the information that one variable $X$
    carries on another variable $Y$, usually a nonlinear dependence. 
    This is done on purpose and consistent with how the machine learning field uses the term ``spurious correlations''
    A large part of this work is itself finding useful mathematical formalizations on this term and the spurious correlations
      problem.
\end{itemize}

\chapter{Robustness and Out of Distribution Generalization}
In order to judge and compare different assumptions and algorithms in the context of OOD, we first need to settle on one or a few ways to quantify
how well a predictor generalizes out of distribution.
\label{chap:rob}
\section{A Framework for Out of Distribution Generalization}
\label{sec:quantification}
There are many ways one could quantify out of distribution generalization. The most classical way to do so, and the one we will mostly employ,
is the approach taken by \emph{distributional robustness} \cite{wald1945}. This is, we only assume that the test distribution belongs to a large,
often infinite, family $\{\PP^e\}_{e \in \Eall}$, and provide the natural quantification
\begin{equation}
  R^\OOD(f) = \max_{e \in \Eall} R^e(f) \label{eq:distrob}
\end{equation}
where (as a reminder), $R^e(f) := \EE_{X^e, Y^e \sim \PP^e}\left[\ell(f(X^e), Y^e)\right]$ is the risk or expected loss for environment $e$.
Assuming that the test distribution, while unknown, belongs to $\Eall$, definition \eqref{eq:distrob} simply means we want to minimize
the worst case over all potential test distributions.

There are many things one could question in the above definition.
To begin with, one could question the need for a maximum in \eqref{eq:distrob}.
If we have access to a prior distribution over potential test environments $P(e)$, then one could think of optimizing for the quantity
$\int_{\Eall} R^e(f) P(e) \d e$. This would make perfect sense, though we find two unsurprising caveats. First of all, while prior knowledge could
be available about which test environments are more likely, it is unclear how often this prior knowledge lends itself to be expressed as a
probability distribution $P(e)$ that we can easily compute. In this book we will obviously make different types of assumptions over $\Eall$ (such as
relationships between the different environments), though expressing them as an explicit $P(e)$ would be a challenging task.
Furthermore, we have the second issue that even if we have access to a $P(e)$ that perfectly captures our prior knowledge, the different integrals
and posteriors involved in minimizing $\int_{\Eall} R^e(f) P(e) \d e$ would often be completely intractable. We elaborate on some of these ideas
in \autoref{sec:erm} and in more technical depth in \autoref{app:meta}.

Another potential concern of definition \eqref{eq:distrob} is the fact that some environments can simply be harder or more stochastic than others,
and the worst case environments would dominate \eqref{eq:distrob}. Thus, $R^\OOD$ could be unable to differentiate two predictors
that perform equally well on the hardest environments, but one performs worse than the other in an easy environment. In that direction, one
could aim to obtain a predictor minimizing \eqref{eq:distrob} and that is Pareto-optimal for the risk of each environment across $\Eall$, as does \cite{meinshausen-buhlman-2015}. Understanding how which algorithms lead to Pareto-optimal OOD predictors for other classes of problems is to us a potentially fruitful and largely unexplored avenue of reseearch.

In conclusion, while quantifying out of distribution generalization as in \eqref{eq:distrob} has many caveats, it is the simplest quantity
we could find that captures the relevant phenomena of OOD in most situations we've encountered.
After all, it is simply saying we want to perform well over all possible test distributions.

What makes out of distribution generalization a hard problem is the fact that we don't have access to samples from the test distribution $\Ptest \in \Eall$.
However, we must have something related to it if we hope to generalize. Therefore, we are in general going to assume that
we have access to data from one or a few training distributions $\{\PP^e\}_{e \in \Etrain}$. We will review some cases where $\Etrain$ contains
a single training distribution, though for the more ambitious tasks we strive to solve, we can easily see that this is insufficient.
Thus, we will often concentrate ourselves in the \emph{multiple environments} framework \cite{peters-ea-icp}, where $\Etrain$
contains more than one probability distribution. In particular, we are interested in the case where $\Etrain$ contains few distributions,
and hence we will aim to obtain strong guarantees (or at least empirical performance) in the nonasymptotic case, where the number
of training environments $|\Etrain|$ is not tending to infinity but is actually rather small (in ways we will make more precise).

As mentioned before, we are on a quest for assumptions; assumptions on our training and testing data. With the current framework, this means
that we want to find valid assumptions on $\Etrain$ and $\Eall$ such that we can make $R^\OOD$ as defined by \eqref{eq:distrob} as small as possible.
In order for this to happen, $\Etrain$ and $\Eall$ have to be related, and in general we will assume $\Etrain \subseteq \Eall$. However,
this is obviously not enough, since $\Eall$ containing some test environments which are unrelated to those in $\Etrain$ would make the problem
futile. The following section provides a few concrete examples that start to elucidate what kind of properties $\Etrain$ and $\Eall$ might
have in real problems of interest.

\section{A Few Motivating Examples}
\label{sec:examples}
Here we provide a few out of distribution generalization problems. These tasks will serve many purposes, but primarily to motivate what
the relationship between $\Etrain$ and $\Eall$ can be, even if we can only describe it in informal terms now. Second, throughout the text
they will be (perhaps simplistic) testbeds for algorithms and assumptions. In particular, we
will at times be able to calculate explicitly what the predictor returned
by an algorithm using $\Etrain$ would be, and how far away this predictor lies from generalizing out of distribution to $\Eall$.

\begin{example}\label{ex:example}
Consider the structural equation model \cite{Wright1921CorrelationAndCausation}:
\begin{align*}
    X_1^e &\leftarrow \mathrm{Gaussian}(0, 1) \\
    Y^e &\leftarrow X_1^e + \mathrm{Gaussian}(0, 1) \\
    X_2^e &\leftarrow \beta(e) Y^e + \mathrm{Gaussian}(0, 1)
\end{align*}
  Defining $X^e = (X_1^e, X_2^e)$, running the above equations as lines of code, each value $\beta(e) \in \RR$ defines a joint probability
  distribution $\PP^e$ over $(X^e, Y^e)$. By varying $\beta(e)$ over $\RR$, we obtain our infinite family of distributions $\{\PP^e\}_{e \in \Eall}$.
  For high values of $\beta(e)$, the correlation between $X_2^e$ and $Y^e$ is large, even larger than that between $X_1^e$ and $Y^e$. However, this
  correlation is \emph{spurious}: for environments with different $\beta(e)$, the correlation between $X_2^e$ and $Y^e$ can vary,
  and even change signs.

  The first thing to see is that the only linear predictor that obtains finite $R^\OOD$ under definition \eqref{eq:distrob} 
  in this setup is $f(X) = X_1$.
  This is because any linear predictor $f$ that assigns a nonzero weight to $X_2$ will have $R^e(f) \to \infty$ for test
  environments with $\beta(e) \to \infty$. Interestingly enough, the predictor $f(X) = X_1$ with optimal $R^\OOD$
  is also what we call the \emph{causal explanation}
  of $Y$: it provides the correct description about how $Y$ is created, and in particular about how it reacts in response to interventions on other variables.

  As for the training environments, we will only assume data from two training environments
  \begin{equation*}
    \Etrain = \{ \text{replace } \beta(e) \text{ by } 10, \text{ replace } \beta(e) \text{ by } 20 \}.
  \end{equation*}
  A crucial thing to note is that for both training environments in $\Etrain$, we have that $X_2$ is more reliable when using it to predict $Y$ than $X_1$ is.
  However, in the larger out of distribution class $\Eall$, there are cases where using $X_2$ would be catastrophical (e.g. when $\beta(e) = -200$).
  This is our first instance of a case where a predictor that is robust across $\Etrain$ can be far from robust across $\Eall$. However,
  we could use the hint that the correlation between $X_2$ and $Y$ (while strong) changes across the training environments and thus discard
  $X_2$, arriving to the predictor $f(X) = X_1$ and obtaining optimal $R^\OOD$. We will study such ideas in depth in the following chapter.

  Another thing to note is that since $\beta(e)$ can vary all across $\RR$, the distribution of $X_2^e$ and hence $\PP^e(X^e)$ can shift arbitrarily
  large amounts across different environments, and in particular from the training ones to a potential test one. This will be a point of failure
  for some of the methods we'll discuss.
\end{example}

\begin{example} \label{ex:cob}
As a thought experiment, consider the problem of classifying images of cows and camels \citep{beery2018recognition}.
To address this task, we label images of both types of animals.
Due to a selection bias, most pictures of cows are taken in green pastures, while most pictures of camels happen to be in deserts.
After training a convolutional neural network on this dataset, we observe that the model fails to classify easy examples of images of cows when they are taken on sandy beaches.
Bewildered, we later realize that our neural network successfully minimized its training error using a simple cheat: classify green landscapes as cows, and beige landscapes as camels. 
Here, we can consider $\Eall$ to be a family of distributions containing pictures of cows and camels on beaches or grass in any proportion.

To solve the problem described above, we need to identify which properties of the training data describe spurious correlations (landscapes and contexts), and which properties represent the phenomenon of interest (animal shapes).
For the training environments, we could assume that we have two distributions with different proportions of cows on beaches.
For instance, the pictures of cows taken in the first environment may be located in green pastures 80\pc{} of the time.
In the second environment, this proportion could be slightly different, say 90\pc{} of the time.
These environments could arise for instance because the pictures were taken in different countries: the first environment 
  in the UK where cows are mostly
  on grass, and the second environment from Goa, India, where there actually is a larger proportion of cows on beaches \cite{cobgoa}.
These two datasets reveal that ``cow'' and ``green background'' are linked by a strong, but varying (spurious) correlation, which should be discarded in order to generalize to new environments containing mostly pictures of cows on beaches.
Learning machines which pool the data from the two environments together may still rely on the background bias when addressing the prediction task.
But, we believe that all cows exhibit features that allow us to recognize them as so, regardless of their context.

This example has a peculiarity, which is that given \emph{infinite} data from a single environment, training on $\Etrain$ would prefer
to pick up the feature ``cow shape'' (which is correlated 1. with the label) rather than ``green background'' (which has a smaller correlation 0.9).
  Hence, this problem can be \emph{asymptotically} solved by minimizing $R^e(f)$ for any environment $e$ with sufficiently large support of $\PP(X^e)$.
The key issue here is the assumption of sufficient data to minimize the \emph{population} error $R^e(f)$, especially in the context of overparameterized models (when the
  number of parameters is larger than the number of datapoints).
  In \autoref{subsub:findata} will show a toy version of this cow on the beach task, which elucidates why overparameterized models
  tend to pick up on spurious correlations and fail to generalize OOD, even when the non-spurious ones have more predictive power in all training environments.
\end{example}

\begin{example}
  \label{ex:adversarial}
  Consider the problem of generalizing to adversarial examples \cite{goodfellow-ea-adversarial}.
  In this situation, we typically have access to a dataset from a single training distribution, hence $\Etrain = \{\PP^\train\}$.
  The attacker takes a sample $X \sim \PP^\train$ and performs a potentially stochastic operation $T: \matX \rightarrow \mathcal{P}(\matX)$
    on top of $X$ to obtain a test sample $X' \sim T(X)$.
  The model is evaluated on the joint distribution of $(X', Y)$, which we denote $\PP^\test$.
  Since $\PP^\test$ depends on the attack, the different types of attacks we aim to generalize to will determine the space of possible $\PP^\test$, which is $\Eall$.
  Looking for assumptions on the potential attacks and hence on $\Eall$ is necessary, since as is widely known after the seminal paper \cite{goodfellow-ea-adversarial}, simply training on $\Ptrain$
tends to generalize poorly to a variety of attacks with small enough perturbations $T(X)$ that are imperceptible to the human eye.
\end{example}

With these examples in mind, and refining them if need be, we are now able to begin to study different algorithms and assumptions to solve them.
\section{Empirical Risk Minimization} \label{sec:erm}
The simplest method we can study is the ubiquitous Empirical Risk Minimization (ERM) principle \citep{vapnik1992}.
Here, ERM would minimize the average error over all environments
\begin{equation} \label{eq:ERM}
  R^\text{ERM}(f) = \sum_{e \in \Etrain} R^e(f)
\end{equation}
In \autoref{ex:example}, ERM would grant a large positive coefficient to $X_2$ if each training environment leads to large $\beta(e)$ (as is the case in \autoref{ex:example}), obtaining catastrophic out of distribution generalization for test distributions with negative $\beta(e)$.

There are cases however where ERM makes a priori some sense.
Indeed, ERM in certain situations is what's known as an optimal learning principle.
In an in-distribution problem, \emph{without making assumptions on the distribution}, no algorithm can obtain better generalization than ERM asymptotically
\cite{vapnik1998statistical}.
Using this result, it is easy to see that if the distributions in $\Etrain$ are samples from a meta-distribution $P(e)$, then ERM is an optimal principle for minimizing
the average out of distribution error $\EE_{e \sim P(e)}[R^e(f)]$.
However, this has two issues already hinted at in the previous paragraph. First, ERM is optimal when test environments are sampled from the same distribution as training ones.
This is not the case in \autoref{ex:example} where training environments are the two with $\beta(e) = 10$ and $\beta(e) = 20$, and we want to generalize to worst case test environments such as $\beta(e) = -200$. Second, even if we don't care about worst-case OOD but the average case and with the same meta-distribution,
ERM is optimal only when making no assumptions on the meta-distribution. However, all the examples mentioned in the
previous section have an enormous amount of structure. In particular, in \autoref{ex:example} all environments share the same underlying data generating process, and indeed $\Eall$ has only one degree of freedom (the varying $\beta(e)$).
In conclusion: the assumptions of ERM are in one axis too strong
(assuming that training and test environments are sampled i.i.d. from a meta-distribution),
and in another axis too weak (assuming no structure on this meta-distribution). However, if the reader has a problem where it is safe to assume test environments come meta-i.i.d,
cares about average case OOD error, and has absolutely no other knowledge on the data, then ERM is essentially the best the reader can do.

It is important to note that ERM was never meant to be a method for out of distribution generalization.
Its theory is based on having many training points i.i.d., and little knowledge about them.
In OOD we will in general have few training distributions, and will have to assume some important amount of prior knowledge and structure about them.
We display ERM here because it is a building block throughout all machine learning, and helps elucidate where methods fall short.
The next method we study was indeed designed with out of distribution generalization in mind.

\section{Robust Optimization}
\label{sec:robopt}
The method of this section stems from a very simple and sensible idea.
Equation \eqref{eq:distrob} asks us to minimize $\max_{e \in \Eall} R^e(f)$.
Since we only have access to $\Etrain \subseteq \Eall$, from an empirical perspective it makes sense to begin by minimizing $\max_{e \in \Etrain} R^e(f)$.
\emph{Robust optimization} \cite{ben2009robust} does essentially this, it attempts to minimize the objective
\begin{equation} \label{eq:robopt}
 R^\text{rob}(f) = \max_{e \in \Etrain} R^e(f) - r_e 
\end{equation}
where the constants $r_e$ serve as environment-specific baselines.
Setting these baselines to zero leads to minimizing the maximum error across environments.
Selecting these baselines adequately prevents noisy environments from dominating optimization.
For example, \cite{meinshausen-buhlman-2015} selects $r_e = \mathbb{V}[Y^e]$ to maximize the minimal explained variance across environments.

At this moment, we want to stress a point.
It is important to separate the notions of distributional robustness \eqref{eq:distrob} as a \emph{goal}, and robust optimization \eqref{eq:robopt} as a \emph{method}.
While the method was designed to address the goal, it is not necessarily the best one.
In particular, in many cases, (perhaps unsurprisingly) other algorithms can exploit the structure of $\Etr$ and $\Eall$ in a way that's more successful at obtaining
  distributional robustness.
In other words, robustness at training time does not in general imply robustness at test time.
However, for specific problems, other conditions at training time may as well imply robustness at test time.

In order to understand the predictors returned by robust optimization, the following proposition will be very useful.
While promising, robust learning turns out to be equivalent to minimizing a weighted average of environment training errors: 
\begin{prop}
    \label{prop:robust}
    Given KKT differentiability and qualification conditions, there exist  $\lambda_e \geq 0$ such that the minimizer of $R^\mathrm{rob}$ is a first-order stationary point of $\sum_{e \in \Etrain} \lambda_e R^e(f)$.
\end{prop}
\begin{proof}
See \autoref{app:proofs}
\end{proof}
This proposition shows that robust learning and ERM (a special case of robust learning with $\lambda_e = \frac{1}{|\Etrain|}$) would never discover the desired predictor $f(X) = X_1$ in \autoref{ex:example}, obtaining infinite $R^{\text{OOD}}$.
This is because minimizing the risk of any positive mixture of environments associated to large positive $\beta(e)$'s yields a predictor with a large positive weight on $X_2$.
Unfortunately, this correlation will vanish for testing environments associated to negative $\beta(e)$, and predictors trained by minimizing $R^\text{rob}$ will fail at
generalizing out of distribution.
This is starting to highlight a key insight: the difference between interpolation and extrapolation. Methods based on robust optimization
interpolate between training environments, and thus when correlations are large but spurious it will associate large weights averaging the correlations. To solve this task
we want a method that \emph{extrapolates}: noticing that the correlation varies across environments, we can aim to discard it and thus generalize to environments whose
$\beta(e)$ vary outside the range seen in the training environments.


\section{Wasserstein Robustness and Adversarial Examples}
\label{sec:wrob}

Following up on \autoref{ex:adversarial}, assuming that $\Etrain = \{\Ptrain\}$, what assumptions do we need on the attacks (and how do they translate to asumptions on $\Eall$) such that we can design an algorithm to improve generalization? Since the inception of adversarial examples, attacks are generally considered to be constrained. In particular,
it is typically assumed that $\|T(x) - x\| < \epsilon$ for some (not necessarily Euclidean) norm $\|\cdot\|$. This is due to the fact that in general adversarial examples
are supposed to be reasonably small perturbations, almost imperceptible to the human eye.

The fact that $\|T(x) - x\| < \epsilon$ means that the distributions $\Ptrain$ and $\Ptest$ are close.
Particularly, it means that they are close according to the Wasserstein distance,
since $T$ induces a coupling between $(X', Y)$ and $(X, Y)$. Therefore, $W(\Ptrain, \Ptest) < \epsilon$ with $W$ any Wasserstein distance in $\matX \times \matY$. Thus,
$\Ptest \in B_\epsilon(\Ptrain)$, where this ball is measured in Wasserstein sense. Hence, $\Eall = B_\epsilon(\Ptrain)$ seems like a fantastic candidate for this task.
Is this enough of an assumption such that if validated in practice (which it is on this task by the above discussion), there is an algorithm with low $R^\OOD$ across $\Eall$?
The answer is largely yes, as shown by the recent paper \cite{sinha2017certifying}. We refer the reader to this paper for the relevant details.

Is the assumption reasonable for our other examples? It doesn't seem like it. In \autoref{ex:example}, replacing $\beta(e)$ by a large value will make the distribution
of $X_2^e$ vary arbitrarily and hence $\PP^e(X)$ can vary arbitrarily in Wasserstein sense.
In the cow on the beach case, the test environment has a much larger proportion of cows on beaches than cows on grass, and beaches
are clearly pixel far from grass.
Hence the assumption is violated in there as well.
This means that for these problems we will need different assumptions, and hints at the
obvious fact that not all out distribution generalization problems are equal. Different tasks will require different assumptions, which will lead to different algorithms.

\section{$KL$ Robustness}
\label{sec:klrob}
We mention here for completeness another very natural out of distribution generalization problem.
What differentiates this one from those in the previous section is that there is a strong negative result despite the assumption being seemingly reasonable, and
in particular almost identical to that of the previous section.

The previous section studied the case where there is a single training environment $\Etrain = \{\Ptrain\}$, and $\Eall = \{\PP: \text{ such that } W(\Ptrain, \PP) < \epsilon\}$ is
  the $\epsilon$-ball around $\Ptrain$ with the Wasserstein distance. Does it make sense to consider a similar situation, but when the test distribution is assumed to be
  close to $\Ptrain$ in some other distance? In particular, what if $\Eall = \{\PP: \text{ such that } D_f(\Ptrain, \PP) < \epsilon\}$ for some $f-$divergence like the
  Kullbach-Leibler divergence? It is important to recall that $f-$divergences measure how close to one the ratio of the densities of the two distributions are. For the above
  examples, this is not true in any of them. As already mentioned, $\PP^e(X)$ can vary greatly in examples \ref{ex:example} and \ref{ex:cob}. For \autoref{ex:adversarial},
  the distribution $\Ptest(X)$ after the attack can have disjoint support (even if geometrically close) to the original one, and hence all $f-$divergences would be arbitrarily
  large. Assuming $D_f(\Ptest, \Ptrain)$ to be small would mean that the support of one of the distributions contains the other, and that the densities are similar.
  This is however the case when $\Ptest$ and $\Ptrain$ contain the same possible examples, only one assigns slightly different probabilities to each example than the other one.

  A surprising result, displayed in complete detail in \cite{hu-ea-robustkl}, is that assuming $\Ptest$ is close to $\Ptrain$ in any $f-$divergence brings nothing to the table when employing a classification loss.
  Namely, we can easily rephrase their result in our framework as showing that if $\Eall = \{\PP: \text{ such that } D_f(\Ptrain, \PP) < \epsilon\}$ then
$$ R^\OOD(h) < R^\OOD(h') \text{ if and only if } R^{\Ptrain}(h) < R^{\Ptrain}(h') $$
 for any two predictors $h, h': \matX \rightarrow \hat \matY$. This means that we cannot expect any algorithm to bring an improvement over ERM if this type of closeness
 is the assumption we make. The reason for this negative result seems to be the fact that an $f-$divergence ball is in a sense too wide. It considers
 all possible distributions which affect the ratios of examples. Since we would have to consider increments in probability to any example, we cannot do anything better than
 minimize the error across all examples equally, which is what ERM does in the first place. Another way of saying this is that this assumption assumes no structure on
 $\Etrain$ and $\Eall$, in the sense that all the possible train-test shifts are considered up to a certain magnitude. This is in contrast to all examples in \autoref{sec:examples},
 where for instance in \autoref{ex:example} we have that a change in $X_1$ yields a statistical change in $Y$, and \autoref{ex:adversarial} considers test cases that are norm close (thus
 assuming geometric structure on the problem) to the training distribution. 

\section{Domain Adaptation}
\label{sec:da}
The methods of the previous two sections assume that $\Ptrain$ is close to $\Ptest$ in some sense, and this is one of the reasons why they fail to address \autoref{ex:example}
  and \autoref{ex:cob}.
The method that we will study in this section has a different assumption, namely that we can extract some variable $\hat H^e = \Phi(X^e)$ by a nonlinear function, that will be useful to predict across environments, and can be identified by assuring $\PP^e(\hat H^e)$ doesn't change across environments.

Domain adaptation \cite{bendavid-domain} considers labeled data from a source environment $e_\train$ and unlabeled data from a target environment $e_\test$ with the goal of training a classifier that works well on $e_\test$.
Many domain adaptation techniques, including the popular Adversarial Domain Adaptation \citep[ADA]{ganin2016domain}, proceed by learning a feature representation $\Phi: \matX \rightarrow \hmatH$ such that (i) the input marginals $\PP^{e_\train}(\Phi(X^{e_\train})) = \PP^{e_\test}(\Phi(X^{e_\train}))$, and (ii) the classifier $w: \hmatH \rightarrow \hmatY$ on top of $\Phi$ predicts well the labeled data from $e_\train$. This can be easily adapted to the multiple environment setting by looking for a $\Phi$ such that $\PP^e(\Phi(X^e))$ is the same for all $e \in \Etrain$.

Since the distribution of $X_1$ doesn't change in \autoref{ex:example}, while the one of $X_2$ does, we can expect domain adaptation to work well in this constrained example.
However, in \autoref{ex:cob}, the fact that the proportion of cows can change means that domain adaptation will discard this information, leading to poor prediction.
We then refine slightly \autoref{ex:example} (which is meant to be a first model of \autoref{ex:cob}) to take into account this issue, and to see how we can fix it.

\begin{example} \label{ex:toy-x1change}
  Consider the structural equation model:
\begin{align*}
    X_1 &\leftarrow \mathrm{Gaussian}(\mu(e), 1) \\
    Y &\leftarrow X_1 + \mathrm{Gaussian}(0, 1) \\
    X_2 &\leftarrow \beta(e) Y + \mathrm{Gaussian}(0, 1)
\end{align*}
\end{example}
The only difference between this example and \autoref{ex:example} is that we now allow $\EE[X_1^e]$ (which we could think as the proportion of cows in the data) to vary.
Thus, now $\Eall$ will contain all the distributions resulting from variations in $\beta(e)$ and $\mu(e)$. The training environments are now adapted to
$$ \Etrain = \{\mu(e=1) = 1, \beta(e=1) = 10; \mu(e=2) = 2, \beta(e=2) = 20\}$$
As can be seen, the discussion of sections \ref{sec:erm} and \ref{sec:robopt} still apply: at training time, in all environments the correlation between $X_2$ and $Y$
  is stronger than the one between $X_1$ to $Y$. Nonetheless, the former correlation is spurious, and using it to predict can be catastrophic at test time,
  while the predictor $f(X) = X_1$ obtains optimal out of distribution error.

The main difference between \autoref{ex:toy-x1change} and \autoref{ex:example} is that the distribution of $X_1$ now varies. This is problematic for domain adaptation, since
enforcing $\PP^e(\Phi(X^e))$ to stay the same across environments would mean that $\Phi$ has to be constant, thus destroying any predictive power.
For a more in-depth study of these failures, we encourage the avid reader to later take a look at \autoref{app:ada} where we also mathematically examine these failures and how they relate to the ideas in the next chapter.

The crucial property that $X_1$ has which makes it useful is indeed not that its marginal distribution stays the same, but the fact that the \emph{conditional} $\PP^e(Y^e | X_1^e)$ stays the same across environments.
Namely: the prediction from $X_1^e$ to $Y^e$ doesn't vary across environments, while that from $X_2^e$ to $Y^e$ does, even if the latter is better at \emph{training} time. We can use
this idea of looking for features $\hH^e$ whose conditional $\PP^e(Y^e | \hH^e)$ doesn't change across $\Etrain$ to discard spurious correlations like that of $X_2$.
This property of $\hH^e$ is what we call \emph{invariant prediction}, which is the central topic of the next chapter.
Our notion of invariant prediction builds directly from the foundational work of \cite{peters-ea-icp}, with similarities and differences that are made precise in \autoref{sec:invdef}.
Since we can see from just our two training environments that $\PP^e(Y^e | X^e_2)$ changes, looking for these
features will filter out $X_2$, use only $X_1$ and obtain optimal $R^\OOD$. As we will show, there is a deep and fundamental connection between
this notion of invariance, causality, and certain kinds of out of distribution generalization. In particular, we will see that in many cases we can
obtain out of distribution generalization by looking for features whose correlation is invariant with the label across just a few training environments.

\chapter{Causality, Invariance, and Generalization}
\label{chap:CIG}

This chapter studies the strong interplay between three ideas: causality, invariance, and out of distribution generalization.

The notion of causality dates back to some of the earliest years of philosophy, starting at least with the notion of \emph{karma} in Hinduism \cite{chapple1986karma}.
In the Mahabharata text (dated around 400 BC), Yudhishthira and Bhishma discuss whether the course of a man has been already predestined, and the existence of free will.
At one point Bhishma replicates that the future is brought into existence by human present actions given by free will, and the current circumstances that are partly set from past actions.
It is karma (intent and action) that shapes the future.
This idea, of actions or \emph{interventions} resulting in a tangible consequence has stood tall throughout centuries until the modern mathematical theories of causality \cite{pearl2009causation, rubin1974estimating}.
These theories formalize in different ways this century old idea that if we know the behaviour of a cause to its effect, we can predict the outcome when
things change due to an action.

The notion of invariance relies upon the idea that some property of a mathematical object remains unchanged \cite{cartwright2003two}.
Any precise definition of invariance then has to determine what quantities are preserved, and across what.
A particular class of invariance that we will mostly focus on is that of \emph{statistical invariance}: the assumption that some statistical patterns are preserved
across a series of distributions.
This kind of invariance (which is still only loosely defined) has been marked as a staple of reasoning and prediction, since in order to figure out what's going to happen
we have to assume that some underlying rules of the world are going to remain invariant.

Then, why are casuality, invariance and out of distribution generalization related?
Causality has often been described as having the central property of \emph{invariance under intervention} \cite{cartwright2003two}.
Namely, certain statistical quantities of the joint data distribution of the cause and the effect are preserved when we change the data distribution with an intervention or action.
In general, if a variable $H$ causes another one $Y$ but not the other way around, intervening by changing the value of $H$ will affect the distribution of $Y$
in a precise and predictable way, while changing $Y$ with an action has no effect on how $H$ changes.
For example, if we see as an intervention in \autoref{ex:example} as changing the value of $\beta(e)$, while many things in $\PP^e$ change, the conditional distribution $\PP^e(Y^e | X_1^e)$ remains stable and invariant. This is not true of $\PP^e(Y^e | X_2^e)$.
As we can see, these three ideas are intertwined: if there are features $H$ whose conditional distribution $\PP^e(Y^e | H^e)$ remains invariant, then using $H^e$ to predict
$Y^e$ will be reliable across different environments and allow us to obtain better out of distribution generalization. Furthermore, the idea of finding variables $H$
that cause $Y$ is inherently tied to the statistical prediction from $H$ to $Y$ being stable under interventions. This means that if different environments come from interventions, finding
the cause is likely to help us generalize. When interventions relate to changes in context, finding the features with stable conditionals with the label will then help us generalize to changes in context.

To us, neither causality nor invariance will ever be the goal. They will be but tools to attack out of distribution generalization. We acknowledge that this is perhaps
not what the reader wants, and finding the causal structures governing the data seems like a worthy goal by itself. To that reader we assure you that (to some extent),
that goal will be satisfied, as a strong part of this chapter will end up showing that in certain circumstances invariance, causality, and out of distribution generalization are
indeed \emph{equivalent}. This might seem strange at first, but really it's unsurprising after a certain closer look. After all, 
we won't be able to predict how $Y$ will react to interventions on the variables we don't know are causes or effects of $Y$.
Perhaps unsurprisingly, causality will often be equivalent to a precise statistical invariance, since $H$ being useful to predict $Y$ across $\Eall$ when $\Eall$
contains all distributions arising from interventions means that certain conditional distributions from $H$ to $Y$ have to be preserved.
However, a central point of this chapter is that even when no traditional notion of causality is
mathematically defined (such as a causal graph between observed variables), invariance will be a statistically testable quantity that enables generalization. And in cases
when causality makes sense, these three notions are often equivalent.

This last point makes us wonder whether we need to redefine what causality is. It is clear that our notions of causality such as graphs between observed variables
are not very useful for many situations of interest (what even is a graph between pixels?). Since invariance is equivalent to causality in some settings when
both are clearly defined, and invariance keeps the fundamental generalization properties we would expect from causation when the latter is undefined, we could consider
empowering the formalization of cause and effect from an invariance point of view. We take a stab at this epistemological qualm in \autoref{sub:causasinv}.

For now, let's make these ambitions more concrete and try to solve \autoref{ex:toy-x1change} by studying which statistical patterns are preserved across environments.

\section{Invariance and Generalization} \label{sec:invdef}
As mentioned before, the peculiarity that the optimal OOD predictor $f(X) = X_1$ for \autoref{ex:toy-x1change} has is that it uses only the feature $X_1$, which
has a stable correlation with $Y$ across environments. A more precise way of saying this is that $\EE[Y^e | X_1^e = x]$ is preserved for all $e \in \Eall$.
How then can we find these features that have stable correlations with the label?
The following observation comes to aide. In general, conditional expectations from $\hH^e$ to $Y^e$
can be written as the optimal predictor minimizing the prediction error from $\hH^e$ to $Y^e$.
\begin{obsdef}[Invariant Prediction] \label{obsdef:invpred} \ \\
  Let $(\hH^e, Y^e)$ follow a joint distribution $\PP^e$, for $e \in \envs$. Then, the two following statements are equivalent:
  \begin{enumerate}
    \item For all $\hh$ in the intersection of the supports $\text{Supp}(\PP^e(\hH^{e})) \cap \text{Supp}(\PP^{e'}(\hH^{e'}))$, we have
      $$ \EE[Y^e | \hH^e = \hh] = \EE[Y^{e'} | \hH^{e'} = \hh] \quad \text{for all $e, e' \in \envs$} $$
    \item There is a classifier $w: \hmatH \rightarrow \hmatY$ that is simultaneously optimal for all risks, i.e. 
      $$ w \in \argmin_{\bar w: \hmatH \rightarrow \hmatY} \EE\left[\ell(\bar w(\hH^e), Y^e)\right]\quad \text{for all $e \in \envs$} $$
      where $\ell: \hat \matY \times \matY \rightarrow \RR_+$ can be either the cross-entropy or mean-squared error losses.
  \end{enumerate}
  Furthermore, in the case where $Y$ is discrete (as in classification), we can replace item (A) with
  $$ \PP^{e}(Y^e | \hH^e = \hh) = \PP^{e'}(Y^{e'} | \hH^{e'} = \hh) \quad \text{for all $e, e' \in \envs$} $$
  If either item (A) or (B) are satisfied, we say that $\hH^e$ leads to an \textbf{invariant predictor} across $\envs$.
  If $\hH^e = \Phi(X^e)$, and $w$ is the classifier in item (B), we will also say $\Phi$ leads to the invariant predictor $w \circ \Phi$ across $\envs$.
\end{obsdef}

This leads to the following simple idea. To predict $Y$ from $(X_1, X_2)$, we can look at
optimal least-squares predictors $\hat{Y}^e = X_1^e\hat{\alpha}_1  +  X_2^e\hat{\alpha}_2$ for environments $e$:
\begin{itemize}
    \item regress from $X_1^e$, to obtain $\hat{\alpha}_1 = 1$ and $\hat{\alpha}_2 = 0$,
    \item regress from $X_2^e$, to obtain $\hat{\alpha}_1 = 0$ and $\hat{\alpha}_2 = \left((\mu(e)^2+2)\beta(e)\right)/\left((\mu(e)^2+2)\beta(e)^2+1\right)$,
    \item regress from $(X_1^e, X_2^e)$, to obtain $\hat{\alpha}_1 = 1 / (\beta(e)^2 + 1)$ and $\hat{\alpha}_2 = \beta(e) / (\beta(e)^2 + 1)$.
\end{itemize}

The regression using $X_1$ is our first example of an invariant correlation: this is the only regression whose coefficients do not depend on the environment $e$.
Conversely, the second and third regressions exhibit coefficients that vary from environment to environment.
This shows that we can identify the optimal predictor $f(X) = X_1$ by looking at \emph{subsets} of variables for which the least square predictor is the same across environments.
This is similar to what the seminal paper \cite{peters-ea-icp}, our main source of inspiration, does.
There, the authors define invariant prediction for a subset of variables $H$ as those that have the same residual 
distributions (the distributions of $Y^e - \EE[Y^e | H^e]$) across environments. 
While this is in general not equivalent to our notion of matching $\EE[Y^e | H^e]$ or $\PP(Y^e | H^e)$, for the 
problems of causality they study the two notions do coincide.
  Algorithmically the authors look for subsets of variables
which have the same regression residual variances across environments. However, the method of \cite{peters-ea-icp} (titled Invariant Causal Prediction, or ICP for short)
has three significant drawbacks. First, looking for subsets of variables in a $d$-dimensional setup has computational cost $\mathcal{O}\left(2^dd^3\right)$.
Second, hypothesis testing for regression residuals
inherently dooms recovering any useful predictor in the presence of mild model mispecification. While these two qualms are important, it is the third drawback that
will matter the most for us.

The third issue of ICP is that in the perceptual problems related to AI, such as \autoref{ex:cob}, looking at subsets of variables makes very little sense.
We cannot expect a fixed subset of pixels to have invariant correlations with the label across images and datasets (since the relevant objects can appear in different pixels for
different images). This becomes even more odd when looking at the causality interpretation of \autoref{ex:toy-x1change}: in this case $X_1$ is the only causal parent
of $Y$, but what even is a causal graph between pixels? In problems related to AI, causality and invariance must operate at a latent variable level: algorithms have to
\emph{learn features} (potentially nonlinear, as in the work of \cite{chalupka-ea-vcfl, heinze2017conditional}) of the inputs which themselves have stable correlations with the label.

The following theorem starts to formalize why looking for features with invariant correlations may be a good idea.
The theorem says that obtaining features with invariant prediction across $\Eall$ and low error across $\Etrain$ is enough to guarantee out of distribution generalization, i.e. low error across $\Eall$.

\begin{theo}\label{theo:invtest}
   Let $\Phi: \mathcal{X} \rightarrow \hmatH$ be a representation function, and $w: \hmatH \rightarrow \hat Y$ a classifier.
   Let's assume that
   $\ell(w(\hh), y) \leq C$ for some constant $C$ (which is trivially satisfied when $l$ is the cross entropy or mean squared
   error if the features and targets are bounded).
   Then, if $\Phi$ satisfies:
   \begin{enumerate}
     \item Approximate invariant prediction:
       $$ D\left(\PP^e(Y^e | \Phi(X^e) = \hh), \PP^{e'}(Y^{e'} | \Phi(X^{e'}) = \hh) \right) \leq \delta_{\text{INV}} $$
       for $e, e' \in \Eall, \hh \in \hmatH$, where $D$ is the total variation distance.
     \item Low training error.\footnote{This is a stronger condition than simply bounding training
       error as in the conclusion of the theorem, since we're stating that for all features, at training time the features are well
       correlated with the label. This could be replaced by standard training error, but we would have to take
       into account the fact that some examples are harder than others (i.e. $\PP(Y | \Phi(X) = \hh) $
       can have different variances for different values of $\hh$), and the test distribution can include
       a larger proportion of harder examples. This is perfectly fine, and indeed it's good that
       the framework allows for detailing generalization errors when distributions have different
       dificulties, but we opted for this exposition simply because the corrresponding result
       is easier to understand.} Let $\Ptrain \in \Eall$ be such that:
       $$ \EE_{Y \sim \Ptrain(Y | \Phi(X) = \hh)}\left[ \ell(w(\hh), Y) \right] \leq \delta_{\text{ERR}} $$
       for all $\hh \in \hmatH$.
   \end{enumerate}
   Then we have the following simple bound on the test error for any $\Ptest \in \Eall$
     $$ \EE_{(X, Y) \sim \Ptest}\left[ 
     \ell(w(\Phi(X)), Y) \right] \leq \delta_{\text{ERR}} + C \delta_{\text{INV}}$$
\end{theo}
\begin{proof}
See \autoref{app:proofs}
\end{proof}

The condition of invariance across $\Eall$ is a priori unverifiable and unatainable without assumptions, since we don't have access to $\Eall$.
What this theorem does is split our work in two:
in \autoref{sec:IRM}, we provide an algorithm that gives invariance across $\Etrain$ and low error across $\Etrain$. Then, section
\autoref{sec:causgen} studies assumptions and theorems that showcase when invariance across $\Etrain$ implies invariance across $\Eall$.
Together with \autoref{theo:invtest}, this means that our algorithm will be able to provide out of distribution generalization in these settings.


\section{Invariant Risk Minimization} \label{sec:IRM}

To discover these invariances from empirical data, we introduce Invariant Risk Minimization (IRM), a learning paradigm to estimate data representations eliciting invariant predictors $w \circ \Phi$ across multiple environments.
In this section, we have two goals in mind for the data representation $\Phi$: we want it to be useful to predict well, and elicit an invariant predictor across $\Etr$.
Mathematically, we phrase these goals as the constrained optimization problem:
\begin{equation} \label{eq:irmconst}
\begin{aligned}
    &\min_{\substack{\Phi : \mathcal{X} \to \hmatH\\ w : \hmatH \to \hmatY}} & & \sum_{e \in \Etrain} R^e(w \circ \Phi) \\
    &\text{subject to} & & w \in \argmin_{\bar w: \hmatH \rightarrow \hmatY} R^e(\bar w \circ \Phi), \text{ for all $e \in \Etrain$}.
\end{aligned}
\tag{IRM}
\end{equation}
This is a challenging, bi-leveled optimization problem, since each constraint calls an inner optimization routine. 
So, we instantiate \eqref{eq:irmconst} into the practical version: 
\begin{mdframed}[roundcorner=5pt, backgroundcolor=yellow!8]
  \begin{equation} \label{eq:irm1}
  \min_{\Phi: \mathcal{X} \rightarrow \hmatY} \sum_{e \in \Etrain} R^e(\Phi) + \lambda \cdot \| \nabla_{w|{w = 1.0}}\, R^e(w \cdot \Phi) \|^2,
  \tag{IRMv1}
\end{equation}
\end{mdframed}
where $\Phi$ becomes the entire invariant predictor, $w = 1.0$ is a scalar and fixed ``dummy'' classifier, the gradient norm penalty is used to measure the optimality of the dummy classifier at each environment $e$, and $\lambda \in [0, \infty)$ is a regularizer balancing between predictive power (an ERM term), and the invariance of the predictor $1 \cdot \Phi(x)$.

\subsection{From \eqref{eq:irmconst} to \eqref{eq:irm1}}

This section is a voyage circumventing the subtle optimization issues lurking behind the idealistic objective \eqref{eq:irmconst}, to arrive to the efficient proposal \eqref{eq:irm1}.

\subsubsection{Phrasing the constraints as a penalty}
We translate the hard constraints in~\eqref{eq:irmconst} into the penalized loss 
\begin{equation} \label{eq:irm_a}
  L_{\text{IRM}}(\Phi, w) = \sum_{e \in \Etrain} R^e(w \circ \Phi) + \lambda \cdot \mathbb{D}(w, \Phi, e)
\end{equation}
where
$\Phi : \mathcal{X} \to \hmatH$, the function $\mathbb{D}(w, \Phi, e)$ measures how close $w$ is to minimizing $R^e(w \circ \Phi)$, and $\lambda \in [0, \infty)$ is a hyper-parameter balancing predictive power and invariance.
In practice, we would like $\mathbb{D}(w, \Phi, e)$ to be differentiable with respect to $\Phi$ and $w$.
Next, we consider linear classifiers $w$ to propose one alternative.

\subsubsection{Choosing a penalty $\mathbb{D}$ for linear classifiers $w$}
Consider learning an invariant predictor $w \circ \Phi$, where $w$ is a linear-least squares regression, and $\Phi$ is a nonlinear data representation. 
In the sequel, all vectors $X^e$, $\Phi(X^e), w, v$ are column vectors.
By the normal equations, and given a fixed data representation $\Phi$, we can write $w^e_\Phi \in \argmin_{\bar w} R^e(\bar w \circ \Phi)$ as: 
\begin{equation}
    w^e_{\Phi} =
    \EE_{X^e}\left[\Phi(X^e) \Phi(X^e)^\top \right]^{-1}
    \EE_{X^e, Y^e}\left[\Phi(X^e) Y^e\right],
    \label{eq:lse}
\end{equation}
where we assumed invertibility.
This analytic expression would suggest a simple discrepancy between two linear least-squares classifiers:
\begin{equation}
    \mathbb{D}_{\text{dist}}(w, \Phi, e) = \| w - w^{e}_\Phi\|^2.
    \label{eq:penalty_dist}
\end{equation}

\begin{figure}[t!]
  \centering \includegraphics[width=\linewidth]{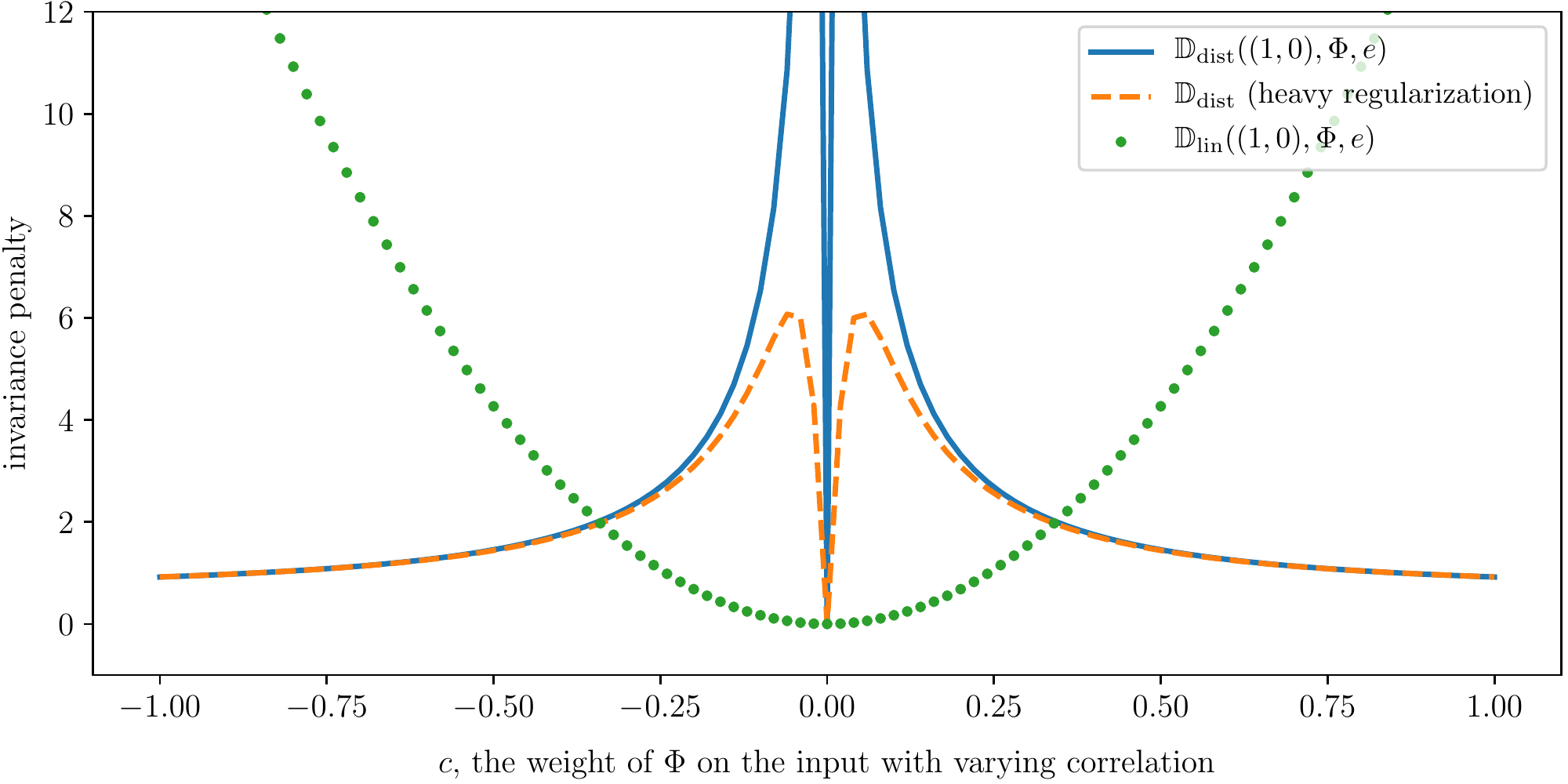} \caption{Different
  measures of invariance lead to different optimization landscapes in
  our \autoref{ex:example}.  The na\"ive approach of measuring
  the distance between optimal classifiers $\mathbb{D}_\mathrm{dist}$
  leads to a discontinuous penalty (solid blue unregularized, dashed
  orange regularized).  In contrast, the penalty
  $\mathbb{D}_{\mathrm{lin}}$
  does not exhibit these problems.} \label{fig:penalties}
\end{figure}

\autoref{fig:penalties} uses \autoref{ex:example} to show why $\mathbb{D}_{\text{dist}}$ is a poor discrepancy.
The blue curve shows \eqref{eq:penalty_dist} as we vary the coefficient $c$ for a linear data representation $\Phi(x) = x \cdot \mathrm{Diag}([1,c])$, and $w = (1, 0)$.
The coefficient $c$ controls how much the representation depends on the variable $X_2$, responsible for the spurious correlations in \autoref{ex:example}. 
We observe that \eqref{eq:penalty_dist} is discontinuous at $c=0$, the value eliciting the invariant predictor.
This happens because when $c$ approaches zero without being
exactly zero, the least-squares rule \eqref{eq:lse} compensates this change by creating vectors $w_\Phi^e$ whose second coefficient grows to infinity.
This causes a second problem, the penalty approaching zero as~$\|c\| \to \infty$.
The orange curve shows that adding severe regularization to the least-squares regression does not fix these numerical problems.

To circumvent these issues, we can undo the matrix inversion in \eqref{eq:lse} to construct:
\begin{equation}
    \mathbb{D}_{\text{lin}}(w, \Phi, e) =
     \left\| \EE_{X^e}\left[\Phi(X^e) \Phi(X^e)^\top \right]  w - \EE_{X^e, Y^e}\left[\Phi(X^e) Y^e\right]  \right\|^2,
    \label{eq:penalty_ones}
\end{equation}
which measures how much the classifier $w$ violates the normal equations.
The green curve in \autoref{fig:penalties} shows $\mathbb{D}_\mathrm{\text{lin}}$ as we vary $c$, when setting $w = (1, 0)$.
The penalty $\mathbb{D}_{\text{lin}}$ is smooth (it is a polynomial on both $\Phi$ and $w$), and achieves an easy-to-reach minimum at $c=0$ ---the data representation eliciting the invariant predictor.
Furthermore, $\mathbb{D}_{\text{lin}}(w, \Phi, e) = 0$ if and only if $w \in \argmin_{\bar w} R^e(\bar w \circ \Phi)$.
As a word of caution, we note that the penalty $\mathbb{D}_{\text{lin}}$ is non-convex for general $\Phi$.

\subsubsection{Fixing the linear classifier $w$}

Even when minimizing \eqref{eq:irm_a} over $(\Phi, w)$ using $\mathbb{D}_{\text{lin}}$, we encounter one issue.
When considering a pair $(\gamma \Phi, \frac{1}{\gamma} w)$, it is possible to let $\mathbb{D}_{\text{lin}}$ tend to zero without impacting the ERM term, by letting $\gamma$ tend to zero. 
This problem arises because \eqref{eq:irm_a} is severely over-parametrized \cite{zhang-ea-rethinking}.
In particular, for any invertible mapping $\Psi$, we can re-write our invariant predictor as
\begin{equation*}
  w \circ \Phi = \underbrace{\left(w \circ \Psi^{-1}\right)}_{\tilde{w}} \circ \underbrace{\left(\Psi \circ \Phi\right)}_{\tilde{\Phi}}.
\end{equation*}
This means that we can re-parametrize our invariant predictor as to give $w$ any non-zero value $\tilde{w}$ of our choosing.
Thus, we may restrict our search to the data representations for which all the environment optimal classifiers are equal to the same fixed vector $\tilde{w}$.
In words, we are relaxing our recipe for invariance into \emph{finding a data representation such that the optimal classifier, on top of that data representation, is ``$\tilde{w}$'' for all environments}.
This turns~\eqref{eq:irm_a} into a relaxed version of IRM, where optimization only happens over $\Phi$:
\begin{equation} \label{eq:irm_b}
    L_{\mathrm{IRM}, w=\tilde w}(\Phi) = \sum_{e \in \Etrain} R^e(\tilde{w} \circ \Phi) + \lambda \cdot \mathbb{D}_\text{lin}(\tilde{w}, \Phi, e).
\end{equation}
As $\lambda \to \infty$, solutions $(\Phi^*_\lambda, \tilde w)$ of \eqref{eq:irm_b} tend to solutions $(\Phi^*, \tilde w)$ of \eqref{eq:irmconst} for linear $\tilde{w}$.

\subsubsection{Scalar fixed classifiers $\tilde{w}$ are sufficient to monitor invariance}

Perhaps surprisingly, the previous section suggests that $\tilde{w} =
(1, 0, \ldots, 0)$ would be a valid choice for our fixed classifier.
In this case, only the first component of the data representation
would matter!  We illustrate this apparent paradox by providing a
complete characterization for the case of \emph{linear} invariant
predictors. In the following theorem, matrix $\Phi\in\RR^{p\times d}$
parametrizes the data representation function, vector $w\in\RR^p$ the
simultaneously optimal classifier, and $v=\Phi^\top w$ the
predictor $w\circ\Phi$.

\begin{theo}
  \label{theo:solutions} For all $e\in\mathcal{E}$, let
  ${R^e:\RR^d\to\RR}$ be convex differentiable cost functions.
  A vector $v\in\RR^d$ can be written $v= w \circ \Phi = \Phi^\top w$,
  where $\Phi\in\RR^{p \times d}$, and where $w\in\RR^{p}$ simultaneously
  minimize $R^e(w\circ\Phi)$ for all $e\in\mathcal{E}$, if and only
  if $v^\top \nabla {R^e(v)}=0$ for all $e\in\mathcal{E}$.
  Furthermore, the matrices $\Phi$ for which such a decomposition exists
  are those whose nullspace $\mathrm{Ker}(\Phi)$ is
  orthogonal to $v$ and contains all the~$\nabla R^e(v)$.
\end{theo}
\begin{proof}
See \autoref{app:proofs}
\end{proof}

So, any linear invariant predictor can be decomposed as linear data representations of different ranks. 
In particular, we can restrict our search to matrices $\Phi \in \RR^{1\times d}$ and let $\tilde{w}\in\RR^1$
be the fixed scalar $1.0$.
This translates~\eqref{eq:irm_b} into:
\begin{equation} \label{eq:irm_c}
    L_{\text{IRM}, w=1.0}(\Phi^\top) = \sum_{e \in \Etrain} R^e(\Phi^\top) + \lambda \cdot \mathbb{D}_{\text{lin}}(1.0, \Phi, e).
\end{equation}
\autoref{sec:causgen} shows that the existence of decompositions with high-rank data representation matrices $\Phi$ are key to out-of-distribution generalization, regardless of whether we restrict IRM to search for rank-1 $\Phi$.

\begin{figure}
  \centering \includegraphics[]{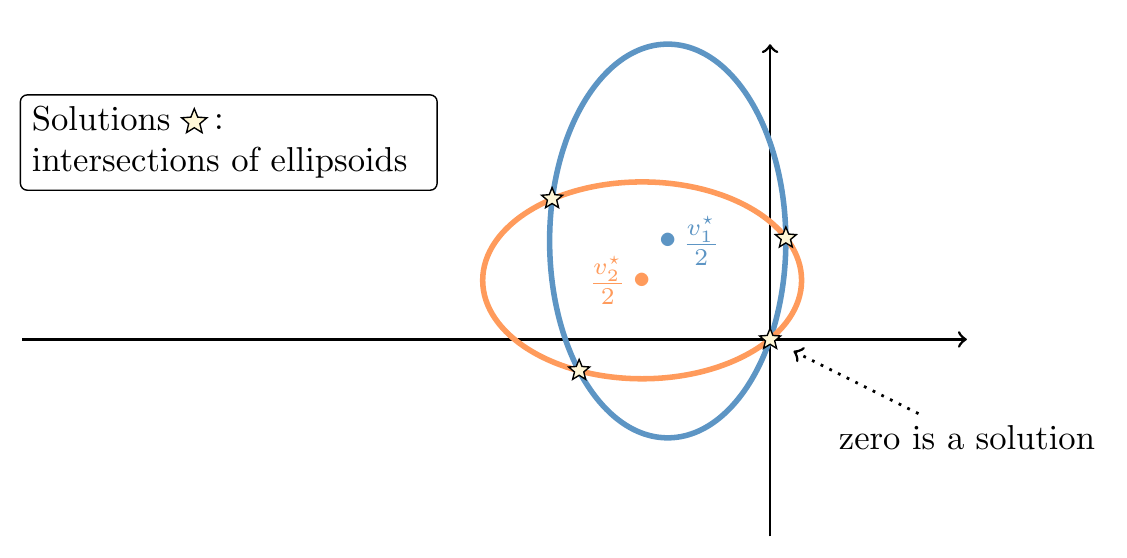}
  \caption{\label{fig:ellipsoids} The solutions of the invariant
    linear predictors $v = \Phi^\top w$ coincide with the intersection of
    the ellipsoids representing the orthogonality condition
    $v^\top\nabla{R^e(v)}=0$.}
\end{figure}

Geometrically, each orthogonality condition $v^\top \nabla R^e(v) = 0$ in \autoref{theo:solutions} defines a~$(d{-}1)$-dimensional manifold in~$\RR^d$.
Their intersection is itself a manifold of
dimension greater than $d{-}m$, where $m$ is the number of
environments.
When using the squared loss, each condition is a
quadratic equation whose solutions form an ellipsoid in
$\RR^d$. \autoref{fig:ellipsoids} shows how their intersection is
composed of multiple connected components, one of which contains the
trivial solution $v=0$. This shows that~\eqref{eq:irm_c} remains
nonconvex, and therefore sensitive to initialization.

\subsubsection{Extending to general losses and multivariate outputs}
Continuing from~\eqref{eq:irm_c}, we obtain our final algorithm~\eqref{eq:irm1} by realizing that the invariance penalty~\eqref{eq:penalty_ones}, introduced for the least-squares case, can be written as a general function of the risk, namely $\mathbb{D}(1.0, \Phi, e) = \| \nabla_{w|w=1.0} R^e(w \cdot \Phi) \|^2$, where $\Phi$ is again a possibly nonlinear data representation.
This expression measures the optimality of the fixed scalar classifier $w = 1.0$ for any convex loss, such as the cross-entropy.
If the target space $\hmatY$ returned by $\Phi$ has multiple outputs, we multiply all of them by the fixed scalar classifier $w = 1.0$.

\subsection{Implementation details}
\label{sub:implementation_details}

When estimating the objective \eqref{eq:irm1} using mini-batches for stochastic gradient descent,
one can obtain an unbiased estimate of the squared gradient norm as
\begin{equation*}
    \sum_{k=1}^b \left[ \nabla_{w|w=1.0} \ell(w \cdot \Phi(X^{e, i}_k), Y^{e, i}_k) \cdot \nabla_{w|w=1.0} \ell(w \cdot \Phi(X^{e, j}_k), Y^{e,j}_k) \right],
\end{equation*}
where $(X^{e, i}, Y^{e, i})$ and $(X^{e, j}, Y^{e, j})$ are two random mini-batches of size $b$ from environment $e$, and $\ell$ is a loss function.

\subsection{About nonlinear invariances $w$} 
\label{sub:nonlinw}

How restrictive is it to assume that the invariant optimal classifier $w$ is linear?
One may argue that given a sufficiently flexible data representation $\Phi$, it is possible to write any invariant predictor as $1.0 \cdot \Phi$.
However, enforcing a linear invariance may grant non-invariant predictors a penalty $\mathbb{D}_\text{lin}$ equal to zero.
For instance, the null data representation $\Phi_0(X^e) = 0$ admits any $w$ as optimal amongst all the \emph{linear} classifiers for all environments.
But, the elicited predictor $w \circ \Phi_0$ is not invariant in cases where $\EE[Y^e] \neq 0$.
Such null predictor would be discarded by the ERM term in the IRM objective.
In general, minimizing the ERM term $R^e(\tilde w \circ \Phi)$ will drive $\Phi$ so that $\tilde w$ is optimal amongst all predictors, even if $\tilde w$ is linear.

From an algorithmic point of view, learning nonlinear invariances is more complicated as well.
Let $\mathcal{W}$ be a subset of functions from $\hmatH$ to $\hmatY$.
We say that a penalty $\mathcal{D}(w, \Phi, e)$ enforces $\mathcal{W}$-invariance if $\mathcal{D}(w, \Phi, e) = 0$ implies that $w \in \argmin_{\bar w \in \mathcal{W}} R^e(\bar w \circ \Phi)$. Namely, the penalty is 0 implies that $w$ is optimal among all the classifiers in $\mathcal{W}$.
The penalty in \eqref{eq:irm1} then enforces \emph{linear}-invariance, or $\mathcal{W}$-invariance when $\mathcal{W}$ is the set of linear classifiers.
The way \eqref{eq:irm1} achieves this by attaining $\nabla_w R^e(w \circ \Phi) = 0$, and this only means $w \in \argmin_{\bar w \in \mathcal{W}}R^e(\bar w \circ \Phi)$ if the loss $\bar w \in \mathcal{W} \mapsto R^e(\bar w \circ \Phi)$ is convex. When $\mathcal{W}$ is the set of linear classifiers and the loss is the MSE or
cross-entropy, this is verified and then the penalty enforces that $w \in \argmin_{\bar w \in \mathcal{W}} R^e(\bar w \circ \Phi)$, i.e. linear invariance.
However, if $\mathcal{W}$ contains nonlinear classifiers, then $R^e(\bar w \circ \Phi)$ is in general not a convex function of $\bar w$, which means that the gradient penalty of
  \eqref{eq:irm1} would only make $w$ a saddle point, of which there can be many meaningless ones \cite{dauphin-ea-saddle, mescheder-ea-numerics}.
Therefore, figuring out algorithms that enforce $\mathcal{W}$-invariance
  for families $\mathcal{W}$ containing nonlinear classifiers is a completely open research problem.

We leave for future work several questions related to this issue.
Are there non-invariant predictors that would not be discarded by either the ERM or the invariance term in IRM?
What are the benefits of enforcing non-linear invariances $w$ belonging to larger hypothesis classes $\mathcal{W}$? 
How can we construct invariance penalties $\mathbb{D}$ for non-linear invariances?


\section{Generalizing with Invariance} \label{sec:causgen}
The introduced IRM principle promotes low error and invariance across training environments $\Etr$.
As mentioned before in \autoref{theo:invtest}, this is one of the two pieces needed to achieve out of distribution generalization (i.e. low error across
$\Eall$).
The missing piece is assumptions and theorems showing when and how invariance across $\Etrain$ can imply invariance across $\Eall$.
So far, we have omitted how different environments should relate to enable out-of-distribution generalization via invariance.
The answer to this question is rooted in the theory of causation.
However, causality (at least the traditional techniques for causality) will just be our starting point.

In general, we will need two things for invariance across $\Etrain$ to imply invariance across $\Eall$ and to obtain from it low out of distribution error.
\begin{enumerate}
  \item We will need an underlying unknown invariance to exist.
    We cannot expect to obtain a low error invariant solution across $\Eall$ if there is none.
  \item We will need $\Etrain$ to provide sufficient \emph{coverage}. This will mean that we will need training environments to be sufficiently diverse,
    we should have enough of them, and they should respect the underlying invariance (e.g. by having $\Etrain \subseteq \Eall$ and the above condition).
\end{enumerate}
This chapter is interested in how to formalize these assumptions in different ways, studying when they are satisfied in relevant situations, and how we can leverage them appropriately.
In particular, item B can be instantiated in many different ways, and as we will see, this can lead to drastically different sample complexities in terms of the needed number of
  environments.
  

The causality literature provides us with a well studied instance of practical interest where these two conditions can be expressed in very clear terms.
More precisely, our starting point will be the ideas in the seminal work of \cite{peters-ea-icp}.
However, we will rapidly depart from the causality language and express everything purely in invariance terms.
This is not simply a change in syntax: by doing this we will be able to express everything in purely statistical terms, and quickly build from that to generalize
  the theory and ideas to nonlinear predictors which work on perceptual inputs, and where all causality and invariance operates at a latent variable level.
The central idea at play is that causality, invariance, and out of distribution generalization are equivalent when data satisfies a causal graph.
However, invariance can be phrased in purely statistical terms and still allow for out of distribution generalization in a far broader range of settings.


\subsection{Causal graphs: a starting point for invariance and generalization}
We begin by assuming that the data from all the environments share the same underlying Structural Equation Model, or SEM \citep{Wright1921CorrelationAndCausation, pearl2009causation}:

\begin{defi}
    A \emph{Structural Equation Model (SEM)} $\matC := (\matS, N)$ governing the random vector $X = (X_1, \ldots, X_d)$ is a set of \emph{structural equations}:
    \begin{equation*}
        \matS_i : X_i \leftarrow f_i(\PA(X_i), N_i),
    \end{equation*}
    where $\PA(X_i) \subseteq \{ X_1, \ldots, X_d \} \setminus \{ X_i \}$ are called the parents of $X_i$, and the $N_i$ are independent noise random variables.
    We say that ``$X_i$ causes $X_j$'' if $X_i \in \PA(X_j)$.
    We call \emph{causal graph of $X$} to the graph obtained by drawing i) one node for each $X_i$, and ii) one edge from $X_i$ to $X_j$ if $X_i \in \PA(X_j)$.
    We assume acyclic causal graphs.
\end{defi}

By running the structural equations of a SEM $\matC$ according to the topological ordering of its causal graph, we can draw samples from the \emph{observational distribution $\PP(X)$}.
In addition, we can manipulate (intervene) a unique SEM in different ways, indexed by $e$, to obtain different but related SEMs $\mathcal{C}^e$. 

\begin{defi} Consider a SEM $\matC = (\matS, N)$. An \emph{intervention $e$ on $\matC$} consists of replacing one or several of its structural equations to obtain an intervened SEM $\matC^e = (\matS^e, N^e)$, with structural equations:
\begin{equation*}
    S^e_i : X^e_i \leftarrow f^e_i(\PA^e(X_i^e), N^e_i),
\end{equation*}
The variable $X^e$ is \emph{intervened} if $S_i \neq S^e_i$ or $N_i \neq N^e_i$.
\end{defi}

Similarly, by running the structural equations of the intervened SEM $\matC^e$, we can draw samples from the \emph{interventional distribution $\PP^e(X^e)$}.
For instance, we may consider \autoref{ex:example} and intervene on $X_2$, by holding it constant to zero, thus replacing the structural equation of $X_2$ by $X^e_2 \leftarrow 0$.
Admitting a slight abuse of notation, each intervention $e$ generates a new environment $e$ with interventional distribution $\PP^e(X^e, Y^e)$.
\emph{Valid interventions} $e$, those that do not destroy too much information about the target variable $Y$, form the set of all environments $\Eall$. 

Prior work \citep{peters-ea-icp} considered valid interventions as those that do not change the structural equation of $Y$, since arbitrary interventions on this equation render prediction impossible.
In this work, we also allow changes in the noise variance of $Y$, since varying noise levels appear in real problems, and these do not affect the optimal prediction rule.
We formalize this as follows.
\begin{defi} \label{defi:valid}
  Consider a SEM $\matC$ governing the random vector $(X_1, \ldots, X_d, Y)$, and the learning goal of predicting $Y$ from $X$. 
  Then, the \emph{set of all environments $\Eall(\matC)$} indexes all the interventional distributions $\PP^e(X^e, Y^e)$ obtainable by \emph{valid interventions} $e$.
    An {intervention $e \in \Eall(\matC)$} is valid as long as (i) the causal graph remains acyclic, (ii) $\EE[Y^e | \PA(Y)] = \EE[Y | \PA(Y)]$, and (iii) $\mathbb{V}[Y^e | \PA(Y)]$ remains within a finite range.
\end{defi}

Condition (ii) can be replaced with assuming that the causal graph stays the same, and one can intervene in the variables not containing $Y$.
This is akin to saying that the underlying causal structure of the world remains unchanged, while we are able to take actions that affect independently (in the probabilistic sense) some of the variables.
However, we found that assumption (while many times reasonable) to be slightly restrictive, since it wouldn't allow for, say, adding an arrow from a variable $X_i$ to another
one $X_j$. In particular, this could be the case when there's an intervention of the type ``let me copy the value of one variable and input it in another one'' (since this would add dependences between the variables), for example
by rerouting the flow of a fluid from one city to another.

Condition (iii) can be waived if one takes into account environment specific baselines into the definition of $R^{\mathrm{OOD}}$, similar to those appearing in the robust learning objective $R^{\mathrm{rob}}$. We leave this direction for additional quantifications of out-of-distribution generalization for future work. 

The previous definitions establish fundamental links between causation and invariance.
Namely, finding the causal parents of $Y$, finding an invariant predictor across $\Eall(\matC)$, and generalizing out of distribution across $\Eall(\matC)$ are \emph{equivalent}.
More precisely, one can show that a predictor $v: \mathcal{X} \rightarrow \mathcal{Y}$ is invariant across $\Eall(\mathcal{C})$ if and only if it attains optimal $R^{\text{OOD}}$, and if and only if it uses only the direct causal parents of $Y$ to predict, that is, $v(x) = \EE_{N_Y}\left[f_Y({\PA(Y)}, N_Y)\right]$.
We state and prove this result in \autoref{theo:cauout} for the case where we assume the causal graph stays the same and the arrow of $Y$ is unchanged.
The extension to the setup of \autoref{defi:valid} is straightforward.

An important comment at this point is that the equivalence, which means that the causal explanation $v(x) = \EE_{N_Y}\left[f_Y({\PA(Y)}, N_Y)\right]$ is the only invariant predictor acoss $\Eall(\matC)$, is only true when allowing all possible interventions
that don't affect $Y$. If instead one where to only allow a subset of all possible interventions, this would result in a
smaller (in the inclusion sense) $\Eall$, and hence a larger set of invariant predictors, of which the causal one $v$ would
be one of potentially many more.

The rest of this section follows on these ideas to showcase how invariance across \emph{training} environments can enable out-of-distribution generalization across all environments.

\subsection{Generalization theory for IRM}

As mentioned before, by \autoref{theo:invtest}, two pieces are needed to achieve out of distribution generalization (i.e. low error across
$\Eall$). The first one, privided by IRM, is low error and invariance across $\Etrain$.
We now examine the remaining condition towards low error across $\Eall$.
Namely, under which conditions does invariance across training environments $\Etrain$ imply invariance across all environments $\Eall$?
We explore this question for \emph{linear} IRM first.

Our starting point to answer this question is the theory of Invariant Causal Prediction (ICP) \citep[Theorem 2]{peters-ea-icp}.
There, the authors prove that ICP recovers the target invariance as long as the data (i) is Gaussian, (ii) satisfies a linear SEM, and (iii) is obtained by certain types of interventions. 
\autoref{theo:lingen} shows that IRM learns such invariances even when these three assumptions fail to hold.
In particular, we allow for non-Gaussian data, dealing with observations produced as a linear transformation of the variables with stable and spurious correlations, and do not require specific types of interventions or the existence of a causal graph.

Before showing \autoref{theo:lingen}, we need to make our assumptions precise.
To learn useful invariances, one must require some degree of diversity across training environments.
On the one hand, extracting two random subsets of examples from a large dataset does not lead to diverse environments, as both subsets would follow the same distribution.
On the other hand, splitting a large dataset by conditioning on arbitrary variables can generate diverse environments, but may introduce spurious correlations and destroy the invariance of interest \citep[Section 3.3]{peters-ea-icp}.
Therefore, we will require sets of training environments containing sufficient diversity and satisfying an underlying invariance.
We say that a set of training environments satisfying these conditions lie in a \emph{linear general position}.

\begin{assu}\label{a:gpos}
    A set of training environments $\Etrain$ lie in a \emph{linear general position} of degree $r$ if 
    $|\Etrain| > d - r + \frac{d}{r}$ for some $r \in \NN$,
    and for all non-zero $x \in \RR^{d}$:
    $$\dim\left(\mathrm{span}\left(\left\{\EE_{X^e}\left[{X^e} {X^e}^\top\right] x - \EE_{X^e, \epsilon^e}\left[{X^e} \epsilon^e\right]\right\}_{e \in \Etrain} \right)\right) > d - r.$$
\end{assu}
Terms $\epsilon^e$ are defined in \autoref{theo:lingen}.
Intuitively, the assumption of linear general position limits the extent to which the training environments are co-linear.
Each new environment laying in linear general position will remove one degree of freedom in the space of invariant solutions.
Fortunately, \autoref{theo:genp1} shows that the set of cross-products $\EE_{X^e}[{X^e} {X^e}^\top]$ not satisfying a linear general position has measure zero.
Using the assumption of linear general position, we can show that the invariances that IRM learns across training environments transfer to all environments.

The setting of the theorem is as follows. $Y^e$ has an invariant correlation with an unobserved latent variable $H^e$ by a linear relationship $Y^e = H^e \cdot \gamma + \epsilon^e$,
with $\epsilon^e$ independent of $H^e$. What we observe is $X^e$, which is a scrambled combination of $H^e$ and another variable $Z^e$ which can be arbitrarily correlated
with $H^e$ and $\epsilon^e$. Simply regressing using all of $X^e$ will then recklessly exploit $Z^e$ (since it gives extra, but spurious, information on $\epsilon^e$ and thus $Y^e$). A particular instance of this setting is when $H^e$ is the cause of $Y^e$, $Z^e$ is an effect, and $X^e$ contains both causes and effects. To generalize
out of distribution the representation would need to filter out $Z^e$ and use $H^e$ to predict.

In words, the implication of the theorem states the following.
If one finds a representation $\Phi$ of rank $r$ eliciting an invariant predictor $w \circ \Phi$ across $\Etrain$, and $\Etrain$ lie in a linear general position of degree $r$, then $w \circ \Phi$ is invariant across $\Eall$.  

\begin{theo} \label{theo:lingen}
    Assume that
    \begin{align*}
        Y^e   &= H^e \cdot \gamma + \epsilon^e, \quad H^e \perp \epsilon^e, \quad \EE[\epsilon^e] = 0,\\
        X^e   &= S (H^e, Z^e).
    \end{align*}
    Here, $\gamma \in \mathbb{R}^c$, $H^e$ takes values in $\mathbb{R}^c$, $Z^e$ takes values in $\mathbb{R}^q$, and $S \in \RR^{d \times (c + q)}$.
    Assume that the $H$ component of $S$ is invertible:
      that there exists $\tilde{S} \in \mathbb{R}^{c \times d}$ such that $\tilde{S} \left(S (h, z)\right) = h$, for all $h \in \RR^c, z \in \RR^q$.
    Let $\Phi\in \RR^{d \times d}$ have rank $r > 0$.
    Then, if at least $d - r + \frac{d}{r}$ training environments $\Etrain \subseteq \Eall$ lie in a linear general position of degree $r$, we have that
    \begin{equation}
      \Phi\, \EE_{X^e} \left[{X^e} {X^e}^\top \right] \Phi^\top w = \Phi\, \EE_{X^e, Y^e}\left[{X^e} Y^e\right]
        \label{eq:theo_cond}
    \end{equation}
    holds for all $e \in \Etrain$ iff $\Phi$ elicits the invariant predictor $\Phi^\top w$ for all $e \in \Eall$.
\end{theo}
\begin{proof}
  See \autoref{app:proofs}.
\end{proof}

The assumptions about linearity, centered noise, and independence between the noise $\epsilon^e$ and the causal variables $H$ from \autoref{theo:lingen} also appear in ICP \citep[Assumption 1]{peters-ea-icp}.
These assumptions imply the invariance $\EE[Y^e | H^e = h] = h \cdot \gamma$. 
Also as in ICP, we allow correlations between the noise $\epsilon^e$ and the `non-causal' variables $Z^e$, which would lead ERM into absorbing spurious correlations (as in our example, where $S = I$ and $Z^e = X^e_2$).

In addition, our result contains several novelties. 
First, we do not assume that the data is Gaussian, the existence of a causal graph, or that the training environments arise from specific types of interventions.
Second, the result extends to ``scrambled setups'' where $S \neq I$.
These are situations where the causal relations are not defined on the observable features $X$, but on a latent variable $(H, Z)$ that IRM needs to recover and filter. 
Third, we show that representations $\Phi$ with higher rank need fewer training environments to generalize.
This is encouraging, as representations with higher rank destroy less information about the learning problem at hand.

We close this section with two important observations.
First, while robust learning generalizes across interpolations of training environments (recall \autoref{prop:robust}), learning invariances with IRM buys extrapolation powers.
We can observe this in \autoref{ex:example} where, using two training environments, robust optimization yields predictors that work well for $\beta(e) \in [10, 20]$, while IRM yields predictors that work well for all $\beta \in \RR$.
Finally, the IRM loss is a differentiable function with respect to the covariances of the training environments.
Therefore, in cases when the data follows an approximately invariant model, IRM returns an approximately invariant solution, being robust to mild model misspecification.
This is in stark contrast to common causal discovery methods based on thresholding statistical hypothesis tests.

\subsection{On the nonlinear case}
\autoref{theo:nonlingen} extends our results to the nonlinear regime. In essence, it says that
if there is an invariant predictor in our function class, the function class has $q$ real parameters,
features are $p$-dimensional, and we have at least $q/p$ training environments, then invariance
at training time implies invariance at test time, and thus generalization.

The theorem is proved similarly to \autoref{theo:lingen}, namely
by a degree of freedom argument. Each extra training environment with invariant prediction removes $q$ degrees
of freedom from the space of possible solutions, which means that once we have $|\Etrain| > q / p$, we end up
with more equations than incognita, and this together with general position implies that the space of possible
solutions has collapsed to the correct one.
In the same way as in the linear version, the theorem can be adapted to deal with mild model misspecification.

The part in which \autoref{theo:nonlingen} falls short and \autoref{theo:lingen} does not is in the general position assumption.
Both assumptions mean a sensible thing, that when the number of parameters gives you $n$ incognita and enforcing invariance gives you $k$ equations, if $k > n$, there is at most one solution.
However, for the linear case we can easily prove that this assumption is satisfied outside of a measure 0 set of environments (see \autoref{theo:genp1}).
For the case of nonlinear general position, we couldn't find how to phrase such a theorem, though we believe such a statement exists.
Hence, right now \autoref{theo:nonlingen} is mostly illustrative and to pave the way for
  future work on nonlinear invariance, rather than a result than can be directly applied to real data.
We leave a better characterization of general position assumptions for nonlinear invariance as a fundamental direction for future work.

An important highlight of both \autoref{theo:lingen} and \autoref{theo:nonlingen} is that both assume the unknown underlying invariance lies approximately in our function class.
In contrast, learning theory for in-distribution generalization does not assume such a thing.
The VC dimension theory implies that in the in-distribution case, a function class with $q$ parameters and $m$ datapoints means with high probability a generalization gap 
  of at most $\mathcal{O}\left(\sqrt{\frac{q}{m}}\right)$, regardless of whether we are close or far to the underlying solution.
There are in general two ways of satisfying that we approximately have the right representation in our function class.
The first one is by having prior knowledge on the underlying invariance: \autoref{theo:lingen} is of this kind, assuming that the underlying invariance is linear.
The second one, is that the function class is sufficiently expressive, which is the case in \autoref{theo:nonlingen} as we increase the capacity of the function class (say, by
using sufficiently large neural networks). However, in the latter one if we increase the capacity of our models naively, we are likely to need more environments to generalize.
In a similar way to current large scale machine learning, it is likely that to learn powerful invariances we will need a good combination of expressivity,
like neural networks provide, and good inductive biases such as the equivariance provided by convolutions in CNNs. However, it is likely that inductive biases
for in-distribution generalization and out-of-distribution generalization will be of different kinds. For instance, as we will ellaborate in \autoref{sec:outlook},
using SGD with overparameterized models is a fantastic regularizer for in-distribution generalization, though sadly leads to models picking up on spurious correlations
and worsening out-of-distribution generalization.

Both \autoref{theo:lingen} and \autoref{theo:nonlingen} are two starting points for a learning theory of invariance, and provide nonasymptotic guarantees in the
number of environments. Unfortunately, they are both very pessimistic, with the needed number of environments scaling with the number of parameters.
Next, we exploit more structure to provide assumptions and guarantees of generalization even in the case where we only have two training environments, and arbitrary featurizers.

\subsection{Generalizing with fewer environments}
While general, \autoref{theo:lingen} is pessimistic, since it requires the number of training environments to scale linearly with the number of parameters in the representation matrix $\Phi$. 
In order to show that this is not always the case, we conclude with a family of problems in which 2 training environments
is sufficient, even in the presence of arbitrarily expressive featurizers. The nonlinear colored MNIST problem in the
next section is a particular case of this family.

In this family of problems, the target $Y^e$ has both causes $H^e$ and effects $Z^e$ which are unobserved. The input $X^e$
is a perceptual representation of both cause and effects, and we will show that any invariant predictor from $X^e$
has to discover the true, latent, causal information.

\begin{theo} \label{theo:nonlingen2} 
  Consider a binary target $Y^e \in \{0, 1\}$ with unobserved discrete cause $H^e$ and binary effect $Z^e$.
  The input, $X^e = F(H^e, Z^e)$ is an unknown invertible function of these unobserved variables.
  In the language of SEMs, we assume the data follows the generative process
\begin{align*}
  H^e &\sim P^e_H \\
  Y^e &\leftarrow \Psi(H^e) \oplus \epsilon_Y \\
  Z^e &\leftarrow Y \oplus \epsilon_Z^e \\
  X^e &= F(H^e, Z^e)
\end{align*}
  where $\epsilon_z^e \sim \text{Ber}(p^e_Z)$ and $\oplus$ is the XOR operator.
  The joint distribution of all variables for environment $e$ is then determined by $P^e_H \in \RR^k$ and $p_Z^e \in [0, 1]$.
  Let $\Etrain = \{e, e'\}$. Then, for any pair $(P^e_H, P^{e'}_H)$ (which could be equal or not), for all pairs $(p_Z^e, p_Z^e) \in [0,1]^2$ outside of
  a set of measure 0 in $[0, 1]^2$,
  if $\Phi: \mathcal{X} \rightarrow \hmatH$ leads to an invariant predictor across $\Etrain$, then $\Phi(F(H^e, Z^e)) = G(H^e)$ for some function $G$.
  That is, $\Phi$ discards the noncausal information. Furthermore, the $\Phi$ that leads to invariant prediction
  with minimal error across $\Etrain$ satisfies that $\Phi(F(H^e, Z^e)) = G(H^e)$
  for $G$ invertible, i.e. $\Phi$ exactly recovers the cause, and leads to the optimal invariant predictor $\Psi \circ G^{-1} \circ \Phi$ across $\Eall$.
\end{theo}
\begin{proof}
  See \autoref{app:proofs}.
\end{proof}

The proof of \autoref{theo:nonlingen2} exploits more structure than the proofs of the previous two theorems.
In this proof, we write $\EE[Y^e | \Phi(X^e)]$ explicitly and show it cannot be the same for two environments unless they are essentially the same (i.e. $p_Z^e = p_Z^{e'}$)
or $\Phi$ is the right causal explanation.
Matching $\EE[Y^e | \Phi(X^e)]$ explicitly would mean enforcing a stronger invariance than a \emph{linear} one since the conditional expectation is the optimal classifier
among \emph{all} classifiers.
The discussion from \autoref{sub:nonlinw} gains relevance here, since enforcing $\mathcal{W}$-invariance for larger families $\mathcal{W}$ should allow discarding more non-invariant predictors with fewer training environments. 

We believe this theorem can be significantly extended
and merged with the approaches of \autoref{theo:lingen} and \autoref{theo:nonlingen}.
All in all, studying what problems allow the discovery of invariances from fewer environments is a promising line of work towards a learning theory of out of distribution generalization.

Before providing targeted experiments to assess the effectiveness of IRM, we conclude this section with a few thoughts about invariance and causation.

\section{Related Work}
\label{sub:causasinv}
Now that we have introduced the relevant material and explained our main contributions, we can discuss how this relates to prior work,
  and contextualize what has been done and what is missing.

The relationship between causation and out of ditribution generalization has been hinted at least as long as causation has been
  approached from a mathematical perspective, particularly the fact that a causal model allows one to predict the outcome of an
  intervention.
In particular, Rubin \cite{rubin1974estimating} was one of the pioneers in doing so.
Rubin considered the problem of learning a predictor $f(X, T)$ from a patient $X$ and treatment $T$ to the patient's outcome $Y$.
He encountered the following scenario: when trained on a single data ditribution the model was making significant errors after
  changing the treatments at test time. More precisely, the model was making errors on pairs $(x, 1)$ on patients $x$ that had
  a high probability of being under treatment 0 in the training distribution (i.e. $(x, 0)$ had high probability under the training data).
He called this a ``missing data'' problem, by making the reasonable assumption that if we had sufficient data from all patients
  under all possible treatments we would be able to predict their outcome.
This assumption, called ``strong ignorability'' is formally written as saying that the single training distribution $\Ptrain$ has
  full support over all $(\mathcal{X} \times \mathcal{T})$, and that $\PP(Y | X, T)$ is invariant (or, in the language of causal 
  graphs, that $(X, T)$ are the causal parents of $Y$.
In the OOD framework we presented, he was considering the setting of $\Etrain = \{\Ptrain\}$, and $\Eall = \Eall(\mathcal{C})$ with 
  $\mathcal{C}$ being the causal graph of $X, T, Y$ such that $X$ causes $T$, and $X, T$ both cause $Y$.
An important comment then is that under the strong ignorabilty assumption, ERM will solve the problem asymptotically, since 
  the training distribution has full support and the conditional probability from input to output is invariant.
This is drastically different from other causality problems like \autoref{ex:example}, where ERM exploits parts of the data
  which have spurious correlations.
However, the problem that Rubin discussed is important when only finite (and very skewed) training data is available, 
  and the methods he designed (based on reweighting examples) have been very successful for his particular problem at hand.

The work of Pearl (see \cite{pearl2009causation} and the references therein) went much more general by (among other contributions)
  proposing the theory 
  of general causal graphs, defining interventional distributions on graphs,
  and studying which aspects of the causal graph can and cannot be recovered from observational data.
Furthermore, Pearl helped clarify the fact that for general cusal graphs and observational data,
  ERM will pick up on things like consequences of the 
  target and confounders to minimize its error, and will fail to predict when testing under different interventions.
He made clear that for general graphs, if one wants to generalize to interventional distributions, one must know the causal parents.
In this way, he showed the importance of finding the causal parents for a large family of problems, and departed from the
  assumptions of Rubin (which assumed the inputs are the causal parents).
He also showed that unless one has access to interventional data, in general it is impossible to recover more than the (undirected) skeleton
  of the causal graph and its V-structures.
Several algorithms for doing such a thing, in particular the widely used PC algorithm \cite{spirtes2000}, became readily available shortly
  after.
Pearl's theorems on the impossibility of recovering causal structures from observational data
  nonetheless were very pessimistic in making almost no assumptions on the kinds of distributions at hand.
Because of this, a dychotomy was presented of either having access to interventional data (which was generally equated to being 
  able to perform interventions, typically impossible in many problems), or being unable to retreive the causal parents.
This is where the work of Peters, Mooij, Sh\"{o}lkopf, and their collaborators came into play, by making stronger assumptions on the training data and thus paving the way for \emph{causal discovery}.

In the work of additive noise models \cite{peters-ea-anm}, Peters et al. realized that one can distinguish whether $X$
  causes $Y$ or $Y$ causes $X$ using only observational data if assuming that their distributions are $\emph{not}$ Gaussian,
  and the relationship between the variables is linear and the noise is additive (i.e. the structural equations were either
  $Y = aX + \epsilon$ or $X = aY + \epsilon$).
Several extensions of this idea have been preseented, including versions with nonlinear dependence and nonadditive noise \cite{peters-ea-causal}.
Nonetheless, their particular distributional assumptions are in general difficult or impossible to verify in practice, 
  and the algorithms typically scale superexponentially with the size of the causal graph when trying to find causal parents.

Later, the seminal work of \cite{peters-ea-icp} (Invariatn Causal Prediction, or ICP) departed from the dycothomy of assuming either
  only observational data, or full access to interventions.
In particular, they introduced the \emph{multiple environments} framework, where one assumes access to different pools of data,
  each one corresponding to different sets of interventions (which don't need to be known).
This is substantially less than assuming the ability to intervene, since it is not uncommon in practice to have data of a system observed
  under different conditions, (e.g. having medical data from multiple hospitals, images from different cities, and so forth).
Furthermore, this assumption of having multiple environments is enough to break the nonidentifiability theorems from Pearl, and
  one also avoids the stringent distributional assumptions of additive noise models.
In this way, it struck much more sensible middle ground between identifiability and having reasonable assumptions on the type of data
  available.
The central idea of ICP was that under the multiple environment setting, one could exploit the fact that the only subset of 
  variables in a causal graph whose regression residuals to $Y$ are distributionally \emph{invariant} are the causal parents of $Y$.
In practice, they chose to look for subsets of variables such that on the training environments, the linear regression residuals
  had the same variance across environments under hypothesis tests,
  and provided careful sensible bounds on the number of environment and samples required
  to retrieve the causal parents.
Furthermore, the work of \cite{heinze2018invariant} lifted the linearity requirement of ICP by using nonlinear regressions 
  and then testing for the residual distributions to match.
While a gigantic progress, a few things hindered the usability of ICP and its nonlinear version. First, the use of statistical hypothesis
  tests meant that the algorithms completely break under mild model mispecification: if the data only approximately (in any weak distributional
  sense) satisfies a causal graph, then the algorithms return ``nothing is causal'' which, while true, doesn't allow us 
  to obtain a predictor that approximately generalizes out of distribution. Second, looking for subsets of variables has exponential
  cost as a function of the number of covariates. Third and most importantly, one has to assume that the observed variables are themselves
  the nodes in a causal graph, rather than a scrambled version of latent features that generalize well. In essence, they don't allow 
  us to learn representations which satisfy the invariance and generalization properties of causal parents.

The problem of learning \emph{causal features} or representations was first attacked by \cite{chalupka-ea-vcfl} and \cite{heinze2018invariant}. However, \cite{chalupka-ea-vcfl} presents a different definition of causal, and in particular the causal parents of a variable
only satisfy their criterion of causal in certain particular settings. Furthermore, in general there is not a clear relationship between
their definition of causality and out of distribution generalization. The paper of \cite{heinze2018invariant}, however, does 
attack the out of distribution problem by finding features that satisfy a particular kind of invariance equivalent to finding the causal
explanation of the target variable. Furthermore, it addresses the mentioned problems of ICP, albeit at the cost of requiring more supervision, as we now detail.

The authors of \cite{heinze2018invariant} assume access to datapoints of the form $(X, \ID, Y)$, where the goal is again to predict 
  $Y$ from $X$, and as such at test time we only observe $X$.
Using our notation, one can phrase their assumptions as asking that there is aun unknown
random variable $H$ (called $X^{\text{core}}$ in their work)
such that for all environments $e \in \Eall$
therE is a joint distribution $(X^e, \ID^e, H^e, Y^e)$ which satisfies that
i)$\PP(Y^e | H^e)$ is invariant across $\Eall$, and ii) for any two given tuples $(X, \ID, H, Y)$, $(X', \ID', H', Y')$ we have that
if $\ID = \ID'$ then $H = H'$.
Therefore, the ID acts as an identifier saying that if two datapoints have the same ID, then they must have the same features that
lead to invariant prediction.
For the case when the data satisfies a causal graph, this means that datapoints with the same ID must have the same values on the causal
explanation of $Y$.
Nonetheless, these assumptions still make sense when there is no causal graph, such as when $Y$ is a person's identity in a picture $X$, 
and we obtain matching ID's for some pairs of datapoints corresponding to the same person but, for instance, different lightning conditions.
Importantly, while $\ID = \ID'$ implies that $H = H'$, the opposite is not true, so they don't assume they have disentangling ID's for
all datapoints.
Under these assumptions, the authors learn a representation $\Phi(X)$ that is trained with a penalty of the form
$$ \sum_{(x, \ID)}\left(\left( \sum_{x' \in \mathcal{X}(\ID)} \Phi(x') \right) - \Phi(x)\right)^2 $$
where $\mathcal{X}(\ID)$ is the set of $x$ points such that $(x, \ID)$ is in the training data.
Namely, they penalize the variance of $\Phi(X)$ conditioned on the ID, directing the model to have constant features for 
  those datapoints that have the same ID and hence the same $H$.
The work of \cite{heinze2018invariant} comes with strong guarantees and performance. 
In the $d$-variable causal graph
  setup they need $d+1$ \emph{datapoints} with matching IDs, this is in contrast to IRM where we need a bit more than $d$ \emph{environments} and thus substantially more datapoints.
This sample complexity, however, comes at a price: needing to have labeled IDs, which in the causal case is equivalent to having
two datapoints with the exact same causal explanation value. Since the causal parents are unknown, even if we can do arbitrary interventions, it is in general not possible to create points with matching IDs. Thus, it requires substantially more supervision than the 
multiple environment setup, but for the problems where this supervision is available, its performance cannot be matched.
Furthermore, by approximating a continuous optimization problem, grouping by ID is (to the best of our knowledge) the first causal 
learning algorithm to not be bound to exponential time constraints, be robust to mild model mispecification, and being able to recover
invariant predictors even when there is no causal graph, or when the graph doesn't take place at the observed variable level. 

As can be seen from this related work, the invariance view of causation transcends some of the difficulties raised by working with causal graphs.
For instance, the ideal gas law $PV = nRT$ or Newton's universal gravitation $F = G\frac{m_1 m_2}{r^2}$ are difficult to describe using structural equation models (\emph{What causes what?}), but they are prominent examples of laws that are invariant across experimental conditions.
When collecting data about gases or celestial bodies, the universality of these laws will manifest as invariant correlations, which will sponsor valid predictions across environments, as well as the conception of scientific theories.

\section{Experiments}
\label{sec:experiments}

We perform two experiments to assess the generalization abilities of IRM across multiple environments.
The source-code is available at\\ \url{https://github.com/facebookresearch/InvariantRiskMinimization}.

\subsection{Synthetic data}

As a first experiment, we extend our motivating \autoref{ex:example}.
We use a slightly different variant of \autoref{ex:example} coming from \cite{arjovsky-ea-irm}, in which the data follows the structural equation model
\begin{align*}
  X_1 &\leftarrow \mathrm{Gaussian}(0, \sigma(e)^2), \nonumber \\
    Y &\leftarrow X_1 + \mathrm{Gaussian}(0, \sigma(e)^2), \\
    X_2 &\leftarrow Y + \mathrm{Gaussian}(0, 1).\nonumber
\end{align*}
and training environments come from changing the value of $\sigma(e)^2$. Here, larger values of $\sigma(e)$ mean that $X_2$ is more correlated with $Y$, albeit the
correlation is spurious since it varies when $\sigma(e)$ varies. Furthermore, the correlation vanishes for low values of $\sigma(e)$. In that sense, $\sigma(e)$ plays a similar
role to $\beta(e)$ in \autoref{ex:example}. This example has the added difficulty that the noise value of $Y$ varies per environment. However, it has the pedagogical complication
that $\Eall$ includes much more environments than just different values of $\sigma$, but includes replacing the arrows of $X_1$ and $X_2$ by arbitrary interventions.

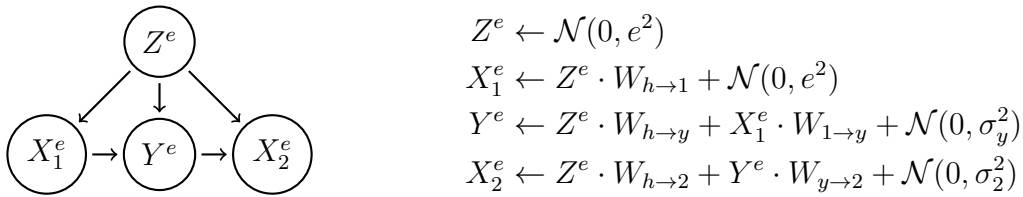
\begin{figure}[ht]
  \centering
  \begin{subfigure}[b]{0.32\linewidth}
    \begin{tikzpicture}[]
        \node[draw=black, thick, circle] (H) at (0, 1.5) {$Z^e$};
        \node[draw=black, thick, circle] (X) at (-1.5, 0) {$X_1^e$};
        \node[draw=black, thick, circle] (Y) at (0, 0) {$Y^e$};
        \node[draw=black, thick, circle] (Z) at (1.5, 0) {$X_2^e$};
        \draw [->, thick, shorten <= 2pt, shorten >= 2pt] (H) -- (X);
        \draw [->, thick, shorten <= 2pt, shorten >= 2pt] (H) -- (Y);
        \draw [->, thick, shorten <= 2pt, shorten >= 2pt] (H) -- (Z);
        \draw [->, thick, shorten <= 2pt, shorten >= 2pt] (X) -- (Y);
        \draw [->, thick, shorten <= 2pt, shorten >= 2pt] (Y) -- (Z);
    \end{tikzpicture}
  \end{subfigure}
  \begin{subfigure}[b]{0.6\linewidth}
    \begin{align*}
        Z^e &\leftarrow \mathcal{N}(0, e^2)\\
        X_1^e &\leftarrow Z^e \cdot W_{h \to 1} + \mathcal{N}(0, e^2)\\
        Y^e &\leftarrow Z^e \cdot W_{h \to y} + X_1^e \cdot W_{1 \to y} + \mathcal{N}(0, \sigma^2_y)\\
        X_2^e &\leftarrow Z^e \cdot W_{h \to 2} + Y^e \cdot W_{y \to 2} + \mathcal{N}(0, \sigma^2_2)
    \end{align*}
  \end{subfigure}
    \caption{In our synthetic experiments, the task is to predict $Y^e$ from $X^e = (X^e_1, X^e_2) \cdot S$ across multiple environments $e \in \mathbb{R}$.}
\label{fig:chains_definitions}
\end{figure}

First, we increase the dimensionality of each of the two input features in $X = (X_1, X_2)$ to $10$ dimensions.
Second, as a form of model misspecification, we allow the existence of a $10$-dimensional hidden confounder variable $Z$.
Third, in some cases the features $X$ will not be directly observed, but only a scrambled version $X \cdot S$, as in \autoref{theo:lingen}.
\autoref{fig:chains_definitions} summarizes the SEM generating the data $(X^e, Y^e)$ for all environments $e$ in these experiments.
More specifically, for environment $e \in \mathbb{R}$, we consider the following variations:
\begin{itemize}
    \item \emph{Scrambled} (S) observations, where $S$ is an orthogonal matrix, or\\
          \emph{unscrambled} (U) observations, where $S = I$.
    \item \emph{Fully-observed} (F) graphs, where $W_{h \to 1} = W_{h \to y} = W_{h \to 2} = 0$, or\\
          \emph{partially-observed} (P) graphs, where $(W_{h \to 1}, W_{h \to y}, W_{h \to 2})$ are Gaussian.
    \item \emph{Homoskedastic} (O) $Y$-noise, where $\sigma_y^2 = e^2$ and $\sigma_2^2 = 1$, or\\
          \emph{heteroskedastic} (E) $Y$-noise, where $\sigma_y^2 = 1$ and $\sigma_2^2 = e^2$.
\end{itemize}

These variations lead to eight experimental setups that we will denote by their initials.
For instance, the setup ``FOS'' considers fully-observed (F), homoskedastic $Y$-noise (O), and scrambled observations (S).
For all variants, $(W_{1 \to y}, W_{y \to 2})$ have Gaussian entries.
Each experiment draws $1000$ samples from the three training environments $\Etrain =  \{0.2, 2, 5\}$.
IRM follows the variant \eqref{eq:irm1}, and uses the environment $e=5$ to cross-validate the invariance regularizer $\lambda$.
We compare to ERM and ICP \citep{peters-ea-icp}.

\begin{figure}[]
    \centering
    \includegraphics[width=\linewidth]{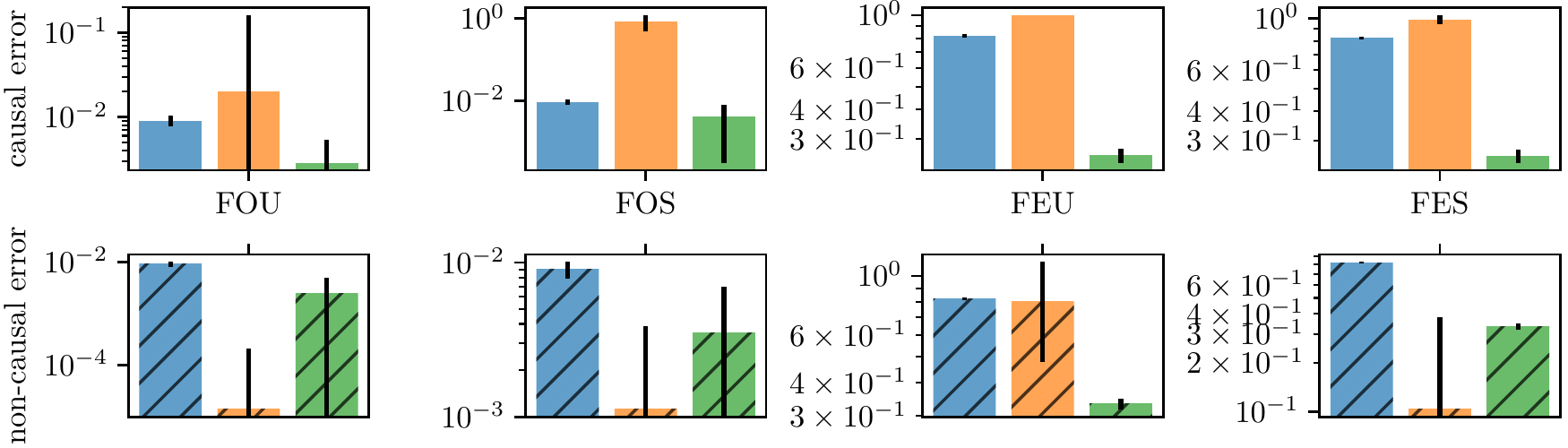}
    \vskip 0.5cm
    \includegraphics[width=\linewidth]{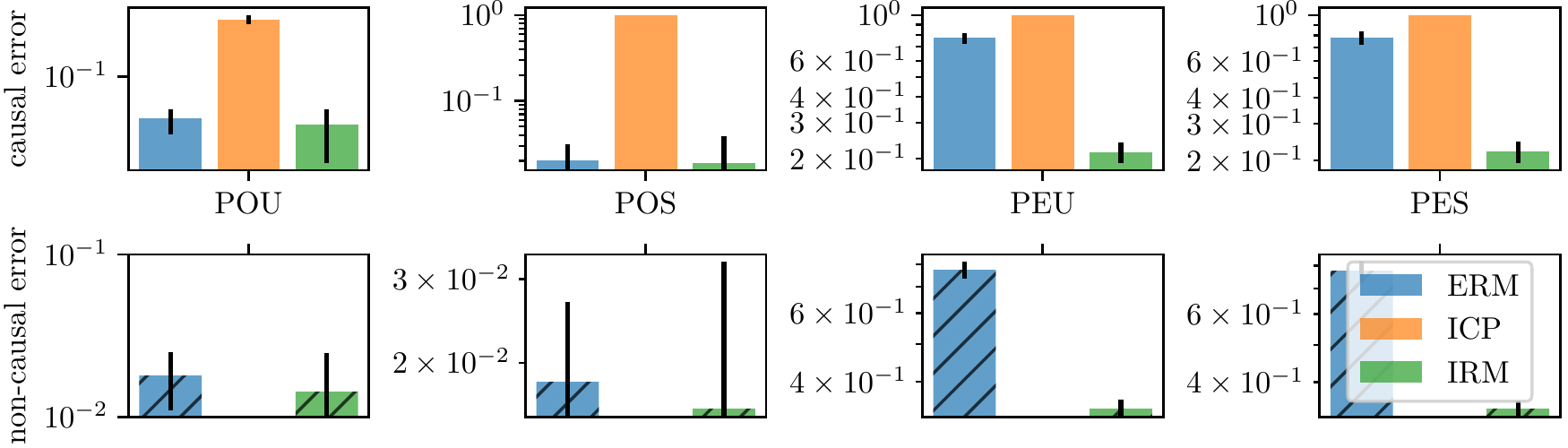}
    \caption{Average errors on causal (plain bars) and non-causal (striped bars) weights for our synthetic experiments. The $y$-axes are in log-scale. See main text for details.}
    \label{fig:chains_results}
\end{figure}

\autoref{fig:chains_results} summarizes the results of our experiments.
We show two metrics for each estimated prediction rule $\hat{Y} = X_1 \cdot \hat{W}_{1 \to y} + X_2 \cdot \hat{W}_{y \to 2}$.
To this end, we consider a de-scrambled version of the estimated coefficients, $(\hat{M}_{1 \to y}, \hat{M}_{y \to 2}) = S^\top (\hat{W}_{1 \to y}, \hat{W}_{y \to 2})$.
First, the plain barplots shows the average squared error between $\hat{M}_{1 \to y}$ and $W_{1 \to y}$.
This measures how well does a predictor recover the weights associated to the causal variables.
Second, each striped barplot shows the norm of estimated weights $\hat{M}_{y \to 2}$ associated to the non-causal variable.
We would like this norm to be zero, as the desired invariant causal predictor is $Y^e = (X_1^e, X_2^e) \cdot S^\top(W_{1 \to y}, 0)$.
In summary, IRM is able to estimate the most accurate causal and non-causal weights across all experimental conditions.
In most cases, IRM is orders of magnitude more accurate than ERM (our $y$-axes are in log-scale).
IRM also out-performs ICP, the previous state-of-the-art method, by a large margin. 
Our experiments also show the conservative behaviour of ICP (preferring to reject most covariates as direct causes).
This leads ICP into large errors on causal weights, and small errors on non-causal weights.

\subsection{Colored MNIST}
\label{sec:color_mnist}

We validate our method for learning nonlinear invariant predictors on a synthetic binary classification task derived from MNIST.
The goal is to predict a binary label assigned to each image based on the digit.
Whereas MNIST images are grayscale, we color each image either red or green in a way that correlates strongly (but spuriously) with the class label.
By construction, the label is more strongly correlated with the color than with the digit, so any algorithm which purely minimizes training error will tend to exploit the color.
Such algorithms will fail at test time because the direction of the correlation is reversed in the test environment.
By observing that the strength of the correlation between color and label varies between the two training environments, we can hope to eliminate color as a predictive feature, resulting in better generalization.

We define three environments (two training, one test) from MNIST transforming each example as follows:
first, assign a preliminary binary label $\tilde y$ to the image based on the digit: $\tilde y=0$ for digits 0-4 and $\tilde y=1$ for 5-9.
Second, obtain the final label $y$ by flipping $\tilde y$ with probability 0.25.
Third, sample the color id $z$ by flipping $y$ with probability $p^e$, where $p^e$ is 0.2 in the first
environment, 0.1 in the second, and 0.9 in the test one. Finally, color the image red if $z = 1$ or green if $z = 0$.

We train MLPs on the colored MNIST training environments using different objectives and report results in \autoref{tab:colored-mnist-results}.
For each result, we report the mean and standard deviation across ten runs.
Training with ERM results in a model which attains high accuracy in the training environments but below-chance accuracy in the test environment since the ERM model classifies mainly based on color.
Training with IRM results in a model that performs worse on the training environments, but relies less on the color and hence generalizes better to the test environments.
For comparison, we also run ERM on a model which is perfectly invariant by construction because it pre-processes all images to remove color.
We find that this oracle outperforms our method only slightly.

\begin{table}[t!]
\centering
\begin{tabular}{ l r r }
  \toprule
  \textbf{Algorithm} & \textbf{Acc. train envs.} & \textbf{Acc. test env.} \\
  \midrule
  ERM & $87.4 \pm 0.2$ & $17.1 \pm 0.6$ \\
  \textbf{IRM (ours)} & $70.8 \pm 0.9$ & $\mathbf{66.9 \pm 2.5}$ \\
  \midrule
  Random guessing (hypothetical) & 50 & 50 \\
  Optimal invariant model (hypothetical) & 75 & 75 \\
  ERM, grayscale model (oracle) & $73.5 \pm 0.2$ & $73.0 \pm 0.4$ \\
  \bottomrule
\end{tabular}
\caption{Accuracy (\pc{}) of different algorithms on the Colored MNIST synthetic task.
         ERM fails in the test environment because it relies on spurious color correlations to classify digits.
         IRM detects that the color has a spurious correlation with the label and thus uses only the digit to predict,
         obtaining better generalization to the new unseen test environment.}
\label{tab:colored-mnist-results}
\end{table}

\begin{figure}[]
  \centering
  \includegraphics[width=\linewidth]{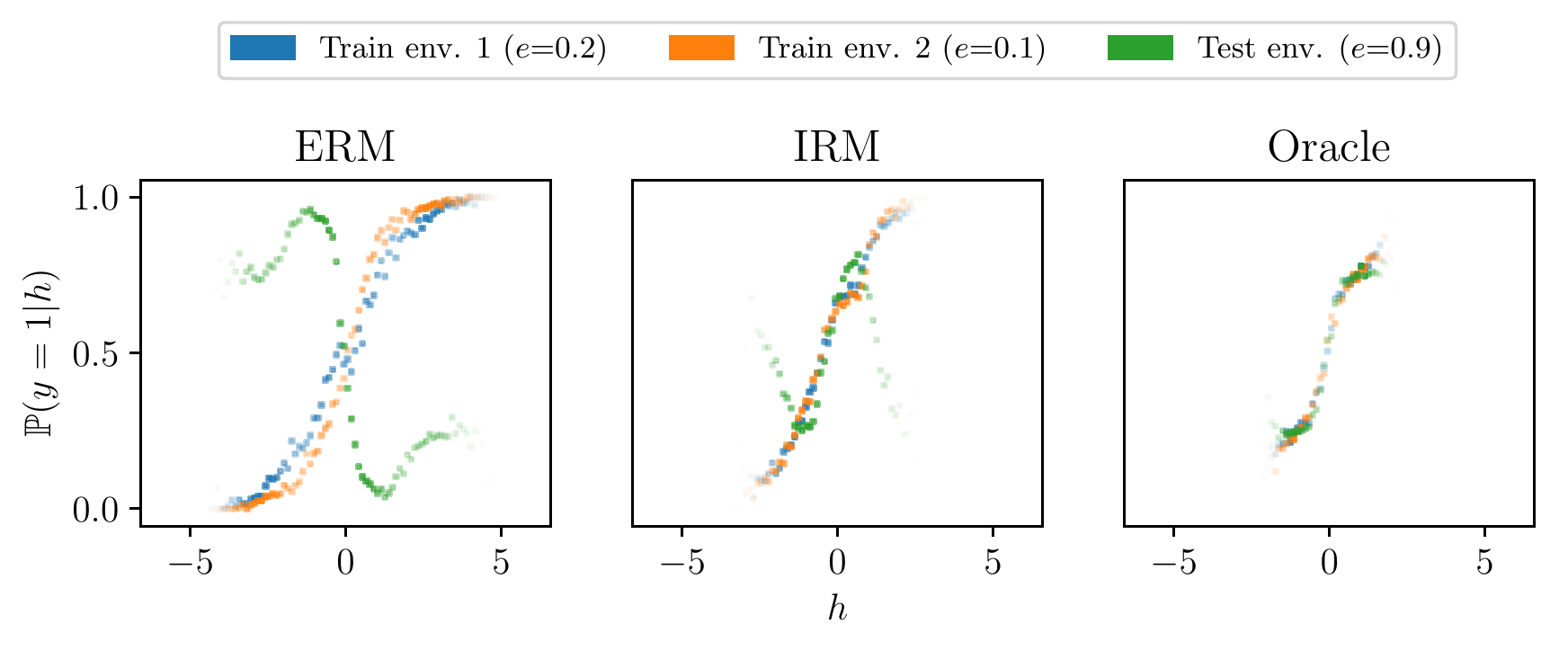}
  \caption{$\PP^e(y=1|\hH^e=\hh)$ as a function of $\hh$ for different models trained on Colored MNIST: (left) an ERM-trained model, (center) an IRM-trained model, and (right) an ERM-trained model which only sees grayscale images and therefore is perfectly invariant by construction. IRM learns approximate invariance from data alone and generalizes well to the test environment.}
  \label{fig:calibration}
\end{figure}

To better understand the behavior of these models, we take advantage of the fact that $\hH = \Phi(X)$ (the logit) is one-dimensional and $Y$ is binary, and plot $\PP^e(Y^e=1|\hH^e=\hh)$ as a function of $\hh$ for each environment and each model in \autoref{fig:calibration}.
We show each algorithm in a separate plot, and each environment in a separate color.
The figure shows that, whether considering only the two training environments or all three environments, the IRM model is closer to achieving invariance than the ERM model.
Notably, the IRM model does not achieve perfect invariance, particularly at the tails of the $\PP(\hh)$.
We suspect this is due to finite sample issues: given the small sample size at the tails, estimating (and hence minimizing) the small differences in $\PP^e(Y^e|\hH^e)$ between training environments can be quite difficult, regardless of the method.

We note that conditional domain adaptation techniques which match $\PP^e(\hH^e|Y^e)$ across environments could in principle solve this task equally well to IRM, which matches
$\PP^e(Y^e|\hH^e)$.
This is because the distribution of the causal features (the digit shapes) and $\PP^e(Y^e)$ both happen to be identical across environments.
However, unlike IRM, conditional domain adaptation will fail if, for example, the distribution of the digits changes across environments.
We prove and discuss this further in \autoref{app:ada}.

Finally, \autoref{fig:calibration} shows that sometimes $\PP^e(Y^e=1|\hH^e=\hh)$ cannot be expressed using a linear optimal classifier $w$.
A method which searches for nonlinear invariances (see \autoref{sub:nonlinw}) might prove useful here.

Now, as with the methods of \autoref{chap:rob} we conclude the chapter with a section highlighting where these ideas are still falling short, and potential avenues on how to fix them.

\section{Looking Forward: Invariance and Generalization}
\label{sec:outlook}
\subsection{Realizable problems}
\label{sub:realizable}
We have discussed how invariance can help us disregard spurious correlations.
However, which combinations of problems and inductive biases lead to our ERM models recklesly exploiting them? 

In common supervised learning datasets, such as the cow on the beach problem,
labels come from multiple human annotators assessing whether there's a cow on the image.
These multiple annotations are often aggregated into a single label by voting.
This process then takes the image $X$, and outputs a label $Y$.
The probability that this procedure takes image $x$ and leads to a possitive label $Y$ is exactly
$\EE[Y^e | X^e = x]$.
If the labeling process stays the same in different environments (regardless of whether the images have the same distribution or not),
then $\EE[Y^e | X^e = x]$ won't change for different environments.
Therefore, if the supports of the environments are shared, $\Phi(x) = x$ leads to an invariant predictor!
We call these problems in which $\EE[Y^e | X^e = x]$ is the same for all $e \in \Eall$, or analogously $\Phi(x) = x$ leads to an invariant predictor,
\emph{realizable}.

This means that as the number of examples and the capacity of our models tends to infinity ERM will 
ignore the biases, recover the right solution and generalize out of distribution.
Why then, do we observe our models picking up on biases in these problems? And can invariance help?
If we hope to use the ideas from \autoref{sec:IRM} and \autoref{sec:causgen} to get rid of these particular biases, several improvements will be needed.

As we mentioned, if $\EE[Y^e | X^e = x]$ stays the same across environments, and i) the supports of $\PP^e(X^e)$ are shared,
ii) the number of datapoints tends to infinity, and iii) the capacity of the model tends to infinity, ERM will recover the
conditional expectation of $Y$ given $X$, and generalize out of distribution.
Thus, if our models pick up on spurious correlations as in \autoref{ex:cob}, one of these three things has to be false.
    \subsubsection{Disjoint environments} \label{subsub:disjoint} If the supports of $\PP^e(X^e)$ are not shared for the different environments,
    then ERM will simply learn the conditional expecation on the training supports, and has a priori no reason to
    generalize to examples out of these supports. Let's see if invariance can help.
    
    Given a data representation $\hH^e = \Phi(X^e)$, we so far have assumed that we can estimate $w^\star_e(\hh) = \EE[Y^e | \hH^e = \hh]$. However,
    $w^\star_e$ is only defined in the support of $\hH^e$. If the supports of $X^e$ are disjoint, then the supports
    of $\hH^e$ can be disjoint for different environments. Thus, it makes a priori no sense to match $w^\star_e$ and $w^\star_{e'}$
    if they are defined on different parts of the space.
   Two potential solutions arise. The first one would be to add a small domain adaptation term forcing $\hH^e$ and $\hH^{e'}$
    to have the same support for different environments. However, if the domain adaptation term has too much weight, this can
    lead to the problems mentioned in \autoref{app:ada}.
    
    Another alternative is to assume that $\EE[Y^e | \hH^e = \hh]$ has a particular
    structure that allows us to estimate it even when $\hH^e$ has very limited support. Indeed, this is what we've
    been doing so far by assuming $\EE[Y^e | \hH^e = \hh]$ is linear on top of the features. If $w^\star_e$ and $w^\star_{e'}$
    are linear functions, then we can estimate them on datapoints that lie on disjoint parts of the space and know what
    values they take on the entire space, and thus match them. This assumption can be violated, as we have seen in the color
    MNIST experiments. Exploring different structural assumptions for the conditional expectations that are reflected
    in practice, and allow for matching them even when estimated in disjoint parts of the space is left for future work.
    We believe the role of compositionality to be crucial in this regard, since compositional assumptions (of which linear
    models are a very particular case) allow for models to be evaluated in different parts of the space than they were estimated on.
  
    \subsubsection{Finite sample and overparameterized models} \label{subsub:findata}
    While data in deep learning problems is abundant, it is certainly not infinite. In fact, a dominant
    characteristic of all deep learning success stories is that models are overparameterized. That is, they 
    have more parameters than examples (typically by an order of magnitude). Indeed, large scale image classifiers
    typically attain zero training error, and exploit the implicit regularization of SGD to generalize in-distribution.
    In these problems, there are many solutions with zero training error, and SGD determines which one to pick, in general
    a low capacity one (in the linear case, one can easily prove that SGD picks up the least-norm solution with 0 training error
    \cite{wilson-marginal2017}). We believe that this is likely the dominant reason why large scale machine learning models
    often pick up on spurious correlations rather than the phenomenon of interest when models are overparameterized. The spurious correlations typically observed 
    require in general much lower capacity than the phenomenon of interest, such as the local texture of grass rather than
    the global shape of a cow \cite{Brendel-ea-texture1, Bruna-ea-texture2, Bruna-ea-texture3}.

    We also believe invariance can help detect these situations, and we illustrate this idea with a last example. Imagine the structural equation model.
    \begin{align*}
      H^e \sim \PP^e \\
      Y^e \leftarrow \beta_H H^e \\
      Z^e \leftarrow \beta_Z(e) Y^e + \epsilon \\
      X^e \leftarrow (H^e, Z^e)
    \end{align*}
    where again $H^e$ and $Z^e$ are unobserved, and we want to predict $Y^e$ from $X^e$.
    As we can see, $Y^e$ is a deterministic invariant function from the `causal' variable $H^e$.
    However, there is an `effect' $Z^e$ which has spurious correlations with $Y^e$.
    If $\epsilon$ is small, then $Z^e$ is strongly (albeit spuriously) correlated with $Y^e$ in each environment.
    The interesting case is when both $\|\beta_H\|$ is large, so the regression from $H^e$ to $Y^e$ requires high capacity, and $\|\beta_Z(e)\|$ is large as well, so the regression from $Z^e$ to $Y^e$ requires \emph{low} capacity (since this regression inverts $\beta_Z(e)$).
    In this cases, under less data than the number of dimensions of $H^e$ (so the problem is overparameterized), there are many solutions with 0 training error.
    SGD will pick among these solutions a regression with 0 training error and minimal norm, and thus has a strong inductive bias towards picking the spurious correlations of $H^e$ to $Y^e$.

    When the bias (the correlation between $Z^e$ and $Y^e$) can be estimated with smaller amounts of data than the high capacity correlation from $H$ to $Y$,
    we can detect that the correlation is spurious, and discard it with invariance. Furthermore, once the unreliable
    variables are discarded, the correlation between $H$ and $Y$ is estimated with a more reasonable (yet obviously imperfect) accuracy,
    and the model generalizes out of distribution in a much better way. In other words, the implicit regularization of SGD
    does a great job at making the overparameterized model generalize in-distribution, but it recklesly exploits the spurious
    correlations which makes it fail at generalizing out-of-distribution. By leveraging invariance, we can drive SGD closer to the right
    solution, and improve out-of-distribution generalization.

    In a sense, realizable problems pick up on biases that could be discarded with invariance when: models are overparameterized (and thus
    there are many zero error solutions),
    there are spurious but strong correlations between parts of the input and the label, and these correlations require lower capacity and data
    to be modeled than the phenomenon of interest. Furthermore, we need that once the nuisance variables are discarded, 
    the phenomenon of interest can be statistically estimated to lower error. This latter fact
    is a fundamental characteristic of high dimensional machine learning problems \cite{meinshausen-highd, wilson-marginal2017, ng98highd}.
    To us, this reflects the characteristics of the large scale machine learning problems that we want to solve. While
    one can already draw parallels between the overparameterized realizable cases and infinite data nonrealizable cases,
    one perhaps can exploit the specific nature of these former problems to (for example) come up with better algorithmic ways of matching
    optimal classifiers on top of features. Furthermore, our theory is in general expressed in terms of conditional expectations, which
    implicitly assume sufficient data per-environment. One could perhaps adapt it to work in a softer sense with empirical distributions.
    
    Since with enough examples these issues are solved, this analysis begs the question: how much data is then enough data
    to not worry about these issues? Is it attainable, or is throwing more data to ERM like throwing more compute at an NP-hard problem,
    when by assuming realistic structure we could transform it into a solvable problem?
    In the example, we need data sizes that increase linearly with the dimension of our problem. However, one could easily
    come up with nonlinear problems where $\PP^e(X^e)$ is multimodal and the modes have a power-law distribution, where we would
    need exponentially many datapoints. We have yet to understand what kind of assumptions we are able to make in practical
    problems of interest that mathematically describe the amount of data required by empirical risk minimization in nonlinear
    realizable problems, and what the (perhaps from exponential to polynomial) reduction of sample complexity is when we exploit invariance.
    We leave all these fascinating research problems for future work.
   
\subsubsection{Finite capacity}
When we have infinite data, but finite capacity (due to modeling constraints or regularization), interesting issues
    arise. If our models are $L^2$ regularized, then they will average all correlations, regardless of the problem 
    being realizable or not. Thus, they will pick up on spurious correlations that can be discarded with invariance. However,
    modeling constraints can create very different phenomena. For example, if $Y^e = Q(H^e) + \epsilon$ with $Q$ a quadratic function, 
    and our model is linear, the only linearly invariant solution is to discard everything. In general, it is likely that invariance has a promise
    of working when our models have enough capacity to capture, at least approximately, the phenomenon of interest. We can see
    hints of this in \autoref{theo:nonlingen}, where it is still necessary that an approximately invariant solution lies in our function class.
    To us, these observations also hint to the fact that out of distribution generalization and invariance
    have a larger chance of success in high-capacity problems, or when we have the right model.
    That being said, the nice property about IRM is that it allows for a smooth trade-off between invariance
    and predictive power, so in cases where model capacity is limited we can decide to attain limited invariance so as to not lower our
    predictive power too much.

In essence, we believe invariance can play a strong role in realizable problems,
especially in the high-dimensional problems where models are overparameterized, for the reasons outlined above.
However, there is much work to be done.

\subsection{Nonrealizable problems}
There are still many problems of interest where $\EE[Y^e | X^e = x]$ changes per environments, such as any
of the classical problems in causation related to genetics and biology where environments come from interventions \cite{igolkina-ea-sem-bio, meinshausen-ea-gene}.
That being said, these are not the only ones.
The difference between realizable and non-realizable cases is related to the difference between supervised
and unsupervised or self-supervised learning, and the difference between causal and anticausal learning \citep{scholkopf2012causal}.
Namely, if the labels don't actually come from an annotation process given the entire input $X$, but for example
by trying to predict parts of the inputs from other parts of the inputs (which is a common and highly succesful technique
in natural language processing \cite{devlin2018bert}), then $\EE[Y^e | X^e = x]$ can easily change for different environments so long as $\PP^e(X^e)$ changes.

In this regime there are many areas in which IRM should still be improved. It is important to mention that 
the disjoint environments and insufficient capacity issues can still be present in nonrealizable problems, but since we've already discussed them we mention in here
other areas for improvement. Of crucial importance is the number of training environments needed to achieve 
invariance with respect to all environments. Our theory considers potentially arbitrary or adversarial spurious correlations,
when it is likely that spurious correlations vary in a structured way, such as the regressors of unreliable variables changing
within a certain subspace, or independently. The work on anchor regression \cite{rot-ea-anchor} examines bounded interventions,
what we are discussing in here is different, and in particular is the assumption of interventions varying in an unbounded way
either in a structured subset of all possible interventions or with a particular type of probability distribution (e.g.
independent interventions). Some of these assumptions could lead to more efficient algorithms. Furthermore,
there are still complicated optimization issues when trying to obtain invariance between the training environments,
particularly in the linear case, which can make learning cumbersome. Finally, as seen in \autoref{sec:color_mnist},
the assumption that $\EE[Y^e | \hH^e = \hh]$ is a linear function of $\hh$ is at times unrealistic, and thus the gradient penalty in \eqref{eq:irm1}
is likely very suboptimal. Both these issues could be solved by studying other ways of matching conditional expectations.
We think there is lots to be gained in this direction.

%

\chapter{Domain-Specific Case Studies}
\label{chap:cases}
In this chapter, we discuss for different applied fields the type of out of distribution problems that appear in practice, and how these fields have dealt with those
    issues in the past. We will not focus on potential solutions, these are all nontrivial tasks that will likely require much research to be attacked. 
Instead, we will try to focus on the particular \emph{structure} that is available in each domain, and what shape the different OOD problems in these fields take.

\section{Robotics}
\label{sec:robotics}

The aspect which characterizes robotics within AI is the study and design of rational agents that act and solve tasks in the \emph{physical world} \cite{russell-ai-2009}.
As such, a large part of robotics deals with the components of the agent that interact with the real world, such as effectors (like legs and joints) and sensors
  (like cameras or eyes).
Strictly speaking, robots do not necessarily require learning, and indeed many tasks in robotics have been solved with 0 learning involved.
However, for tasks which are difficult to hand engineer, it is common for learning to be used as part of a solution.

Learning in robots is particularly hard, and there are many reasons for this. The principal challenge for learning with robots is sample complexity:
while a simulated agent could play millions of games in a day \cite{silver-ea-zgo}, physical robots are bound by real time constraints. For instance,
a robot learning to bat a baseball needs to wait for the ball to actually come, which takes at least a few seconds, implying the agent can act (being very optimistic)
at most a few thousand times a day. While efforts have been developed to train robots in a distributed fashion \cite{yahya-ea-distrob}, it is still the sad reality that
robots that want to leverage learning often have to deal with 4 to 6 orders of magnitude less samples than simulated agents. As such, 
learning robots need to leverage additional structure to be efficient.

One of the main tools to aid in these difficulties is sim-to-real transfer: the process by which a robot learns a policy from a simulated environment
and aims for it to generalize to the real world \cite{chebotar-ea-closing, openai-dext}.
Another way of leveraging data that doesn't come from robot interactions is to learn from observations of another more experienced agent, such as a human \cite{yu-ea-oneshot}.
An able human's behaviour gives supervision to a robot of what good performance looks like, thus providing a strong learning signal.
However, it is not perfectly clear how to optimally learn from said supervision,
  especially when mapping the action space of a human to that of a robot is impossible (e.g. when actions are at a joint level), or only raw visual information of the human is available.

It should be clear to the reader that these are different versions of out of distribution generalization.
Simulations are inherently imperfect and share many differences with the real world.
This is particularly true when robots deal with raw sensory imput like vision.
Furthermore, sensors are noisy and simulations can often be very imprecise, and the real sensor noise can be difficult to model \cite{openai-dext}. Therefore, generalizing
from simulation to reality is out of distribution.
Finally, humans' effectors and sensors are different from those of the robot, a mapping from one to the other is in general not available, and sometimes
  only raw sensory data is available of the human solving the task (such as a video). 

Several domain-specific techniques have been developed to deal with these OOD problems, and we will not attempt a thorough review.
What we will do is highlight some of the most used and successful techniques, to illustrate the type of \emph{structure} that is often available,
  and the shape particular OOD problems in robotics can take.

\subsection{Sim-to-Real}
Our starting point for this brief review is \cite{openai-dext}. There, the authors train a robotic hand to perform dexterity tasks, such as moving a cube
from one position to another. The training is done entirely in simulation, and the robot is then deployed in the real world. The robot decision making process
has two components, a CNN predicting the object pose, and a control policy that takes the predicted pose and actual fingertip locations and outputs actions.
In order for the sim-to-real transfer to succceed, both components have to generalize out of distribution: if the pose prediction is wrong, the policy's inputs are meaningless;
if the policy doesn't adequately transfer its knowledge from the fingertip simulations to the real sensors, its actions can be suboptimal.

Without a priori knowledge of the simulator, none of these two components have any reason to generalize.
If the simulator is perfect, this is an in-distribution problem with higher chances of success.
Hence, one would think the goal of the simulator designers is to handcraft a simulation as close to the reality as possible.
However, this turns out to be surprisingly untrue, at least from a practical standpoint.

Simulations will never be perfect, and trying to approximate the real world with a single simulation is doomed to fail unless we have essentially a perfect model of the problem at hand.
Therefore, what the authors opt to do is to employ \emph{domain randomization}, which simply means that the simulation parameters are randomized across different training
runs. We can understand this with the multiple-environment setup, where $\Etrain$ then consists of the many possible simulations, and $\Eall$ of both real and simulated
environments. Domain randomization then makes sense in the presence of a simulator.
Assuming that there is a policy which solves the task in all environments (i.e. 
we're in the realizable problem\footnote{see \autoref{sub:realizable}}),
$\PP(Y^e | X^e)$ (where $Y^e$ could be the poses from visual sensory inputs $X^e$) is the same or at least very similar across
environments. Thus, using a single simulator $\Ptrain$ would limit learning to the support of $\Ptrain(X)$, while using a family of environments would allow
learning the correct policy for $\cup_{e \in \Etrain} \Supp(\PP^e(X))$, a potentially much larger space. 

Now, an important comment to make is that likely $\cup_{e \in \Etrain} \Supp(\PP^e(X))$ and $\Supp(\Ptest(X))$ are still disjoint, since visual simulations are
inherently different to the real world. Therefore, we can see that any policy that aims for success in this task has to generalize outside of its support. This can be
achieved in two ways: by having a \emph{stable} policy with respect to small changes according to some distance of the $\PP^e(X)$ and trying to make $\PP^e(X)$ close to $\PP^\test(X)$ (see \autoref{sec:wrob}), or by exploiting
some form of compositionality in the representation space on which the policy is operating on. Both these routes are fruitful topics of future work that
could be merged with invariant prediction, as discussed in \autoref{subsub:disjoint}. Looking for invariances across simulated environments
could be an interesting line of research: we don't want to exploit correlations that are specific only to certain simulated parameters or noise distributions,
since these are very unlikely to generalize to the real world; we want to take decisions that are optimal for all simulated environments in the same way.

\subsection{Imitation Learning from Humans}

Given how hard it is for robots to learn with very few trials, a reasonable route is to \emph{teach} them how to do things by \emph{showing} them how to perform certain tasks.
As very eloquently phrased by \cite{reed-ea-npi}: ``While
unsupervised and reinforcement learning play important roles in perception and motor control, other
cognitive abilities are possible thanks to rich supervision and curriculum learning. This is indeed
the reason for sending our children to school.''.

To discuss imitation learning from humans in the context of robotics, we will use \cite{yu-ea-oneshot} as a case study.
There, the authors have access to both robot and human supervision for several training environments $\Etrain$, and exactly one human demonstration from a test task 
inside a family of possible ones $\Eall$.
Different environments correspond to different objects to be manipulated, humans, and scenes.
Denoting $\matX$ as the space of human video demonstrations, and $\Pi$ as the space of robot policies, the goal is to learn a mapping $f: \matX \rightarrow \Pi$ such that
  $f(X)$ is close to optimal for all the tasks or MDPs in $\Eall$.
More concretely, training data consists of pairs $(X, A)$, with $X$ being a human demonstration and $A$ the actions of a close-to-optimal robot demonstration.

The authors of \cite{yu-ea-oneshot} approach this task with a MAML-like \cite{finn-ea-maml} objective.
There, they learn a loss $L$ defining then $f(X) := \argmin_{\pi \in \Pi} L(\pi, X)$, 
  and optimize $L$ so that the actions taken by policy $f(X)$ are close to $A$ for the training pairs. Since actions $A$ are close to optimal robot demonstrations
  for the training environments, effectively optimizing this means that $f(X)$ will be so for these training tasks.
The argmin is approximated in a differentiable way by interpreting SGD as a computational graph, and everything is backpropagated onto the learned loss $L$.

While this is a reasonable approach, this is arguably doing empirical risk minimization on the space of mappings from $\matX$ to $\Pi$, and it is likely that
  as such it has the usual pitfalls outlined in \autoref{sec:erm} and \autoref{app:meta}.
Could we then leverage the ideas of \autoref{chap:CIG} to find demonstration-to-actions decisions that are optimal across the different objects, scenes and demonstrators?
We identify two research questions that would probably need to be addressed to do so.

The first issue is that of defining environments as individual combinations of humans, objects and scenes. Practically, as mentioned in \cite{finn-ea-maml}, only
few pairs $(X^e,A^e)$ are available for the same combination $e$, many times a single one. Having a single datapoint per environment means that the population statistics assumed
in \autoref{sec:IRM} are likely very different to the empirical estimates. We imagine two ways of dealing with this. The simplest way would be to search for invariances across
\emph{groups} of humans, objects and scenes rather than all possible combinations, therefore allowing for slightly weaker invariances but reducing this potential overfitting
  in individual environments. The second way, more complex, would be to leverage the ideas of \autoref{sec:outlook}, particularly those of the realizable setup with
  overparameterized models, since it is reasonable to assume that this is a realizable problem where a common policy could be optimal across the series of tasks. This second
  approach though, we believe will likely require much work before it can be put to scale.

The second issue, perhaps more fundamental, is that of looking for invariant mappings from human demonstrations $\matX$ to the policy space $\Pi$.
Item B in the definition of invariant prediction (see \autoref{obsdef:invpred}) is well defined for arbitrary losses. We could then aim to
build a representation $\Phi: \matX \rightarrow \hmatH$ such that there is a single $w: \hmatH \rightarrow \Pi$ with $w \in \argmin_{\bar w:\hmatH \rightarrow \Pi}
\EE_{(X^e, A^e) \sim \PP^e} \left[\ell\left((\bar w \circ \Phi)(X^e), A^e\right)\right]$ for $e \in \Etrain$,
with $\ell$ being as in \cite{yu-ea-oneshot} the cross-entropy loss between the actions of $f(X^e)$ and the robot demonstrations $A^e$. This however is not trivial.
The main issue is that the mapping from representation to policy space $w$ will likely have to be nonlinear (what even is a linear predictor that outputs policies?). As such,
the loss over $w$ will likely be nonconvex and the gradient penalty in \eqref{eq:irm1} will not enforce that $w$ is in the argmin, but merely a saddle point, of which
there can be many uninteresting ones. Because of this, we suspect that finding invariant policies is somewhat intrinsically dependent on being able to learn
  nonlinear invariances. The discussion of \autoref{sub:nonlinw} might be a useful starting point for this research endeavour.

\section{Computer Vision}
\label{sec:cv}

Since the inception of computer vision, there have been two approaches for solving complicated sensory tasks like object recognition:
  that of specifying a model by hand, and that of learning the concepts from data.
As more data and computing power became available, models started relying more and more on statistical patterns than on human-defined inductive biases,
  arguably with great success.
That being said, learning statistical patterns is not the only role of datasets in computer vision.
Datasets also provide a way to evaluate progress by giving a way to quantitatively measure success in a certain task.

However, success in a dataset is only a proxy for the ability to perform a vision task.
If a model has high accuracy on a given object recognition dataset like imagenet, does that imply that the model knows how to classify objects? Did it learn anything
  about how vision works?
The seminal paper of \cite{torralba2011unbiased} reminds us that proxies can often be imperfect, especially when one spends a lot of time trying to beat the proxy,
  and not the background problem the proxy is meant to represent.
In this work, the authors start with a fascinating human experiment called ``Name That Dataset''.
The authors argue that if computer vision datasets were faithful complete proxies for the real visual world, then one wouldn't be able to easily distinguish which dataset individual pictures come from. However, both humans and machine learning algorithms trained to do so can easily name the dataset each picture belongs to.

The authors go one step forward, and show the perhaps unsurprising fact that even when the task is the same (saying whether there's a car in the picture), the generalization
  of models trained in one dataset and tested on another one is poor. This last fact clearly implies that the model didn't learn what is a car, but how to say whether
  there's a car \emph{in the particular dataset it was trained on}.

Let's examine these two issues more concretely. The ``Name That Dataset'' experiment tells us that humans can, given an image, say what dataset it came from.
Considering the distributions of different datasets as environments in a family $\Eall$ of possible CV datasets for a given task,
  this is proof of existence of a function $f: \matX \rightarrow \Eall$ such that if $X^e \sim \PP^e(X)$, then $f(X^e) = e$ with probability close to 1.
If $f(X^e) = e$ with probability 1 for $X^e \sim \PP^e(X)$, this implies that the supports of the different environment distributions are disjoint, i.e. $\Supp(\PP^e(X)) \cap 
\Supp(\PP^{e'}(X))) = \emptyset$ for all $e, e' \in \Eall$. This helps understand why there can be a high generalization error across datasets (i.e. a large $R^\OOD$).
Without strong priors, models trained on a given dataset $e$ can take arbitrary values outside of $\Supp(\PP^e(X))$, and hence have a priori no reason to generalize out of distribution.

Years after this, with the deep learning revolution and the advent of adversarial domain adaptation (ADA)\cite{ganin2016domain},
	many authors have tried to address this issue with ADA approaches \cite{tzeng2017AdversarialDD, long-ea-ada}.
The justification seems natural: training a neural network discriminator to recognize the dataset images come from and optimizing the representation so that this 
  discriminator has high error is in a way making the representation not susceptible to the ``Name That Dataset'' phenomenon. It is literally a loss making it so that
  the dataset cannot be named from the features.

For the reasons outlined in \autoref{app:ada} we believe these approaches are misguided, but for completeness we include a short example in here.
We refer the reader to the appendix for more details on this issue.
While ADA techniques are increasingly being presented as catch-all solutions to generalization across domains, let us imagine a simple task.
Imagine environment $e=1$ is MNIST, and environment $e=2$ is MNIST with a different class balance, where $90\%$ of the images are of 9's. Then,
for a representation $\hH^e = \Phi(X^e)$ to satisfy $\PP^{e=1}(\hH^e) = \PP^{e=2}(\hH^e)$ it will have to map most images of all digits to images of 9's (or vice-versa) to match
the marginal feature distributions. Trivially we can see that any such domain-invariant representation will have at most $20\%$ generalization accuracy in this problem,
barely more than chance. This is in a problem where simply training in $e=1$ and testing in both environments would suffice.

With a bit of a deeper look, it also becomes clear that being able to name that dataset with some success is not necessarily a problem for generalization. Shifting $\PP^e(X)$ distributions
without making the supports disjoint doesn't a priori imply any increase in generalization error. In particular, if a predictor has low error across the support of the
training distribution, and the testing distribution simply shifts the probability assignments of $\PP^e(X)$, we will be able to name that dataset with high accuracy but the predictor will generalize perfectly.
It is easy to prove that the best possible accuracy for a ``Name That Dataset'' discriminator for two environments $e=1, e=2$ (i.e.
  $\max_{f: \matX \rightarrow \{0, 1\}} \EE_{X \sim \PP^{e=1}(X)}[f(X)] - \EE_{X \sim \PP^{e=2}(X)}[f(X)]$) is exactly the total variation distance $TV(\PP^{e=1}(X), \PP^{e=2}(X))$. As such, the ``Name That Dataset'' game is in a way extreme: a total variation distance of 1 (perfect 
domain discrimination) means that supports are
  disjoint and hence predictors have no a priori incentive to generalize; however, a total variation distance less than 1 is compatible with shifting only frequencies
  of $\PP^e(X)$ and perfect generalization.

In the case where the total variation distance $TV(\PP^e(X), \PP^{e'}(X))$ is less than 1 and environments overlap, the OOD error of a predictor with low error across a training
environment is not due to shifting $\PP^e(X)$, but has to be due from shifting conditionals between the features it's using and the target (at least when sufficient data is
available). Another way of saying this is that when environments overlap, dataset biases have to be from shifting $\PP^e(Y | \hH)$. These are spurious correlations that
invariance can detect and eliminate.

However, what can we do if environments are effectively disjoint (as is the case when the dataset is perfectly nameable)? Without assumptions on the data, there is nothing
we can do. However, data has structure, and indeed ADA has enjoyed limited success. As mentioned in \autoref{sub:realizable}, there are two things we could do in this case.
We can try to use an ADA term with a small contribution to the overall loss, and apply invariance on top of those features, though if the ADA contribution is too high
this brings about the problems of \autoref{app:ada}. Finally, we could make assumptions in the data that allow us to estimate conditionals $\PP^e(Y | \hH)$ on one
part of the space $\hmatH$, and evaluate it reliably in other parts of $\hmatH$. We effectively did this (solving in this way simple potentially
disjoint environment problems like \autoref{ex:example} if the distributions of $\PP^e(X_1)$ are set to be disjoint) by assuming a linear conditional expectation
$\hh \mapsto \EE^e[Y | \hH=\hh]$ (i.e. a linear invariance $w: \hmatH \rightarrow Y$). However, we could attempt to do this for more complicated nonlinear models
by learning compositional invariances, since compositional assumptions allow models to be trained in one part of the input space and evaluated on another.
For instance, compositionality-inspired architectures for visual tasks like relational networks \cite{santoro-ea-relational} are to us good candidates for compositional
  invariances $w$. However, before we can leverage these compositional nonlinear invariances, we have to design penalties that search for nonlinear invariances, which 
  was discussed in \autoref{sub:nonlinw}.
This brings us to the next chapter, in which we study a bit more precisely the role of compositionality in language, and its relationship with out of distribution generalization.

\section{Natural Language Understanding}
\label{sec:nlp}
Humans are surprisingly efficient learners when it comes to language.
Given a new word, it is sufficient for us to hear it only once in the right context to understand how to use it.
For example, if I say ``A blep is a fruit that looks like an orange and tastes like an apple'', then we would know what the sentence ``Yesterday I ate a blep pie''
means, and we could even imagine what such a pie would taste like.
Furthermore, we would also know that the sentence ``Yesterday I was wearing a blep instead of a shirt'' wouldn't make much sense.

The skill underlying this extremely fast learning has been denoted as \emph{systematic compositionality}: the ability to produce an inifinite number
  of syntactical structures by combining different components \cite{Chomsky57a}.
The ability to generalize what has been learned from these components to the larger structures (as in the above
  paragraph) is known as \emph{systematic generalization} \cite{lake-ea-systematic, bahdanau2018systematic}.
 
As has been empirically demonstrated recently \cite{lake-ea-systematic, bastings-etal-2018-jump, weber-etal-2018-fine} standard sequence to sequence models \cite{sutskever-ea-seq2seq, bahdanau-ea-attention} often fail to exhibit systematic generalization.
In particular, \cite{lake-ea-systematic} define a task in which $\matX$ corresponds to sequences of commands that ought to be translated to sequences of actions
$\matY$. They observe that when the training set consists of randomly sampled pairs $(x, y)$, and the test set is i.i.d. from the same distribution, then seq2seq models
generalize perfectly. However, when they test on longer sequences than those seen at training time (a task that can be solved perfectly with
systematic compositionality), the models fail dramatically. Furthermore, the authors perform a similar experiment to the ``blep pie'' mentioned earlier, in which
the training set consists of all types of command sequences (such as ``turn twice'') but only one utterance of the primitive command ``jump''. If the model understood
the compositional structure of the problem, it would understand that ``jump'' is a primitive command and ``twice'' a modifier, both of which it has seen
how they work. Therefore, it should understand new commands constructed from these components, and be able to translate ``jump twice'' into actions. However, all
models fail terribly at this task.

It is clear that these last two examples are particular out of distribution problems. One is where $\Etrain$ contains commands up to a certain length and $\Eall$ of arbitrary lengths; the second one where $\Etrain$ contains commands using all modifiers and primitive commands, but only one instance of a particular primitive command, while $\Eall$ contains
  all \emph{combinations}.
The work of \cite{bahdanau2018systematic} makes this idea a lot more concrete, and studies what is required to solve some of these issues, as we now detail.

How much inductive bias is it needed to solve systematic compositionality tasks?
It is clear that by hand-specifying a solution we can solve the tasks of \cite{lake-ea-systematic} very easily. However, this is unlikely to work in more complex
tasks like machine translation. The work of \cite{bahdanau2018systematic} studies this question in a lot more detail.
They create a synthetic task in which the input is an image containing two symbols, and a question asking about the relationship between these symbols (e.g. ``Is `X' above `Y'?''). 
The output is a yes or no answer to this question. The authors define the precise training and testing distributions in such a way that a particular instance of
systematic compositionality is largely necessary and sufficient to solve this task. For each triple $(X, R, Y)$ of symbols $X$, relationship $R$ and symbol $Y$,
there is a distribution $\PP^{(X, R, Y)}$ of images for which the relationship $(X, R, Y)$ is manifested. The authors then construct training environments $\Etrain$
containing a subset of these $\PP^{(X, R, Y)}$ (with a million images per training environment), and test on $\Eall$ containing \emph{all} possible $(X, R, Y)$ combinations.
The subset $\Etrain$ is such that all $(X, R)$ pairs appear, but only a very limited number of $(X, R, Y)$ triples do so. A model then will be able to generalize
to all triples if and only if it learned to appropriately
combine the different relations and symbols with a few training environments that contain all components but few combinations (a
form of systematic generalization).

The findings of \cite{bahdanau2018systematic} are quite illuminating. In their first line of experiments, they assess whether \emph{generic} methods like attention models with
little task inductive bias generalize systematically to $\Eall$. In line with \cite{lake-ea-systematic}, they show that these models fail to generalize.
However, \cite{bahdanau2018systematic} goes two steps forward. In addition, they show that architectures with a small task-specific inductive bias like Neural Module Networks (NMNs \cite{andreas-ea-nmn})
with a tree-structure generalize almost perfectly. Finally, they ask the question of whether this tree structure can also be learned from the task itself,
reducing the need of any task-specific inductive bias. The answer to this final question is largely no: published algorithms which attempt to learn this structure
by empirical risk minimization fail to meta-learn the tree structure and generalize to $\Eall$. This is in line with \autoref{app:meta}: the role of a
successful meta-prior in here comes in the form of a tree-structure which limits how environments can relate to each other, while employing a flat meta-prior inhibits out
of distribution generalization from a limited number of training environments.

We identify two follow up questions in this regard. The first one is whether an invariance inductive bias could be used. All experiments above were performed with empirical
risk minimization, and hence the meta-prior with the tree structure came from hand-specifying the NMN layout. We are certainly not arguing that no inductive bias should be present,
we simply wonder if the particular tree structure could be discovered by looking at structures that have invariant prediction across the training environments $\PP^{(X,R,Y}) \in \Etrain$.
Secondly, it is important to note that $\Etrain$ are disjoint environments: there is no overlap between the images for different triples $(X, Y ,R)$
(see also \autoref{sub:realizable}).
Thus, the systematic generalization of NMNs with tree layouts imply that some architectures with compositional
inductive biases can generalize outside of its training support. We did this in \eqref{eq:irm1} with a $w: \hmatH \rightarrow \matY$
which has to generalize outside of its support because it is linear, but this is obviously limited.
Along these lines we find the results of \cite{bahdanau2018systematic} then particularly promising for attacking disjoint environment problems.

As a final note, we would like to mention the work of \cite{radford2019language}, in which the authors train a massive-scale language model on 40GB's of text
  from links shared on Reddit. The samples generated by the model are remarkably similar to those of humans, and this work has often been discussed
  on the performance improvements that can come essentially just from scale. The authors also use this model in several zero-shot generalization
ways for other language tasks. However, to us their results are evidence that scale is not enough. The authors succesfully obtain remarkable language modelling performance,
even when generalizing to other datasets. However, their results in terms of text sumarization, question answering and machine translation are far from good. In particular,
the question answering answers that come from the model often exploit spurious cues, like answering `who' questions by mentioning the name of any character in the text.
These cues are evidence of a lack of systematic generalization, and would fail to generalize out of distribution. This is \emph{even} in the presence of an unprecedented scale,
with most laboratories in the world simply being unable to train such large models. Our opinion is that this paper helps clarify that scale is not enough, and in line with
\cite{bahdanau2018systematic}, highlighting the importance of smart inductive biases for generalizing with systematicity and out of distribution.

\section{Fairness in Machine Learning}
\label{sec:fairness}
Automatic decision making, powered in part by machine learning, is becoming an increasingly large part of our daily lives.
As such, it's becoming increasingly important to be able to assess that these decisions are not discriminatory against certain sectors of society.
This is not only important from an ethical standpoint, but current and future regulations such as GDPR \cite{gdpr} are starting to make this mandatory from a legal standpoint.
Therefore, machine learning practitioners are going to have to start worrying about their classifiers and predictors being fair whether they want it or not.
The immediate question that follows is: what is a fair classifier?

\subsection{Oblivious and Causal Criteria for Fairness}
It is clear that we need a way to quantify fairness, or at the very least to be able to objectively assess whether a machine learning system is fair or not.
The following are a few of the relevant attempts at answering this question, with the seminal work of \cite{hardt-ea-equality} as our starting point.
However, an important note in this regard is that at the moment of writing, no universally accepted fairness criterion is available.

Let $X$ be the features available of an individual, and $Y \in \{0, 1\}$ the target we are trying to predict, such as ``would individual $X$ repay their loan?''.
As well, $A \in \matA$ is a variable (which for simplicity we assume discrete) determining whether $A$ belongs or not to a certain protected group.
Different works assume different amount of knowledge over $A$ for each example.
For now we shall assume $A$ is known, and hence datapoints take the form $(x^{(i)}, y^{(i)}, a^{(i)})$ for $i = 1, \dots, n$.
A predictor $f: \matX \times \matA \rightarrow \hmatY$ has the objective of predicting $Y$ as best as possible without being discriminatory, and these are several ways
of defining what that could mean.
Assuming $\hat Y = f(X, A) \in \{0, 1\}$ determines the outcome of an action, such as ``is $X$ going to be accepted for a loan?'', then $f$ can satisfy one or both of the following
notions:
\begin{enumerate}
  \item A predictor $f$ is said to provide equality of opportunity \cite{hardt-ea-equality} if $\hat Y = f(X, A)$ satisfies
    $$ \PP(\hat Y = 1 | Y = 1, A = a) = \PP(\hat Y = 1 | Y = 1, A = a') \quad \text{for all $a, a' \in \matA$} $$
      This is akin to saying ``If $X$ is acceptable for a loan (i.e. ends up satisfying $Y = 1$), then on average they
      will be considered for a loan with the same likelihood regardless of whether they belong to the group $A$ or not''.
  \item A predictor $f$ is said to provide equality of odds if $\hat Y = f(X, A)$ satisfies
    $$ \PP(\hat Y = 1 | Y = y, A = a) = \PP(\hat Y = 1 | Y = y, A = a') \quad \text{for all $y \in \{0, 1\}, a, a' \in \matA$} $$
      This is trivially stronger than equality of opportunity, with the added element that bad clients will also be on average
      considered in the same way by the predictor regardless of their membership to the protected group.
\end{enumerate}
One of the problems that both these notions have is that they are \emph{oblivious}, which means that they depend only on the joint distribution of $(X, A, Y)$ and $f$.
While this doesn't seem a priori unreasonable, \cite{hardt-ea-equality, kil-ea-fair}
  show with a simple example that any oblivious measure cannot distinguish between
  two case studies with the same joint distribution of $(X, A, Y)$ and $f$ but that humans consider discriminatory in one case, and not the other.

More concretely, \cite{hardt-ea-equality} construct two causal graphs for $(X, A, Y)$ which have the same joint distribution, but intuitively very different interpretations.
In the first one, $A$ causes $X$ which causes $Y$, and in the second one $A$ causes $Y$ and both of them cause $X$.
The difference in whether $X$ is cause or effect of $Y$ intuitively tells us whether we should use it or not to predict $Y$.
Intuitively, if $X$ is a cause of $Y$, such as the wealth, then it would make sense to use it to predict, but if it is a consequence of $A,Y$, then using it to predict
  would introduce profiling based on $A$ that doesn't provide causal information on $Y$ (a behaviour judged intuitively as discriminatory).
However, the analysis of \cite{hardt-ea-equality} stops there, without being able to provide an answer as to why or how behaviours on these two graphs should be considered
  as different, aside from the intuition that one seems discriminatory and the other is not.

\cite{kil-ea-fair} goes one step forward. More concretely, they assume a causal graph for variables $(A, Z, H)$, of which $X = (Z, H)$. The variables
$H$ (specified by the user of the framework) are said to be \emph{resolving}, which means that using them to predict $Y$ is not judged as discriminatory.
In the example provided $Y$ is acceptance into a program, and $H$ is the choice of program, of which there are ones that are more dificult to get in than others.
Then, the authors say that a predictor $\hat Y = f(X, A)$ has unresolved discrimination if there is a path from $A$ to $\hat Y$ that is not blocked by a variable in $H$.
This has a natural interpretation: if the predictor $f$ has unresolved discrimination, then it is using $A$ to predict $Y$ in a way that is not determined acceptable.
Since the notion of unresolved discrimination depends on the causal structure of the data, and not just the joint distribution, then it is not an oblivious property, which
  is why it is able to distinguish between the two different scenarios of \cite{hardt-ea-equality}.

The notion of unresolved discrimination though is problematic to implement, since a priori it requires specifying the causal nature of the data, which is in general not accessible.
Furthermore, determining whether an unresolved path from $A$ to $\hat Y$ exists is in general not feasible since it would require intervening on $A$ to do so, 
  something that is in general impossible.
Therefore, the authors of \cite{kil-ea-fair} take the `benevolent viewpoint' and seek only to restrict the effects that proxies $P$ of $A$(which can be intervened on) take on
  $\hat Y$. This means, enforcing $\PP(\hat Y | \do(P=p)) = \PP(\hat Y | \do(P=p'))$.
While this can be at times enforced, it still requires being able to intervene on proxies, and it is unclear that the benevolent viewpoint is a reasonable way of
  specifying whether a classifier is fair, or how close it is to the original notion of not having unresolved discrimination.
  However, these ideas help us shed a light into what is doable, at least in theory, when using non-oblivious criteria for fairness.

\subsection{Invariance, Fairness, and Out of distribution Generalization}
Taking a step back, we want to understand what are examples of bad behaviour, before trying to theoretically define fairness.
In 2015, a flaw in Google's image recognition system was made very public when users spotted that their photos which contained black people were
  being wrongly tagged as `Gorilla' instead of `People' \cite{guynn-2015,mulshine-2015}.
This is a clear example of a spurious correlation, between dark skin texture and a gorilla being present in the picture.
While in many photos this correlation is strong (such as photos of jungles), in certain users' photos, it was completely off.
A particular concern is that this error exists solely on humans with dark skin, since the correlation is never present for white skin individuals.
Therefore, what makes this issue more problematic is the fact that the classifier performs poorly only on a protected group of individuals.

In line with recent work \cite{hashimoto-ea-dro}, this points to a definition of fairness involving low errors across all possible demographics and protected groups.
Spurious correlations which produce out of distribution errors to certain protected groups would then violate this notion of fairness.
To put it more concretely, if we assume that there is an unknown function $f^*(x)$ mapping an image $x$ to the labels,
and $\mathcal{Q}$ is a family of distributions over pairs $(x, a)$, then we could define the possible test environments $\Eall$ as 
the space of the distributions of $(Y, X^{\PP,a})$ with $Y = f^*(X^{\PP,a})$, and $X^{\PP,a} \sim \PP(X | A = a)$ with $\PP \in \mathcal{Q}$.
Comparing predictors based on $R^\OOD$ across this family of environments means judging them by their worst out of distribution error across protected groups.
Hence, in order for the predictor to perform well, it has to perform well for all protected groups.

Now, a central question is why are predictors picking up on these spurious and discriminatory biases?
The work of \cite{hashimoto-ea-dro} provides evidence to support a simple explanation: in many datasets (especially those collected in the first world),
  some protected groups have a much smaller number of examples, and hence contribute less to the overall loss.
In the absense of infinite data, as argued in \autoref{sub:realizable}, overparameterized models trained with SGD tend to pick up on spurious correlations and data biases.
This is worse for protected groups with less data: the less data available the stronger the effect of SGD's inductive bias, which favours low capacity spurious correlations.
Along these lines, looking for invariances across environments and protected groups could help reduce some of these issues, by avoiding spurious correlations
  that generalize poorly for some protected groups. Importantly, this kind of invariance would not be an oblivious property, since it would depend on
  the particular environments that the user or society specify. These environments could be different data splits according to temporal or geographical locations
  across which we would like to generalize, and across which spurious correlations could fluctuate and thus be detected.
We then think that the ideas of \autoref{sub:realizable} could help ameliorate some of the issues of building classifiers that work robustly for all protected groups.

\chapter{Looking Forward}
\label{chap:forward}
In \autoref{sec:outlook} we have discussed what the main downsides and next steps are in the theory and praxis of invariance as a tool for out of distribution generalization.
In this chapter, we take a stab at other potentially fruitful avenues of research that stem from ideas in this book.
In particular, we will be concerned with the need of \emph{agents} to generalize out of distribution.
While we phrase the goal as reducing the sample complexity of agents learning from their environment, it will be clear that these ideas apply to agents attempting to generalize
to new environments or tasks.
A particularly exciting new topic will be the relationship between uncertainty and out of distribution generalization: saying which actions are the most likely
to provide information about how to generalize out of distribution, and how to use this information to drive exploration.

\section{Reinforcement Learning and Out of Distribution Generalization}
\label{sec:rl}
Here we briefly discuss how out of distribution generalization, and some of the exposed tools, could help reduce the sample complexity of reinforcement learning.
Using standard notation, let's assume a task is given by an episodic Markov Decission Process or MDP $(\matS, \matA, P, R)$ where $\matS$ is our state space,
$\matA$ is the action space, $P: \matS \times \matA \rightarrow \mathcal{P}(\matS)$ is the transition probability function, and $R: \mathcal{S} \rightarrow [0, 1]$ is the
reward function. For simplicity, we assume rewards are state-dependent, deterministic
and an episodic setting where the initial state is $s_0 \in \mathcal{S}$, and the maximal time-step or time horizon is $\Tmax \in \NN$,
though extensions of these ideas to other common settings are straightforward.

For now, let us begin with the policy gradient setup. Here, $\pi_\theta: \matS \rightarrow \mathcal{P}(\matA)$ is a policy parameterized by $\theta \in \RR^q$.
In order to make improvements towards an optimal policy, it is customary to make gradient steps attempting to maximize the cumulative return function
$$ J(\pi_\theta) = \sum_{t=0}^\Tmax \EE[R(S_t) | S_0 = s_0, A_k \sim \pi(S_k)]$$
The policy gradient theorem \cite{sutton-ea-pg} tells us that the gradient of $J(\pi_\theta)$ is exactly
\begin{equation} \label{eq:pg}
  \nabla_\theta J(\pi_\theta) = \sum_{t=0}^\Tmax \EE\left[Q^{\pi_\theta,t}(S_t, A_t) \nabla_\theta \log \pi_\theta ( A_t | S_t) | S_0 = s_0, A_k \sim \pi_\theta(S_k)\right] 
\end{equation}
Where the Q function, defined as 
\begin{equation} \label{eq:q}
  Q^{\pi,t}(s, a) = \EE\left[\sum_{t'=t}^\Tmax R(S_{t'}) | S_t = s, A_t = a, A_k \sim \pi(S_k) \text{ for all $k > t$}\right]
\end{equation}is the expected cumulative reward
of policy $\pi$ after time $t$ when being at state $s$ and taking action $a$ at the beginning.

Normally, \eqref{eq:pg} is estimated in one of two ways. The simplest one, by replacing both expectations in \eqref{eq:pg} and \eqref{eq:q} by monte-carlo returns.
The problem with this is that the variance created by the expectation in \eqref{eq:q} is enough to make learning extremely slow \cite{kakade-exploration}.
This is due to the fact that \emph{each call} to $Q^{\pi,t}$ involves an expectation over all the possible \emph{sequences} of actions taken by $\pi$, of which there
are exponentially many. Therefore, a common alternative to the monte-carlo return procedure is to do actor-critic learning, where we try to approximate
$Q^{\pi, t}$ with a parametric function $Q^{\psi,t}$, typically a neural network of some kind (called the critic). Unless the critic is linear and optimally trained,
this injects bias into the procedure, albeit greatly reducing variance. The quality of the approximation $Q^{\pi, t} \approx Q^{\psi,t}$ typically determines the speed
of learning.

Now, the gain from actor critic typically comes in leveraging data from past iterations of the policy $\pi_\theta$ (the actor), as opposed to only using monte carlo
estimates from the current policy. However, this is not easy.
Let us denote by $\PP^{\pi, t, t', s, a}(S)$ the distribution of $S_{t'}$ when starting at $S_t = s$, taking $A_t = a$, and $A_k \sim \pi(S_k)$ for $k > t$. Then,
by definition $Q^{\pi, t}(s, a) = \sum_{t'=t}^\Tmax \EE^{\pi, t, t', s, a}[R(S_{t'})]$.
Imagine we have access to data from a previous policy $\pi$, and we want to estimate $Q^{\pi',t}$. This would mean we have access to data from the regression
problems $X = S_t, Y = R(S_{t'})$, with $S_{t'}$ sampled from $\PP^{\pi, t, t', s, a}$. In order to estimate $Q^{\pi',t}(s, a)$ we would need to solve the regression
problems $X = S_t, Y = R(S_{t'})$, but with $S_{t'}$ sampled according to $\PP^{\pi', t, t', s, a}$ (i.e. changing the distribution of the future states when acting
with $\pi'$ rather than $\pi$).

But this is an out of distribution generalization problem! Namely: we have data for regression problems with distributions $\PP^{\pi, t, t', s, a} \in \Etrain$,
and we want to solve the same regression problem for other policies $\pi'$ resulting in $\PP^{\pi', t, t', s, a} \in \Eall$.
Making no assumptions on how these distributions relate, the best estimates for $Q^{\pi',t}$ we can get come from importance sampling. But the particular MDP and
how the policies relate provide an enormous amount of structure, and thus implies a strong relationship between $\Etrain$ and $\Eall$.

On how to set-up this problem, there are two avenues. One way to do so
would be to directly consider the out of distribution problem of $X = S_t, Y = R(S_{t'})$ of predicting future returns. The optimal predictor for this task
would be $\EE^{\pi, t, t', s, a}[R(S_{t'})]$, and summing these terms for $t' = t, \dots, \Tmax$ we would obtain the desired Q values. This is what we call the \emph{model-free} out
of distribution approach, in which we try to make a robust predictor for expected \emph{future rewards} that generalizes out of distribution.
Another way to phrase this task would be to instead consider the prediction task of $X = S_t, Y = S_{t'}$, that is directly predicting future states.
Given access to one or many previous policies $\pi$, we would like to leverage the knowledge of this prediction task (which takes the shape of datapoints
for distributions $\PP^{\pi, t, t', s, a}(S_{t'})$) to estimate future states from policy $\pi'$, i.e. $\PP^{\pi', t, t', s, a}$. If we can accurately sample future
states $S_{t'}$ according to $\PP^{\pi', t, t', s, a}$, and the reward function is known, we could simply calculate $Q^{\pi',t}$ by calculating and summing
$\EE^{\pi', t, t', s, a}[R(S_{t'})]$.
We call this the \emph{model-based} approach, where we search for a predictor of \emph{future states} that generalizes out of distribution to the resulting $\PP^{\pi', t, t', s, a}$,
and from these states calculating the cumulative returns.

An important difference between these two approaches is that the optimal regression for predicting future rewards gives us the future expected rewards, and by summing them
for different times we obtain $Q^{\pi', t}$. However, the optimal regression for predicting states would be $\EE^{\pi', t, t', s, a}[S_{t'}]$, which
averages states in whatever representation space the states are, and is a priori unrelated to the expected return over states $\EE^{\pi', t, t', s, a}[R(S_{t'})]$. In order
to leverage a predictor of future states to estimate Q functions, we would need the predictor to provide samples from the entire \emph{distribution} of future states rather
than just the expected future state. For this to be doable with the ideas of \autoref{chap:CIG} we would need to adapt them for working not just with regression
or classification problems where we want the expected outcome, but for predicting entire conditional distributions. We leave these interesting ideas for future work.

Again, we could consider training an invariant $Q$ estimate by leveraging access from several $\PP^{\pi, t, t', s, a}$ coming from multiple past policies $\pi$, and initial states
and actions $s, a$. Another approach would be to learn a Q-function that's invariant over states: finding state features such that for a single policy, the expected future reward is invariant across groups of states. This is inspired from the fact that we would like a single agent to estimate a Q function when reaching a new state that in a representation space
obeys the same rules as previous states (such as changing a new room in Montezuma's Revenge  where only the background color changes).
We believe understanding the sample complexity of agents from invariant Q-functions to be a fruitful topic of future work.
A word of caution is that a priori $\PP^{\pi, t, t', s, a}(S)$ and $\PP^{\pi', t, t', s, a}(S)$ can be disjoint (or at least not completely overlapping), since
one agent might only explore a part of the state space, and a previous iteration another part. These issues have been briefly discussed in \autoref{subsub:disjoint},
we encourage the reader to take a look at it before embarking on this research topic.

We have developed this section based on the policy gradient theorem.
However, the out of distribution component needed to make policy improvements was the Q-function $Q^{\pi,t}$.
If we are able to make Q-functions that generalize better (as is the dream of function approximation), we can use them as well for generalized policy improvement schemes like policy iteration or SARSA.

With these basic ideas in mind, we now discuss how we believe they could set the basis for the much harder (and AI-relevant) problem of exploration.

\section{Exploration, Uncertainty, and Out of Distribution Generalization}
\label{sec:exploration}
A central topic in AI is not how to learn quickly from rewards, but how to efficiently explore the world looking for rewards in the absence of them.
Most reinforcement learning agents opt for naive exploration strategies, where in the absence of reward they revert to random exploration.
However, if we think of how humans explore, we try one thing, and if it doesn't work we try a different thing, and so on, rather than simply moving at random.
This difference of acting random and remembering what we've tried and trying new things is central to solving exploration problems in polynomial time rather than exponential.

Consider an MDP with state space $\matS = \{0, \dots, n\}$, with two actions $\matA = \{0, 1\}$, where the initial state is $s_0 = 0$, and
the only reward is at the end $s=n$. The transitions are ruthless: at each state $k$, one action (not necessarily the same at each state) moves it forward to the next state $k+1$,
while the other one sends it back to the beginning, resetting the episode.
An agent that acts randomly in the absense of reward will take an expected number of episodes on the order of $2^n$ to see any reward,
since only one sequence of actions is satisfactory.
An agent that keeps a simple count of tried transitions and gives itself a reward for trying new state action pairs will need $n$ episodes to reach the final step.
This is the basis of the provably optimal and near optimal exploration methods \cite{kakade-exploration, azar-ea-minimax}, which simply assign various intrinsic rewards
based on the visit counts to state-action pairs and the observed rewards.

A commmon way to figure out the mathematical shape of intrinsic rewards is from the point of view of uncertainty.
Visit counts are but a proxy for how uncertain we are about expected returns.
In the simplest case of bandits, the expected reward $R(a)$ of taking arm $a$ lies with high probability in the range $[\hat r(a) - c\frac{\log k}{\sqrt{N_k(a)}},
\hat r(a) + c\frac{\log k}{\sqrt{N_k(a)}}]$ where $\hat r(a)$ is the average of empirical rewards from trials of arm $a$, 
$N_k(a)$ is the number of times the agent's tried arm $a$ after $k$ overall trials, and $c$ is a constant.
This comes from applying Hoeffding's inequality and simple union bounds over $a$ and $k$ to make sure that it holds with high probability uniformly
over arms and time-steps. This simple but important fact trivially implies that $\argmax_a R(a) = \argmax_a \hat r(a) + c \frac{\log k}{\sqrt{N_k(a)}}$
for all but $\mathcal{O}(\log k)$ trials \cite{bubeck-ea-bandits}.
In particular, it means that most of the times we're selecting the best possible arm when taking actions that maximize
the empirical \emph{extrinsic} reward $\hat r(a)$ plus an \emph{intrinsic} reward $\frac{\log k}{\sqrt{N_k(a)}}$.

This idea generalizes almost verbatim to the general tabular MDP case. Assuming an episodic MDP with time-horizon $H$, $S$ states and $A$ actions,
an agent that acts by maximizing
$\hat r(s, a) + c H \frac{\log\left(SAT/\delta\right)}{\sqrt{N_k(s, a)}}$
(where $T = Hk$ is the number of time-steps and $k$ the number of episodes the agent has traversed)
obtains a regret bounded by $\tilde{\mathcal{O}}(H\sqrt{SAT})$ with probability at least $1 - \delta$,
close to the provably optimal $\tilde{\mathcal{O}}(\sqrt{HSAT})$ \cite{azar-ea-minimax} \footnote{We recall that $f \in \tilde{\mathcal{O}}(g)$ means that $f \in \mathcal{O}(g \log g)$, so optimal $\tilde{O}$ means optimal up to $\log$ factors.}.
The proofs are more involved, but the underlying idea is the same: apply concentration inequalities like Hoeffding's and Bernstein's (or their martingale versions) to
prove the optimal $Q^*$ is bounded above by the Q-function for total rewards 
$\hat r(s, a) + b_k(s, a)$ with $b_k(s, a)$ the intrinsic reward (the above one being $b_k(s,a) = c H \frac{\log\left(SAT/\delta\right)}{\sqrt{N_k(s, a)}}$).

The faster the intrinsic reward (also called exploration bonus) decays to 0,
the faster the optimal Q function for the agent's reward converges uniformly to the optimal Q function and then the policy converges to optimal.
However, the faster the intrisic rewards decay the harder it will be to prove (or for it to be true) that the the optimal Q function is upper bounded by the Q function arising
from the sum of the (empirical) extrinsic and intrinsic rewards. A more involved exploration bonus
decaying as $\frac{1}{N_k(s, a)}$ (faster
than the typical $\frac{1}{\sqrt{N_k(s, a)}}$) and including extra empirical variance terms obtains the optimal (up to log-factors) regret of $\tilde{\mathcal{O}}(\sqrt{HSAT})$
using the same principle but tighter concentration inequalities \cite{azar-ea-minimax}.

The way in which these methods fall short is in that they're not easily adapted to work with function approximation in a non-tabular setting.
The typical workaround is by defining a parametric density model over states and using it to create \emph{psuedocounts} as a novelty signal for an intrinsic reward,
  analogous to state counts $N_k(s) := \sum_a N_k(s,a)$ \cite{bellemare-ea-pseudocounts}.
  The pseudocounts are defined as follows: let $\rho_k(s) = \rho(s; (s_1, \cdots, s_k))$ be a density model $\rho$
  trained with datapoints $(s_1, \cdots, s_k)$, and evaluated on $s$.
  Let $\rho_k'(s) = \rho(s; (s_1, \cdots, s_k, s))$, the same density model algorithm evaluated in $s$, but trained in $(s_1, \cdots, s_k, s)$.
  Then, the pseudocount is defined as
  \begin{equation}
    \label{eq:pseudocounts}
  \hat N_k(s) = \frac{\rho_k(s) (1 - \rho'_k(s))}{\rho'_k(s) - \rho_k(s)}
  \end{equation}
  This definition might seem odd at first, but it is more reasonable once we realize that if $\matS$ is discrete, and $\rho$ is the count density model $\rho(s; s_1, \cdots, s_k) := \frac{\sum_{i=1}^k \Ind\{s_i = s\}}{k} = \frac{N_k(s)}{k}$ then indeed $N_k(s)$ equals equation \eqref{eq:pseudocounts}. In that sense, $\rho_k(s)$ generalizes the role of
  the discrete density model $\frac{N_k(s)}{k}$, and this is exploited to define pseudocounts that represent this change in density in a way faithful to the behaviour of discrete
  counts.
  The exploration bonuses are then defined analogously to the tabular case, $b_k(s, a) \propto \frac{1}{\sqrt{\hat N_k(s)}}$.

However, let us remember that in the tabular case, counts are used as a proxy for how uncertain we are of the optimal Q function.
In this tabular case, with high probability, the Q function arising from the extrinsic empirical rewards and the exploration bonuses always upper bounds the optimal Q function.
As the counts grow higher, the intrinsic rewards decrease to 0, and this upper bound becomes tight: our agent's Q function converges to the optimal Q function and the policy
converges to the optimal one. Is this the case for pseudocounts? Well it must depend on the density model. If the ratio of the pseudocounts and actual counts exists as $k \to \infty$, then the same
analysis holds. However, we would gain little from using pseudocounts if that's valid.
Pseudocounts' usefuleness lies in generalization of the density model: the fact that some new states
where counts are low are similar for the density model to previously seen states, and thus reducing the corresponding exploration bonus. However, density models
can over-generalize, in which case they assign little novelty to states which are different from a potential reward perspective,
or under-generalize, in which case they assign too little novelty to similar states (assimilating the behaviour of counts, where no state similarity is present)
\cite{ostrovski-ea-density}.
These generalization issues depend not only on the capacity of the density model, but on its precise inductive bias, which can often be difficult to maneuver.
Thus, in order to define a more precise analog to the role of counts, but exploiting generalization and function approximation, we choose to take a step back,
and return to the idea of uncertainty of the Q function.

Let us make our line of attack more precise. Let $b_k: \matS \times \matA \rightarrow \RR$ be a function called exploration bonus (depending implicitly on 
the history $(s_1, a_1, r_1, \cdots, s_k, a_k, r_k)$ from all the episodes). In the tabular case our goal is to define $b_k$ in such a way that the solution
$Q^{*,b_k}$ of the empirical Bellman fixed point equation
$$Q^{*,b_k}(s_t, a_t) = r_t + b_k(s_t, a_t) + \gamma \EE_{s_{t+1} \sim \hat \PP(s' | s_t, a_t)}\left[\max_{a'} Q^{*,b_k}(s', a')\right]$$
upper bounds the true optimal Q function $Q^*$ for all state action pairs (trivially satisfied with $b_k$ sufficiently large) and has $b_k$ going to 0 as fast as possible.
In an analogous way for function approximation, doing a slight abuse of notation, we can formalize our goal by looking for functions $b_k$ such that the solution $Q^{*,b_k}$ of
the regression problem 
\begin{equation} \label{eq:qref-emp}
  Q^{*,b_k} := \argmin_{Q: \matS \times \matA \rightarrow \RR} \sum_{t=1}^k \left(Q(s_t, a_t) - r_t + b_k(s_t, a_t) + \gamma \max_{a'} Q(s_{t+1}, a')\right)^2 + \lambda I(Q)
\end{equation}
upper bounds the optimal Q function. In equation \eqref{eq:qref-emp}, the term $I(Q)$ is some form of regularizer.
If $Q^{*,b_k} \geq Q^*$ and $b_k \to_{k \to \infty} 0$,
then as $k \to \infty$, assuming enough capacity and $\lambda_k \to 0$, our $Q^{*, b_k}$ converges to the optimal policy.

Now, in order for exploration to occur as efficiently as possible, we need $Q^{*,b_k}$ to upper bound $Q^*$ but $b_k$ to vanish as fast as possible. Here is when
generalization comes into play. In the extreme case, if $Q^{*,b_k}$ actually minimizes the expected Bellman residual
\begin{equation} \label{eq:qref-exp}
  \min_{Q: \matS \times \matA \rightarrow \RR} \max_{s,a \in \matS \times \matA} \left(Q(s, a) - R(s) + b_k(s, a) + \gamma \EE_{s' \sim P(s' | s, a)} \max_{a'} Q(s', a') \right)^2
\end{equation},
then it will \emph{always} upper bound $Q^*$, so long as $b_k \geq 0$. In particular, if we were able to minimize \eqref{eq:qref-exp} exactly, we could set $b_k = 0$ and
obtain the solution $Q^*$. In essence, as the minimizer of \eqref{eq:qref-emp} is closer to the minimizer of \eqref{eq:qref-exp}, we will be able to select a smaller $b_k$ and
explore more efficiently. However, a lack of generalization will mean that in order to upper bound $Q^*$ we will need $b_k$ to be nonzero and reasonably large.

In the same way as in the analysis of the previous section, the generalization from \eqref{eq:qref-emp} to \eqref{eq:qref-exp} can be seen as an out of
distribution generalization problem,
which we can also see due to the appearance of the $\max_{s, a}$ term in \eqref{eq:qref-exp} in bearance of distributional robustness.
Setting the $b_k$ terms $\eqref{eq:qref-emp}$ can then be seen as having a regression problem $(X^e, Y^e)$ with for $e \in \Etrain$,
and looking for functions $b^e: \matX \rightarrow \RR$
such that the predictor arising from an algorithm (which could either be ERM or an IRM variant) with the training environemnts given by the modified regression
problems $(X^e, Y^e + b^e(X^e))$ upper bounds the optimal $R^\OOD$ predictor for $\Eall$. Again, if we were able to find the optimal OOD predictor from $\Etrain$, we could simply set $b^e = 0$. This analysis to us means two things: first, we should probably be starting to use regularizers $I(Q)$ in equation
\eqref{eq:qref-emp} that are designed to attack this particular out of distribution generalization problem, perhaps with an invariance flavour.
Second, we should look into the particular structure of this
task (i.e. the $\Etrain$ and $\Eall$) such that we can calculate $b^e$ for which we (even loosely) upper bound the optimal $R^\OOD$ predictor. This
obviously doesn't assume we know $R^\OOD$, just that we can bound it by adding decaying exploration bonuses.

A crucial difference from this line of attack and that of density models is that generalization will always play in our favor.
In the case of pseudocounts, the over-generalization of the density model can lead to us failing to assign novelty to important states, and reverting back to random exploration afterwards.
Here, this is impossible: setting the inductive bias of $Q$ (either by the regularizer in \eqref{eq:qref-emp} or the function class) to reduce the generalization gap
to \eqref{eq:qref-exp} will always mean that we can pick a smaller $b_k$, thus geting closer to the optimal Q function and better approximating the optimal policy.

\printbibliography

\clearpage

\begin{appendices}
\chapter{Additional Theorems and Assumptions} \label{app:theo}

\begin{theo} [Equivalence between Invariance, Causality and Out of Distribution Generalization] \label{theo:cauout}
  Let $\mathcal{G} = (V, E)$ be the graph of a structural equation model $\mathcal{C}$, $Y$ one of its nodes, $X$ be $V \setminus Y$,
  and $\Eall(\mathcal{C})$ the corresponding set of environments arising
from intervening in any (or none, thus including the observational environment) of the structural equations except the one from $Y$.
  We assume that for any $i \in \PA(Y)$, there is an $x \in \mathcal{X}$ such that
  $\partial_i \EE_{N_Y}[f_Y(x, N_Y)] \neq 0$ (otherwise changes in $X_i$ wouldn't actually affect $Y$ in any
  useful way). 

  Given a predictor $v: \mathcal{X} \rightarrow \mathcal{Y}$ let the risk for environment $e$ be defined
  $R^e(v) = \EE\left[ \ell(v(X^e), Y^e)\right]$ with $\ell$ the MSE or cross-entropy.
  Let $\Phi: \mathcal{X} \rightarrow \hmatH$ and $w: \hmatH \rightarrow \mathcal{Y}$.
  Then, the following three statements are equivalent:
  \begin{enumerate}
    \item The featurizer $\Phi$ leads to an invariant predictor $(w \circ \Phi)$ with respect to $\Eall(\mathcal{C})$.
    \item We can recover the causal parents of $Y^e$ from $(w \circ \Phi)$,
      and all the information of nodes that are not causal parents is thrown away by $\Phi$.
      More formally, if $\hH^e = \Phi(X^e)$ then 
      $w(\hat H^e) = \EE_{N_Y}\left[f_Y(X^e_{\text{Pa}(Y)}, N_Y)\right]$
      and if we denote the set of non-causal variables by $X^\perp = V \setminus \PA(Y)$, then
      $$X^\perp = \{i: \partial_i \left(w \circ \Phi\right)(x) =  0 \text{ for all $x \in \RR^p$}\}$$
      $$\PA(Y) = \{i: \partial_i \left(w \circ \Phi\right)(x) \neq  0 \text{ for some $x \in \RR^p$}\}$$
      namely the causal parents are exactly the variables which change the predictor. Furthermore,  
      $\EE^e[Y^e | \Phi(X^e) = \Phi(x)] = \EE^e[Y^e | X^e_{\text{Pa}(Y)} = x_{\text{Pa}(Y)}]$ for all $x \in \mathcal{X}$, $e \in \Eall(\mathcal{C})$.
      That is, $\Phi$ already filters out all non-causal information.
    \item Any predictor trained on top of the features obtains optimal out of distribution generalization.
      More specifically,
      $$ \max_{e \in \Eall(\mathcal{C})} R^e\left(w \circ \Phi\right) = \min_{v: \mathcal{X} \rightarrow \mathcal{Y}} \max_{e \in \Eall(\mathcal{C})} R^e\left(v\right)$$
      In addition, $w \in \argmin_{\bar w: \hmatH \rightarrow \mathcal{Y}} R^e(\bar w \circ \Phi)$ for all $e \in \Eall(\mathcal{C})$,
      which means that it can be estimated from any environment where $\PP^e(\hH^e)$ has full support.
  \end{enumerate}

The same equivalence holds if we only consider a linear structural equation model (a SEM
where all equations are linear), and linear interventions.
In this case, both the featurizer $\Phi$ and $w$ are linear. Furthermore, we
also have $\PA(Y) = \{i \in V: (w \circ \Phi)_i \neq 0\}$, so we can recover
the causal parents from an invariant predictor trivially by checking which entries
are nonzero.
\end{theo}

\begin{theo} \label{theo:genp1}
  Let $\Sigma_{X,X}^e := \EE_{X^e}[{X^e}^\top X^e] \in \SSS^{d \times d}_+$, with $\SSS^{d \times d}_+$ the space of symmetric positive semi-definite matrices, and $\Sigma_{X,\epsilon}^e := \EE_{X^e}[X^e \epsilon^e] \in \RR^d$.
  Then, for any arbitrary tuple $\left(\Sigma_{X, \epsilon}^e\right)_{e \in \Etrain} \in {\left(\RR^d\right)}^{|\Etrain|}$, the set 
  $$ \{(\Sigma_{X,X}^e)_{e \in \Etrain}\text{ such that $\Etrain$ does \emph{not} satisfy general position} \}$$
  has measure zero in $(\SSS_+^{d \times d})^{|\Etrain|}$.
\end{theo}

\begin{assu}[Nonlinear general position] \label{a:ngpos}
  Let $\mathcal{F}$ be a $q-$dimensional smooth manifold of functions $\Phi:\matX \rightarrow \RR^p$
  with $\nabla_x \Phi(x)$ full rank for all $x \in \RR^d, \Phi \in \mathcal{F}$.
  Let $c^e: \mathcal{F} \rightarrow \RR^p$ be defined as 
  $$c^e(\Phi) = \EE_{X^e}\left[\Phi(X^e) \Phi(X^e)\right]^{-1} \EE_{X^e, Y^e} \left[\Phi(X^e)^T Y^e\right]  - w$$
    for a fixed vector $w \in \RR^p$.
  Let $F(\Phi) \in \RR^{p |\Etrain|}$ be defined as the concatenation of the $c^e(\Phi)$.
  We say that $e \in \Etrain$ lie in general position with respect to $\mathcal{F}$ if
  $$ \nabla_\Phi F(\Phi) \quad \text{has full rank for all $\Phi\in \mathcal{F}$}$$
  and if the problem $F(\Phi) = 0$ is strictly overdetermined (i.e. $p |\Etrain| > q$, it has more equations than incognita, namely) and
  the set of $\Phi$ such that $F(\Phi)$ consists of isolated points, then the set of solutions is at most a single point.
\end{assu}

\begin{theo}
  \label{theo:nonlingen}
  Let $(X^e, Y^e) \sim \PP^e$ with $e \in \Eall$ be a family of distributions. Let $\Etrain \subseteq \Eall$ be a finite subset.
  Let $\mathcal{F}$ be a smooth manifold of functions from $\mathcal{X}$ to $\hmatH = \RR^p$ that we call featurizers, and $q$
  the dimension of $\mathcal{F}$. Furthermore,
  let's assume that $\nabla_x \Phi$ has full rank for all $x \in \mathcal{X}$, $\Phi \in \mathcal{F}$.
    We also assume that there is a unique unknown $\Phi^* \in \mathcal{F}$  and $w_* \in \RR^p$ such that
      $$ \EE[Y^e | \Phi^*(X^e) = \hh] = w_*^T \hh \quad \text{ for all $e \in \Eall$} $$
    Then, if $\Phi \in \mathcal{F}$ leads to an invariant predictor $v = w \circ \Phi$ across 
    $\Etrain$ where $w$ is linear, the environments in $\Etrain$ lie in nonlinear general position (see \autoref{a:ngpos}),
    and if $|\Etrain| > q/p$, this implies that  $\Phi = \Phi^*$, which leads to the invariant predictor $v = w_* \circ \Phi$ for all $e \in \Eall$.
\end{theo}

\chapter{Proofs of things} \label{app:proofs}

\section{Proof of \autoref{prop:robust}}
\begin{proof}
Let
\begin{align*}
f^\star &\in \min_{f} \max_{e \in \Etrain} R^e(f) - r_e,\\
M^\star     &= \max_{e \in \Etrain} R^e(f^\star) - r_e.
\end{align*}
Then, the pair $(f^\star, M^\star)$ solves the constrained optimization problem
\begin{align*}
\min_{f, M}& \quad M \nonumber \\
\text{s.t.}& \quad R^e(f) - r_e \leq M \quad \text{for all $e \in \Etrain$},
\end{align*}
with Lagrangian $L(f, M, \lambda) = M + \sum_{e \in \Etrain} \lambda^e (R^e(f) - r_e - M)$.
If the problem above satisfies the KKT differentiability and qualification conditions, then there exist $\lambda^e \geq 0$ with $\nabla_f L(f^\star, M^\star, \lambda) = 0$, which means
$$ \nabla_f|_{f=f^\star} \sum_{e \in \Etrain} \lambda^e R^e(f) = 0$$
finishing the proof
\end{proof}

\section{Proof of \autoref{theo:invtest}}
\begin{proof}
  Let $\Psi^\hh: \mathcal{Y} \rightarrow \RR$ be defined as
  $$ \Psi^\hh(y) = \ell(w(\hh), y) $$
  Then, $|\Psi^\hh(y)| \leq C$ by the assumption that $\ell$ is bounded by $C$. Therefore, we have that
  $$ | \EE_{y \sim \PP^e(Y^e | \Phi(X^e) = \hh)}[\psi^\hh(y)] - \EE_{y \sim \PP^{e'}(Y^{e'} | \Phi(X^{e'}) = \hh)}[\psi^\hh(y)]| \leq C \delta_{\text{INV}} $$
  by the fact that the total variation distance between both conditional distributions is less than $\delta_{\text{INV}}$.
  Taking $e' = e_{\text{train}}$ and $e = e_{\text{test}}$ since both are in $\Eall$, we get
  \begin{align*}
    \EE_{(x, y) \sim \Ptest}\left[ \ell(w(\Phi(x)), y) \right] 
      &= \EE_{x \sim \Ptest(X)} \left[ \EE_{y \sim \Ptest(Y | \Phi(X) = \Phi(x))}\left[ \ell(w(\Phi(x)), y)\right] \right] \\
      &= \EE_{x \sim \Ptest(X)} \left[ \EE_{y \sim \Ptest(Y | \Phi(X) = \Phi(x))}\left[ \Psi^{\Phi(x)}(y)\right] \right] \\
      &\leq C \delta_{\text{INV}} + \EE_{x \sim \Ptest(X)} \left[ \EE_{y \sim \Ptrain(Y | \Phi(X) = \Phi(x))}\left[ \Psi^{\Phi(x)}(y)\right] \right] \\
      &\leq C \delta_{\text{INV}} + \EE_{x \sim \Ptest(X)} \left[ \delta_{ERR} \right] \\
      &\leq \delta_{ERR} + C \delta_{\text{INV}}
  \end{align*}
  finishing the proof.
\end{proof}

\section{Proof of \autoref{theo:solutions}}
\begin{proof}
Let $\Phi\in\RR^{p\times d}$, $w \in \RR^{p}$, and $v = w \circ \Phi = \Phi^\top w$.
The simultaneous optimization
\begin{equation}
\label{eq:pb-in-Sw}
  w^\star \in \argmin_{w \in\RR^{p}} R^e(w\circ\Phi) \quad \text{for all $e \in \envs$}
\end{equation}
is equivalent to 
\begin{equation}
\label{eq:pb-in-v}
  v^\star \in \argmin_{v\in\mathcal{G}_{\Phi}} R^e(v)\quad \text{for all $e \in \envs$},
\end{equation}
where $\mathcal{G}_{\Phi}=\{\Phi^\top w : w\in\RR^{p}\}\subset\RR^{d}$ is the
set of vectors~$v=\Phi^\top$ reachable by picking any $w\in\RR^{p}$.
It turns out that $\mathcal{G}_{\Phi}=\mathrm{Ker}(\Phi)^\perp$, that
is, the subspace orthogonal to the nullspace of $\Phi$. Indeed, for
all $v=\Phi^\top w\in\mathcal{G}_{\Phi}$ and all $x\in\mathrm{Ker}(\Phi)$,
we have $x^\top v=x^\top \Phi^\top w = (\Phi x)^\top w = 0$. Therefore
$\mathcal{G}_{\Phi}\subset\mathrm{Ker}(\Phi)^\perp$. Since both
subspaces have dimension $p - \mathrm{rank}(\Phi)=\mathrm{dim}(\mathrm{Ker}(\Phi))$,
they must be equal.

We now prove the theorem: let $v=\Phi^\top w$ where $\Phi\in\RR^{p\times d}$
and~$w\in\RR^{p}$ minimizes all~$R^e(w\circ\Phi)$. Since $v\in\mathcal{G}_\Phi$,
we have~$v\in\mathrm{Ker}(\Phi)^\perp$. Since~$w$ minimizes~$R^e(\Phi^\top {w})$, we can also write
\begin{equation}
\label{eq:vderivative}
\frac{\partial}{\partial w}\:R^e(\Phi^\top {w})
= \Phi\,\nabla{R^e}(\Phi^\top w) = \Phi \nabla{R^e}(v) =  0~.
\end{equation}
Therefore~$\nabla{R^e}(v)\in\mathrm{Ker}(\Phi)$.
Finally~$v^\top\nabla{R^e}(v)=w^\top \Phi\,\nabla{R^e}(\Phi^\top w)=0$.

Conversely, let $v\in\RR^d$ satisfy~$v^\top\nabla{R^e}(v)=0$ for
all~$e\in\mathcal{E}$. Thanks to these orthogonality conditions, we can construct a
subspace that contains all the~$\nabla{R^e}(v)$ and is
orthogonal to~$v$. Let~$\Phi$ be any matrix whose nullspace satisfies
these conditions. Since $v\perp\mathrm{Ker}(\Phi)$, that is, $v\in\mathrm{Ker}(\Phi)^\perp=\mathcal{G}_{\Phi}$,
there is a vector $w \in \RR^{p}$ such that $v=\Phi^\top {w}$.
Finally, since $\nabla{R^e}(v)\in\mathrm{Ker}(\Phi)$, the
derivative~\eqref{eq:vderivative} is zero.
\end{proof}

\section{Proof of \autoref{theo:lingen}}
\begin{proof}

Observing that $\Phi \, \EE_{X^e, Y^e}\left[{X^e} Y^e\right] = \Phi\, \EE_{X^e, \epsilon^e}[{X^e}({(\tilde S X^e)}^\top \gamma + \epsilon^e)]$, we re-write~\eqref{eq:theo_cond} as
\begin{equation}
  \Phi \left(\underbrace{\EE_{X^e}\left[{X^e} {X^e}^\top \right] (\Phi^\top w - \tilde S^\top \gamma) - \EE_{X^e, \epsilon^e}\left[{X^e} \epsilon^e\right]}_{:= q_e}\right) = 0.
    \label{eq:optmatchnew}
\end{equation}
To show that $\Phi$ leads to the desired invariant predictor $\Phi^\top w = \tilde S^\top \gamma$, we assume $\Phi^\top w \neq \tilde S^\top \gamma$ and reach a contradiction.
First, by Assumption~\ref{a:gpos}, we have $\text{dim}(\text{span}(\{q_e\}_{e \in \Etrain})) > d - r$.
Second, by~\eqref{eq:optmatchnew}, each $q_e \in \text{Ker}(\Phi)$.
    Therefore, it would follow that $\text{dim}(\text{Ker}(\Phi)) > d-r$, which contradicts the assumption that $\text{rank}(\Phi) = r$.
\end{proof}

\section{Proof of \autoref{theo:nonlingen2}}
\begin{proof}
If $\Phi: \mathcal{X} \rightarrow \hmatH$ leads to an invariant predictor in this problem, this means that
for $\hat H^e = \Phi(X^e)$ we have
$$ P(Y^e | \hat H^e) = P(Y^{e'} | \hat{H^{e'}}) $$
Let us define $G_\Phi(H^e, Z^e) = (\Phi \circ F) (H^e, Z^e)$. 
Now, for all $\hat h \in \hmatH$, we have
\begin{align*}
  P(Y^e = 1 | \hat H^e = \hat h) &= \sum_{(h, z) \in \mathcal{H} \times \mathcal{Z}} P(Y^e = 1, Z^e = z, H^e = h | \hat{H}^e = \hat h) \\
  &= \frac{\sum_{(h, z) \in G^{-1}(\hat h)} P(Y^e = 1, Z^e = z, H^e = h)}{P(\hat{H}^e = \hat h)} \\
  &= \frac{\sum_{(h, z) \in G^{-1}(\hat h)} P(Z^e = z | Y^e = 1) P(Y^e = 1 | H^e = h) P^e(H^e= h)}{P(\hat{H}^e = \hat h)} \\
\end{align*}
We now introduce the notation $P(Y^e = 1 | H^e = h) = v_h$, with $v \in \RR^d$ (and $d$ the number of possible values of $P^e$). Note
that since $H^e$ is the cause of $Y^e$, we have that $v$ does not depend on the environment. Furthermore, we denote as $B^0_{\Phi, \hat h}$ the set
$$B^0_{\Phi, \hat h} = \{h \in \mathcal{H} \text{ such that } G_\Phi(h, 0) = \hat h, G_\Phi(h, 1) \neq \hat h\}$$
In the same way, we denote as $B^1_{\Phi, \hat h}$ the set of $h$ for
which $G_\Phi(h, 0) \neq \hat h, G_\Phi(h, 1) = \hat h$, and finally $B^2_{\Phi, \hat h}$ the set of $h$ for which both $G_\Phi(h, 0) = G_\Phi(h, 1) = \hat h$.
We also denote by $A^0_\Phi \in \RR^{a \times d}$ the matrix such that $(A^0_\Phi w)_{\hat h} = \sum_{h \in B^0_{\Phi, \hat h}} w_h$, and in the same way for
$A^1_\Phi$ and $A^2_\Phi$. Thus, we have that the right hand side of the above equation equals
\begin{align*}
  P(Y^e = 1 | \hat H^e = \hat h) &= \frac{\sum_{(h, z) \in G^{-1}(\hat h)} P(Z^e = z | Y^e = 1) P(Y^e = 1 | H^e = h) P^e(H^e= h)}{P(\hat{H}^e = \hat h)}\\
    &=\frac{1}{P(\hat{H}^e = \hat h)}\left(
    \sum_{h \in B^1_{\Phi, \hat h}} (1 - p^e_Z) (v \odot P^e_H)_h +
    \sum_{h \in B^2_{\Phi, \hat h}} p^e_Z (v \odot P^e_H)_h + \sum_{h \in B^e_{\Phi, \hat h}} (v \odot P^e_H)_h\right) \\
    &= \frac{1}{P(\hat{H}^e = \hat h)}\left((1 - p_Z^e) (A^1_\Phi (v \odot P_H^e))_{\hat h} + p_Z^e (A^2_\Phi (v \odot P_H^e))_{\hat h} +
    (A^3_\Phi (v \odot P_H^e))_{\hat h}\right)
\end{align*}
Rewriting $P(Y^e = 1 | \hat H^e = \hat h)$ as $c^e_{\Phi, h}$, we have
\begin{align}
  c^e_\Phi &= \frac{1}{P(\hat{H}^e = \hat h)}\left((1 - p_Z^e) A^1_\Phi + p_Z^e A^2_\Phi + A^3_\Phi \right) (v \odot P_H^e) \nonumber\\
  &= \frac{1}{P(\hat{H}^e = \hat h)}\left((A^1_\Phi + p_Z^e (A^2_\Phi - A^1_\Phi) + A^3_\Phi \right) (v \odot P_H^e) \nonumber\\
  &= \frac{\left((A^1_\Phi + p_Z^e (A^2_\Phi - A^1_\Phi) + A^3_\Phi \right) (v \odot P_H^e)}{\left((A^1_\Phi + p_Z^e (A^2_\Phi - A^1_\Phi) + A^3_\Phi \right) (\matr{1}_d \odot P_H^e)} \label{eq:rhsinvmn}
\end{align}
We denote the last right hand side \eqref{eq:rhsinvmn} as $J_\Phi(p_Z^e)$. Now, $\Phi$ leads to an invariant predictor among $\Etrain$ if and only if $c^e_\Phi = c^{e'}_\Phi$.
Now, it is clear that if $A^2_\Phi - A^1_\Phi \neq 0$, then $\eqref{eq:rhsinvmn}$ is different for almost all pairs $(p_Z^e, p_Z^{e'})$. Namely, the set
$$\{(p_Z^e, p_Z^{e'}) \text{ such that } J_\Phi(p_Z^e) = J_\Phi(p_Z^{e'})\}$$
has measure 0 in $[0, 1] \times [0,1]$ if $A_\Phi^2 - A^1_\Phi \neq 0$. Thus, for almost all pairs $(p_Z^e, p_Z^{e'})$ we have that those $\Phi$ with invariant
prediction satisfy $A^2_\Phi = A^1_\Phi$. But $A^2_\Phi$ and $A^1_\Phi$ are indicator matrices of disjoint sets. Therefore, for these $\Phi$ with invariant prediction
we must have $A^2_\Phi = A^1_\Phi = 0$, and by defnition, $B^2_{\Phi, \hat h} = B^1_{\Phi, \hat h} = \emptyset$.
This trivially implies that for all $\hat h \in \hat{\mathcal{H}}$, either both $(h, 1), (h, 0) \in G_\Phi^{-1}(\hat h)$, or neither are. In particular, this trivially
means that $G_\Phi(h, 0) = G_\Phi(h, 1)$, and thus that $G_\Phi(h, z) = G(h)$ for some function $G$, concluding the proof.
\end{proof}

\section{Proof of \autoref{theo:cauout}}
\begin{proof}
  For brevity, throughout the proof we will use $\Eall$ to mean $\Eall(\mathcal{C})$ since no other family of environments will be involved.
\item
  \paragraph{$(1) \Rightarrow (2)$:} By the definition of invariant prediction, $(1)$ means that $w \in \argmin_{\bar w: \hmatH \rightarrow \mathcal{Y}} 
  R^e(\bar w \circ \Phi)$ for all $e \in \Eall$. If $x \in \mathcal{X}$, we will denote $e(x) \in \Eall$ the environment coming from intervening with the structural
  equations $X_i^{e(x)} = x_i$ (so standard DO interventions \cite{pearl2009causation}).

  Let us denote by $\Psi: \mathcal{X} \rightarrow \mathcal{Y}$
  the function$\Psi(x) =\EE_{N_Y} \left[ f_Y(x_{\text{PA}(Y)}, N_Y) \right]$. 
  Since $w \circ \Phi$ has invariant prediction with respect to $\Eall$, then $w \in \argmin_{\bar w: \hmatH \rightarrow \mathcal{Y}} R^{e(x)}(\bar w \circ \Phi)$ for all
  $x \in \mathcal{X}$. Now, in these environments, $X^{e(x)}$ is constant, and hence independent of $Y^e$. This implies that
  \begin{align*}
    (w \circ \Phi)(x) &= \EE\left[Y^{e(x)} | \Phi\left(X^{e(x)}\right) = x\right] \\
      &= \EE\left[Y^{e(x)}\right] \\
    &= \EE_{X^{e(x)}} \left[ \EE_{N_Y}\left[f_Y(X^{e(x)}_{\text{Pa}(Y)}, N_Y) \right] \right] \\
    &= \EE_{N_Y} \left[ f_Y(x_{\text{PA}(Y)}, N_Y) \right] \\
    &= \Psi(x)
  \end{align*}
  
  The first equality is due to $c$ being optimal on top of $\Phi$ for environment $e(x)$, the second one by the independence between $X^{e(x)}$ and $Y^{e(x)}$, and the
  last three by the definition of $Y^e$, the fact that $X^{e(x)} = x$, and the definition of $\Psi$ respectively.

  Since this equality is true for all $x \in \mathcal{X}$, then $(w \circ \Phi) = \Psi$ as functions. Therefore, $(w \circ \Phi)(X^e) = \Psi(X^e)$ for all $e \in \Eall$, which
  is the first part of $(2)$. We now have to prove that the set of causal parents are exactly those that change $(w \circ \Phi)$. But this is easy, since
  if $i \notin \PA(Y)$, we have $\partial_i (w \circ \Phi) = \partial_i \Psi = 0$ by definition. Finally, if $i \in \PA(Y)$, then $\partial_i (w \circ \Phi)
  = \EE_{N_y} \left[ \partial_i f_Y(x_{\PA(y)}, N_Y) \right]$ which is nonzero for some $x \in \mathcal{X}$ by assumption.

  Finally, we need to show that $\EE^e[Y^e | \Phi(X^e) = \Phi(x)] = \EE^e[Y^e | X^e_{\text{Pa}(Y)} = x_{\text{Pa}(Y)}]$ for all $x \in \mathcal{X}$, $e \in \Eall$.
  But this is trivial, since the fact that $w \circ \Phi$ is an invariant predictor for all $e \in \Eall$ means
  \begin{align*}
    \EE^e[Y^e | \Phi(X^e) = \Phi(x)] &= c(\Phi(x)) = \Psi(x) \\
      &= \EE^e[Y^e | X^e_{\text{Pa}(Y)} = x_{\text{Pa}(Y)}] 
  \end{align*}
  finishing this part of the proof.
\item
\paragraph{$(2) \Rightarrow (3)$:} We have to prove two things. First, that $w \circ \Phi$ obtains optimal out of distribution
  generalization, and then that $w \in \argmin_{\bar w: \hmatH \rightarrow \mathcal{Y}} R^e(\bar w \circ \Phi)$ for all $e \in \Eall$.
  
  Denoting $\Psi(X^e) = \EE\left[f_Y(X^e_{\text{Pa}(Y)}, N_Y)\right]$, by $(2)$ we know that
  that $c(\hat H^e) = \Psi(X^e)$, or equivalently $w \circ \Phi = \Psi$. We will show now that $\Psi$ obtains optimal out of distribution generalization,
  and then that $c$ is in the argmin of the risks for all environments on top of $\Phi$.

  Let us denote $R^{\text{OOD}}(v) = \max_{e \in \Eall} R^e(v)$.
  We claim that $\Psi \in \argmin_{v:\mathcal{X} \rightarrow \mathcal{Y}} R^{\text{OOD}}(v)$.
  We proceed by contradiction, assuming that there is a $\tilde \Psi: \mathcal{X} \rightarrow \mathcal{Y}$ such that $R^{\text{OOD}}(\tilde \Psi) < R^{\text{OOD}}(\Psi)$.
  We first note that
  \begin{align}
    R^e(\Psi) &= \EE_{X^e, Y^e}\left[ l(Y^e, \Psi(X^e)) \right] \nonumber \\
      &= \EE_{X^e} \left[ \EE_{N_Y} \left[ l\left( f_Y(X^e_{\text{Pa}(Y)}, N_Y), \Psi(X^e) \right) \right] \right] \nonumber \\
      &\leq \max_{x \in \mathcal{X}} \left[ \EE_{N_Y} \left[ l\left( f_Y(x_{\text{Pa}(Y)}, N_Y), \Psi(x) \right) \right] \right] \label{eq:worstx}
  \end{align}
  If $\mathcal{X}$ is compact and $\Psi$ is continuous, then there is an $\bar x \in \mathcal{X}$ with 
  \begin{equation} \label{eq:worsenv}
    \max_{x \in \mathcal{X}} \left[ \EE_{N_Y} \left[ l\left( f_Y(x_{\text{Pa}(Y)}, N_Y), \Psi(x) \right) \right] \right] = 
    \EE_{N_Y} \left[ l\left( f_Y(\bar x_{\text{Pa}(Y)}, N_Y), \Psi(\bar x) \right) \right] 
  \end{equation}
  Denoting by $e^* \in \Eall$ the environment that comes from the interventions $X^{e^*}_i \leftarrow \bar x_i$, we can rewrite the right
  hand side of \eqref{eq:worsenv} as $R^{e^*}(\Psi)$. Continuing from \eqref{eq:worstx}, this means $R^e(\Psi) \leq R^{e^*}(\Psi)$ for all $e \in \Eall$,
  and thus
  $$ R^{e^*}(\Psi) \leq \max_{e \in \Eall} R^e(\Psi) \leq R^{e^*}(\Psi) $$
  which implies $R^\text{OOD}(\Psi) = R^{e^*}(\Psi)$.
  
  Now, we have that $R^{e^*}(\tilde \Psi) \leq R^\text{OOD}(\tilde \Psi) < R^\text{OOD}(\Psi) = R^{e^*}(\Psi)$. By construction,
  $\Psi(x)$ is the conditional expectation of $Y^e$ given its causal parents, and in $e^*$ all the noncausal parents are constant and independent
  of $Y^e$. This means that
  $\Psi(\bar x) = \EE_{X^{e^*}, Y^{e^*}}\left[Y^{e^*} | X^{e^*} = \bar x\right]$, which means that
  $\Psi \in \argmin_{g: \mathcal{X} \rightarrow \mathcal{Y}} R^{e^*}(g)$ contradicting the fact that $R^{e^*}(\tilde \Psi) < R^{e^*}(\Psi)$. This concludes the part
  of the proof showing that $\Psi$ has optimal out of distribution generalization.

  Now, the part remaining is to show that $w \in \argmin_{w: \hmatH \rightarrow \mathcal{Y}} R^e(w \circ \Phi)$ for all $e \in \Eall$. But this is trivial, 
  since $\Phi$ discards the noncausal variables, and $w \circ \Phi$ is the conditional expectation of $Y^e$ given the causal ones, which means that it achieves optimal
  risk in all environments among all the predictors that don't use the noncausal variables.
  
  More formally, 
  we know by $(1)$ that $\EE^e[Y^e | \Phi(X^e)] = \EE^e[Y^e | X^e_{\text{Pa}(Y)} ]$ for all $e \in \Eall$.
  This then trivially means
  \begin{align*}
    R^e(w \circ \Phi) &= R^e(\Psi) \\
      &= R^e(\EE[Y^e | X^e_{\text{Pa}(Y)}]) \\
      &= R^e(\EE[Y^e | \Phi(X^e)]) \\
      &\leq \min_{w: \hmatH \rightarrow \mathcal{Y}} R^e(w \circ \Phi)
  \end{align*}
  the last inequality due to the fact that the conditional expectation is optimal for the loss $e$.
  This by definition means that $w \in \argmin_{\bar w: \hmatH \rightarrow \mathcal{Y}} R^e(\bar w \circ \Phi)$.
\item 
\paragraph{$(3) \Rightarrow (1)$:} Since $w \in \argmin_{\bar w: \hmatH \rightarrow \mathcal{Y}} R^e(\bar w \circ \Phi)$ for all $e \in \Eall$, by definition $\Phi$ leads to the invariant predictor
  $w \circ \Phi$ for all $e \in \Eall$.
\end{proof}

\section{Proof of \autoref{theo:genp1}}
\begin{proof}
  Let $m = |\Etrain|$, and define $G: \RR^d \setminus \{0\} \rightarrow \RR^{m \times d}$ as
  $ \left(G(x)\right)_{e, i} = \left(\Sigma^e_{X,X} x - \Sigma_{X,\epsilon}^e\right)_i.$

  Let $W = G\left(\RR^d \setminus \{0\} \right) \subseteq \RR^{m \times d}$, which is a linear manifold of dimension at most $d$, missing a single point (since $G$ is affine, and its input has dimension $d$).

  For the rest of the proof, let $(\Sigma^e_{X, \epsilon})_{e \in \Etrain} \in {\RR^{d}}^{|\Etrain|}$ be arbitrary and fixed.  
  We want to show that for generic $(\Sigma^e_{X,X})_{e \in \Etrain}$, if $m > \frac{d}{r} + d - r$, the matrices $G(x)$ have
  rank larger than $d - r$. Analogously, if $\text{LR}(m, d, k) \subseteq \RR^{m \times d}$ is the set of $m \times d$ matrices with rank $k$, we want to show
  that $W \cap \text{LR}(m, d, k) = \emptyset$ for all $k < d - r$.

  We need to prove two statements.
  First, that for generic$(\Sigma^e_{X,X})_{e \in \Etrain}$
  $W$ and $\text{LR}(m, d, k)$ intersect transversally as manifolds, or don't intersect at all. This will be a standard argument using Thom's transversality theorem.
  Second, by dimension counting, that if $W$ and $\text{LR}(m, d, k)$ intersect transversally, and $k < d - r$, $m > \frac{d}{r} + d - r$, then
  the dimension of the intersection is negative, which is a contradiction and thus $W$ and $\text{LR}(m, d, k)$ cannot intersect.

  We then claim that $W$ and $\text{LR}(m, d, k)$ are transversal for generic $(\Sigma^e_{X,X})_{e \in \Etrain}$.
    To do so, define $$F: (\RR^d \setminus \{0\}) \times \left(\SSS_+^{d \times d}\right)^m \rightarrow \RR^{m \times d},$$
  $$F\left(x, \left(\Sigma^e_{X,X}\right)_{e \in \Etrain}\right)^{e'}_l = \left(\Sigma_{X,X}^{e'} x - \Sigma_{X,\epsilon}^{e'}\right)_l$$
  If we show that $\nabla_{x, \Sigma_{X,X}} F: \RR^d \times (\SSS^{d \times d})^m \rightarrow \RR^{m \times d}$ is a surjective linear transformation,
  then $F$ is transversal to any submanifold of $\RR^{m \times d}$ (and in particular
  to $\text{LR}(m, d, k)$). By the Thom transversality theorem, this will imply that the set of $\left(\Sigma^e_{X,X}\right)_{e \in \Etrain}$ such that $W$ is not transversal
  to $\text{LR}(m, d, k$) has measure zero in $\SSS_{+}^{d \times d}$, proving our first statement.

  Let us then show that $\nabla_{x, \Sigma_{X,X}} F$ is surjective. This follows by 
  showing that $\nabla_{\Sigma_{X,X}} F: (\SSS^{d \times d})^m  \rightarrow \RR^{m \times d}$
  is surjective, since adding more columns to this matrix can only increase its rank.
  We then want to show that the linear map $\nabla_{\Sigma_{X,X}} F: (\SSS^{d \times d})^m  \rightarrow \RR^{m \times d}$ is surjective.
  To this end, we can write:
  $ \partial_{\Sigma_{i,j}^e} F^{e'}_l = \delta_{e,e'}\left(\delta_{l,i} x_j + \delta_{l,j} x_i\right)$.
  Let $C \in \RR^{m \times d}$. We want to construct a $D \in \left(\SSS^{d \times d}\right)^m$ such that
  $$ C^{e'}_l = \sum_{i, j, e} \delta_{e,e'} \left(\delta_{l,i} x_j + \delta_{l,j} x_i\right) D^e_{i,j}. $$
  The right hand side equals
  \begin{align*}
    \sum_{i, j, e} \delta_{e,e'} \left(\delta_{l,i} x_j + \delta_{l,j} x_i\right) D^e_{i,j} &= \sum_{j} D^{e'}_{l,j} x_j +\sum_{i} D^{e'}_{i,l} x_i 
      = (D^{e'} x)_l + (x D^{e'})_l
  \end{align*}
  If $D^{e'}$ is symmetric, this equals $(2 D^e x)_l$. Therefore, we only need to show that for any vector $C^e \in \RR^d$, there is a symmetric matrix
  $D^e \in \SSS^{d \times d}$ with $C^e = D^e x$.
  To see this, let $O \in \RR^{d \times d}$ be an orthogonal transformation such that $O x$ has no zero entries,
  and name $v = O x, w^e = O C^e$. Furthermore, let $E^e \in \RR^{d \times d}$ be the diagonal matrix with entries $E^e_{i,i} = \frac{w^e_i}{v_i}$. Then,
  $C^e = O^T E^e O x$. By transposing, $O^T E^e O$ is symmetric, showing that $\nabla_{\Sigma_{X,X}} F: (\SSS^{d \times d})^m  \rightarrow \RR^{m \times d}$
  is surjective, and thus that $W$ and $\text{LR}(m, d, k)$ are transversal for almost any $\left(\Sigma_{X,X}^e\right)_{e \in \Etrain}$.

  By transversality, we know that $W$ cannot intersect $\text{LR}(m, d, k)$ if $\dim(W) + \dim\left(\text{LR}(m, d, k)\right) - \dim\left(\RR^{m \times d}\right) < 0$.
  By a dimensional argument (see \cite{lee2003introduction}, example 5.30), it follows that 
  $\codim(\text{LR}(m, d, k)) = \dim\left(\RR^{m \times d}\right) - \dim\left(\text{LR}(m, d, k)\right) = (m - k) (d - k)$.
  Therefore, if $k < d - r$ and $m > \frac{d}{r} + d - r$, it follows that
  \begin{align*}
      \dim(W) + \dim\left(\text{LR}(m, d, k)\right) - \dim\left(\RR^{m \times d}\right) &= \dim(W) - \codim\left(\text{LR}(m, d, k)\right) \\
      &\leq d - (m - k)(d - k) \\
      &\leq d - (m - (d - r))(d - (d - r)) \\
      &= d - r (m - d + r) \\
      &< d - r\left(\left(\frac{d}{r} + d - r\right)-d + r\right) \\
      &= d - d = 0.
  \end{align*}
  Therefore, $W \cap \text{LR}(m, d, k) = \emptyset$ under these conditions, finishing the proof.
\end{proof}

\section{Proof of \autoref{theo:nonlingen}}
\begin{proof} 
  We assume that $\tilde \Phi \in \mathcal{F}$ leads to an invariant predictor $(w \circ \tilde \Phi)$ across $\Etrain$, with $w$ linear,
  and hence we denote $v \in \RR^a$.
  Using the same notation as in \autoref{a:ngpos}, for any $\Phi$ we denote $c^e(\Phi) \in \RR^p$ as the optimal linear least squares
  regressor on top of $\Phi$ on environment $e \in \Etrain$ minus $w$.
  Furthermore, we denote by $F(\Phi) \in \RR^{p |\Etrain|}$ the concatenation of these $c^e(\Phi)$.
  Since $(w \circ \tilde \Phi)$ is an invariant predictor across $\Etrain$, we have that $F(\tilde \Phi) = 0$.
  Since $F$ is defined from a manifold of dimension $q$ to $\RR^{p |\Etrain|}$, with $\frac{q}{p} < |\Etrain|$, the problem $F(\Phi) = 0$ is strictly
  overdetermined. In combinatioin with $\nabla_\Phi F(\Phi)$ being full rank, this means that the set of $\Phi$ such that $F(\Phi) = 0$ is at
  most a set of isolated points, and by \autoref{a:ngpos}, this means the solution is unique. Since both $\tilde \Phi$ and $\Phi^*$ are
  solutions, this means $\tilde \Phi = \Phi^*$, finishing the proof.
\end{proof}

\chapter{Failure Cases of Domain Adaptation}
\label{app:ada}

Domain adaptation \cite{bendavid-domain} considers labeled data from a source environment $e_\train$ and unlabeled data from a target environment $e_\test$ with the goal of training a classifier that works well on $e_\test$.
Many domain adaptation techniques, including the popular Adversarial Domain Adaptation \citep[ADA]{ganin2016domain}, proceed by learning a feature representation $\Phi$ such that (i) the input marginals $P(\Phi(X^{e_\train})) = P(\Phi(X^{e_\test}))$, and (ii) the classifier $w$ on top of $\Phi$ predicts well the labeled data from $e_\train$.
Thus, are domain adaptation techniques applicable to finding invariances across multiple environments?

One shall proceed cautiously, as there are important caveats.
For instance, consider a binary classification problem, where the only difference between environments is that $P(Y^{e_\train} = 1) = \frac{1}{2}$, but $P(Y^{e_\test} = 1) = \frac{9}{10}$.
Using these data and the domain adaptation recipe outlined above, we build a classifier $w \circ \Phi$.
Since domain adaptation enforces $P(\Phi(X^{e_\train})) = P(\Phi(X^{e_\test}))$, it consequently enforces $P(\hat{Y}^{e_\train}) = P(\hat{Y}^{e_\test})$, where $\hat{Y}^e = w(\Phi(X^e))$, for all $e \in \{ e_\train, e_\test \}$.
Then, the classification accuracy will be at most 20\pc{}.
This is worse than random guessing, in a problem where simply training on the source domain leads to a classifier that generalizes to the target domain.

Following on this example, we could think of applying conditional domain adaptation techniques \cite[C-ADA]{Li_2018_ECCV}.
These enforce one invariance $P(\Phi(X^{e}) | Y^{e}) = P(\Phi(X^{e'}) | Y^{e'})$ per value of $Y^e$, and all pairs $e, e' \in \Etrain$.
Using Bayes rule, it follows that C-ADA enforces a stronger condition than invariant prediction when $P(Y^{e}) = P(Y^{e'})$.
However, there are general problems where the invariant predictor cannot be identified by C-ADA.

To see this, consider a discrete input feature $X^e \sim P(X^e)$, and a binary target $Y^e = F(X^e) \oplus \text{Bernoulli}(p)$.
This model represents a generic binary classification problem with label noise.
Since the distribution $P(X^e)$ is the only moving part across environments, the trivial representation $\Phi(x) = x$ elicits an invariant prediction rule.
Assuming that the discrete variable $X^e$ takes $n$ values, we can summarize $P(X^e)$ as the probability $n$-vector $p^{x, e}$.
Then, $\Phi(X^e)$ is also discrete, and we can summarize its distribution as the probability vector $p^{\phi, e} = A_\phi p^{x, e}$, where $A_\phi$ is a matrix of zeros and ones.
By Bayes rule,
\begin{align*}
    \pi^{\phi, e} := P(\Phi(X^e) | Y^e=1) = \frac{P(Y^e = 1 | \Phi(X^e)) \odot p^{\phi, e}}{\langle P(Y^e = 1 | \Phi(X^e)), p^{\phi, e} \rangle} = \frac{\left( A_\Phi \left(v \odot p^{x, e} \right)\right)\odot (A_\Phi p^{x,e})}{\langle\left( A_\Phi \left(v \odot p_{x, e} \right)\right), A_\Phi p^{x, e} \rangle},
\end{align*}
where $\odot$ is the entry-wise multiplication, $\langle , \rangle$ is the dot product, and $v := P(Y^e = 1 | X^e)$ does not depend on $e$.
Unfortunately for C-ADA, it can be shown that the set $\Pi_\phi := \{ (p^{x, e}, p^{x, e'}) : \pi^{\phi, e} = \pi^{\phi, e'} \}$ has measure zero.
Since the union of sets with zero measure has zero measure, and there exists only a finite amount of possible $A_\phi$, the set $\Pi_\phi$ has measure zero for \emph{any} $\Phi$. 
In conclusion and almost surely, C-ADA disregards any non-zero data representation eliciting an invariant prediction rule, regardless of the fact that the trivial representation $\Phi(x) = x$ achieves such goal. 

As a general remark, domain adaptation is often justified using the bound \cite{bendavid-domain}:
\begin{equation*}
    \text{Error}^{e_\test}(w \circ \Phi) \leq \text{Error}^{e_\train}(w \circ \Phi) + \text{Distance}(\Phi(X^{e_\train}), \Phi(X^{e_\test})) + \lambda^\star.
\end{equation*}
Here, $\lambda^\star$ is the error of the optimal classifier in our hypothesis class, operating on top of $\Phi$, summed over the two domains.
Crucially, $\lambda^\star$ is often disregarded as a constant, justifying the DA goals (i, ii) outlined above.
But, $\lambda^\star$ depends on the data representation $\Phi$, instantiating a third trade-off that it is often ignored.
For a more in depth analysis of this issue, we recommend \citep{Johansson2019SupportAI}.

\chapter{Meta-Learning}
\label{app:meta}

Meta-learning is at the moment of writing an umbrella term encompassing many things: using data to learn an optimizer \cite{andry-ea-meta} or architectures \cite{zoph-ea-nas}
for future learning,
generalizing to new classes other than those seen during training \cite{snell-ea-meta}, self-modifying code \cite{schmidhuber-ea-self},
increasing speed of adaptation to a new dataset from the same task \cite{finn-ea-maml}, and many other topics \cite{thrun-ea-ltol}.

The philosophy generally attached to meta-learning is particularly appealing: phrasing the above OOD problems as variants of assuming that the training and test distributions come from a meta-distribution, and try to reuse as much as possible the technical machinery that works so well in i.i.d. problems \cite{silver-principles, finn-ea-maml}. Furthermore, in the last few
decades models with less inductive bias and feature engineering have overwhelmingly surpassed the performance of more hand-crafted approaches in complicated sensory
tasks, with variants of neural networks now being state of the art in an increasingly large family of language, vision, and speech problems. Therefore, could using less inductive
bias about the distributions at hand not only make things easier by reusing the i.i.d. machinery, but ultimately increase performance?

There is no yes or no answer to this question. As we have seen, different OOD tasks have different structure. ERM and other approaches make sense in some OOD tasks,
and catastrophically fail in others. However, it is our conjecture that for an important class of AI related problems, the meta-learning dream of simply reusing the i.i.d.
machinery is futile, and in this chapter we try to elaborate on why.


\section{A Bayesian Meta-Learner}
In order to understand how meta-learning and other ideas may fall short, it is quite useful to see what the `right thing' to do is, and quantitatively study how we deviate from that.
We can describe a version of the right thing by proceeding in a fully Bayesian way.
Obviously, the output of this Bayesian derivation will be completely intractable from a computational point of view, so from an algorithmic standpoint it will be useless,
  but from an understanding viewpoint it will help us see precisely how we are deviating from the Bayes-optimal solution.

We utilize the notation and setup of \cite{finn-ea-maml} slightly rewritten, which for completeness we now detail.
An important assumption that we'll be dealing with, following from that of \cite{finn-ea-maml}, is that the different training and test environments
  come i.i.d from a meta-distribution $e \sim \fancyp(e)$. This is obviously a strong limitation of meta-learning
  for many real life settings, since it is likely that new environments
are not independent from previous ones. For instance, in a nonstationary environment such as an agent observing new states, the previous environments will lead to actions
that will impact the new states the robot will be in, and hence the new environments. As well, previous users' data will alter the decisions taken by services
like recomendation systems or search engines, and hence alter the future users' reactions. Nonetheless, we need a starting point for this technical discussion,
and this disclaimer aside we could imagine problems where this assumption is closer to the truth.
For instance, we could obtain datasets from different hospitals with data independently collected across them, and want to generalize predictions to a new hospital selected in the same manner. In essence, while this assumption is limiting, it will make the following analysis clearer (and not any less valid),
thus helping elucidate other important downsides of meta-learning approaches.

Proceeding in a full Bayesian manner, let us assume a meta-prior $P(\fancyp)$ that encompasses all our prior knowledge about the meta-distribution $\fancyp$.
Following \cite{finn-ea-maml} we assume access to datasets $\{S^j\}_{j=1, \cdots, |\Etrain|}$, and potentially empty sample sets from the test distribution $S^\test$. Here,
$S^j = \{x^{(i), j}, y^{(i), j}\}_{i=1}^{n_j}$ are sampled i.i.d. from $P(x, y | e_j)$, the data distribution for environment $e_j$.
Given a new test datapoint $x^*$, our goal will be to estimate $P(y^* | x^*, S^1, \dots, S^{|\Etrain|}, S^\test)$.
The expectation of this distribution is the Bayes-optimal prediction for losses such as cross-entropy or MSE. 

For simplicity we assume that $P(x | e)$ is the same for all environments $e \in \Eall$, and that all the distributions involved admit nonzero densities.
This is not strictly necessary for the analysis, but will help us avoid cumbersome calculations that bring nothing to the table.
This then means that $P(y, x | e) = P(y | x, e) P(x)$.
By Bayes rule, we can see
\begin{equation} \label{eq:bopt}
  P(y^* |  x^*, S^1, \dots, S^{|\Etrain|}, S^\test) = \int_{\Eall} P(y^* | x^*, e^\test) P(e^\test | S^{1}, \dots, S^{{|\Etrain|}}, S^\test) \d e^\test
\end{equation}
Now, $P(y^* | x^*, e^\test)$ is a trivial part, since the conditioning assumes we know what the test environment is. In more traditional Bayesian notation, this is written
$P(y^* | x^*, \theta)$, just running the forward pass once we know the parameters of the test environment. The difficulty of calculating \eqref{eq:bopt} lies
in sampling or estimating $P(e^\test | S^{1}, \dots, S^{{|\Etrain|}}, S^\test)$, i.e. what the likely forward pass is for the test samples given the previous environments'
and limited test data. $S^{1}, \dots, S^{{|\Etrain|}}$ will give us information about $\fancyp$ and thus what things are likely across most environments,
and $S^\test$ will help us nail down things that are specific for the test distribution.

More concretely,
\begin{align*}
  P(e^\test | S^{1}, &\dots, S^{{|\Etrain|}} , S^\test) \\
  &= \int_{\mathcal{P}(\Eall)} P(e^\test | \fancyp, S^{1}, \dots, S^{{|\Etrain|}} , S^\test) P(\fancyp | S^{1}, \dots, S^{{|\Etrain|}} , S^\test) \d \fancyp \\
  &= \int_{\mathcal{P}(\Eall)} P(e^\test | \fancyp, S^\test) P(\fancyp | S^{e_1}, \dots, S^{{|\Etrain|}} , S^\test) \d \fancyp
\end{align*}
Now, the sampling term $P(e^\test | \fancyp, S^\test)$ describes how to nail down to things that are specific to $e^\test$ once we already know $\fancyp$ (i.e.
the things that are likely for all environments), and $S^\test$ which tells  us additional information to the specifics of the test environment.
By Bayes rule, this can be written as $P(e^\test | \fancyp, S^\test) = \frac{P(S^\test | e^\test) \fancyp(e^\test))}{P(S^\test | \fancyp)}$. 
This is the standard decomposition of the posterior into likelihood, prior, and marginal.
In more traditional Bayesian language, this would be written down as $P(\theta | S) = \frac{P(S | \theta) P(\theta)}{P(S)}$.

Since this last problem of approximating the posterior once we have the prior $\fancyp$ has been studied in the entire literature of Bayesian optimization
and is essentially a standard i.i.d. problem, we will focus on the term that's more relevant to us: estimating $P(\fancyp | S^{1}, \dots, S^{{|\Etrain|}}, S^\test)$.
This term describes what things are shared with high probability across all environments, given training data from multiple environments.
An example of this would be given data from multiple environments, recovering the shared causal structure between them.
This is the term that all of \autoref{chap:CIG} implicitly focused on, and the term that enables zero-shot out of distribution generalization.
One could interpret \autoref{sec:causgen} as assumptions on the meta-prior $P(\fancyp)$ such that $P(\fancyp)$
  assigns probability only to those $\fancyp$ whose environments $e \sim \fancyp$ all share a low error invariant predictor. This invariant predictor can be different
  for different $\fancyp$'s, but for each $\fancyp$ there is a shared invariant predictor across its supported environments.

Rewriting $S^\test$ as $S^{0}$, we have
\begin{align*}
  P(\fancyp | S^{0}, \dots, S^{{|\Etrain|}}) &= \frac{P(S^{0}, \dots, S^{{|\Etrain|}} | \fancyp) P(\fancyp)}{P(S^{0}, \dots, S^{{|\Etrain|}})} \\
    &= \frac{ P(\fancyp) \prod_{j=0}^{|\Etrain|} P(S^{j} | \fancyp)}{P(S^{0}, \dots, S^{{|\Etrain|}})}
\end{align*}
We will be interested in maximum a posteriori (MAP) estimates for $\fancyp$, since these ones will describe the most likely commonalities shared across environments with high
probability, and as briefly mentioned in the previous paragraph, enable OOD.
Since the term in the denominator doesn't depend on $\fancyp$, this term will be phrased as a constant $C$.
Taking logarithms, we finally obtain the posterior probability of $\fancyp$
\begin{align}
  \log P(\fancyp | S^{0}, \dots, S^{{|\Etrain|}}) &= \log C + \log P(\fancyp) + \sum_{j=0}^{|\Etrain|} \log P(S^{j} | \fancyp) \nonumber \\
  &= \log C + \log P(\fancyp) + \sum_{j=0}^{|\Etrain|} \log \int_{\Eall} \prod_{i=1}^{n_j}P(x^{(i), j}, y^{(i), j} | e_j) \fancyp(e_j) \d {e_j} \nonumber \\
  &\geq \log C + \log P(\fancyp) + \sum_{j=0}^{|\Etrain|} \int_{\Eall} \log \prod_{i=1}^{n_j}P(x^{(i), j}, y^{(i), j}| e_j) \fancyp(e_j) \d {e_j} \label{eq:vbound}\\
  &= \log C + \log P(\fancyp) + \sum_{j=0}^{|\Etrain|} \sum_{i=1}^{n_j} \EE_{e \sim \fancyp(e)} \left[\log P(x^{(i), j}, y^{(i), j}| e) \right] \label{eq:plike} 
\end{align}

Thus, we have arrived to \eqref{eq:plike} which is the addition of three terms: a constant, the meta-prior over $\fancyp$, and the likelihood of the data under $\fancyp$.
If the inequality \eqref{eq:vbound} was an equality, we would have obtained a guarantee that MAP estimation for $\fancyp$ (the left hand side) is equivalent to maximizing the likelihood of the data under $\fancyp$ and its meta-prior probability.
This is almost identical to the result on the i.i.d. case, in which MAP estimation over conditional distributions is equivalent to maximizing log-likelihood and a prior.
Since using a flat prior and optimizing likelihood (i.e. empirical risk minimization) can work so well in the i.i.d. setup, we might wonder whether we can do the same in \eqref{eq:plike}.
There are however a few differences from the i.i.d. setup, which will show why reussing the i.i.d. machinnery and just optimizing the likelihood term of \eqref{eq:plike}
  may not be such a good idea.

The first difference in this meta-i.i.d. derivation is the variational bound in equation \eqref{eq:vbound}, which is due to Jensen's inequality.
Since Jensen's inequality is an equality if and only if the random variable inside the expectation is deterministic, we can see that this is not the case, and indeed
  the inequality is strict unless we are in the pathological case where the meta-prior is concentrated on a delta (which would be assuming the test distribution is the same as the training one).
This means that if our goal is to estimate the $\fancyp$ with the highest probability under the data, optimizing the likelihood and the meta-prior can be suboptimal (which
is different than in the i.i.d. setting). However, it is a good exercise that we leave for the reader to prove that this bound becomes an equality after normalizing and letting
$n_j \to \infty$. This means that if we assume that we have enough samples from each environment but not a large number of environments (as assumed in
\autoref{chap:CIG}), MAP estimation over $\fancyp$ is approximately equivalent to optimizing the data likelihood and the meta-prior. Then, what
could be the difference with the i.i.d. setup where just optimizing likelihood can yield such good results? The answer lies in the meta-prior.

In the typical i.i.d. setting we have
\begin{equation} \label{eq:iid}
  \log P(e | S) = \log C + \log P(S | e) + \log P(e)
\end{equation}
or in more traditional notation $\log P(\theta | S) = \log C + \log P(S | \theta) + \log P(\theta)$. Successful approaches for tasks like language modelling and object
classification often require three simple things: carefully restricting the architecture (i.e. the structure of $P(S | \theta)$), using SGD and clever initialization schemes,
and optimizing the likelihood term $\log P(S | \theta)$. Restricting the architecture, SGD and initialization schemes can easily be phrased as using particular 
priors $P(\theta)$ \cite{goodfellow-ea-dl, chaudhari2018stochastic}. The combination of these priors and optimizing the likelihood leads to good in-distribution generalization
for the mentioned problems with large amounts of data.

Restricting the architecture and optimizers in the case of equation \eqref{eq:plike} would mean having $P(\fancyp)$ supported on those
$\fancyp(e)$ that assign probability to environments $e$ with conditional distributions $P(y | x, e)$ that are of these kinds of architectures.
However, it doesn't say how these different environments should relate to each other. To make this more concrete, in a problem with two variables $X, Y$ where
we want to find whether $X$ causes $Y$ or $Y$ causes $X$, we need two environments to figure this out by looking at whether $P(Y | X)$ or $P(X | Y)$ are invariant.
This means having a meta-prior $P(\fancyp)$ where each $\fancyp$ has only one of those conditionals invariant across supported environments. Put it another way,
restricting the architectures and using SGD is akin to defining a shared support of the possible $\fancyp$,
while OOD problems like cause and effect require $P(\fancyp)$ such that we can identify the stable properties over environments in $\fancyp$ from a few of its meta-samples.
In essence, we need meta-priors $P(\fancyp)$ that give us \emph{identifiability} over what properties are stable over all $\Supp(\fancyp)$,
not just sets that contain $\Supp(\fancyp)$. To stress this important point once more: architectural and optimization priors state a set $A$ such that $\Supp(\fancyp) \subseteq
A$ with high probability under the meta-prior $P(\fancyp)$. We want more. We want, given training environments, that there is an identifiable
predictor with low error across all $e \in \Supp(\fancyp)$. Since this predictor can be different for different $\fancyp$'s,
it is not enough to specify a common superset of the supports $\cup_{\fancyp \in \Supp(P(\fancyp))} \Supp(\fancyp) \subseteq A$.

These differences in possible meta-priors wouldn't matter given sufficient number of environments (at least for the meta-i.i.d. case we are assuming in this chapter), since
the role of the meta-prior dissapears as the likelihood dominates. However, we are interested in the nonasymptotic case where the number of environments is 
reasonably small.
As seen in \autoref{chap:CIG}, we are sometimes able to generalize with two environments. This is in contrast to the typical i.i.d. case where we have millions
of samples. We have a lot less meta-samples in the meta-i.i.d. setup than we have samples in traditional i.i.d. problems. Luckily in most tasks, and in particular in
those of \autoref{sec:examples}, there is a shared structure of the environments, which can be boiled down as a specific meta-prior. For instance, in \autoref{ex:example}
a meta-prior that allows us to solve this problem is one that supports $\fancyp$ for which with high probability given $e_1, \dots, e_{|\Etrain|} \sim \fancyp$
there is a single low-error invariant predictor across $\{e_1, \dots, e_{|\Etrain|}\}$ that is invariant across $\Supp(\fancyp)$.

MAML \cite{finn-ea-maml} goes one step forward of simply restricting the family of possible $P(Y | X, e)$. In addition to this, it implicitly has a meta-prior
that assumes a low error solution is obtainable by few steps of SGD on all environments by a common initialization point. This already starts to look at things which are
shared across environments and identifiability, since we can discard possible test environments when those would be far away from common initialization points of training
environments.
As such, it has a very non-flat meta-prior.
Indeed, this is likely a reasonable meta-prior for some problems. However, in \autoref{ex:example} the initialization
point would make the conditional of $P(Y | X_2)$ lie in the middle of $[\beta(e=1), \beta(e=2)]$, and thus far from that of the test environment.
It is clear that invariant prediction is not identifiable by MAML or other methods that look for lowest-error classifiers close to a common initialization.

A final work that deserves mention in this chapter is that of \cite{bengio2019meta}. There, the authors show that speed of adaptation from one
environment to another is enough to identify the causal directions given two environments. 
This work shows that meta-priors like assuming shared causal structure across environments
make optimal OOD predictors algorithmically identifiable by the meta-learning flavored idea of speed of adaptation.
In this work, as in \autoref{chap:CIG}, this meta-prior is crucial. However, the algorithmic means to exploit this prior are different, and more research needs
to be done so that we can figure out how to efficiently leverage this and other shared structures.

As a final comment, we would like to mention that we find figuring out how to merge learning invariances across training environments with finetuning when 
some limited data is available from the test domain to be a particularly promising direction for future work.

\end{appendices}
\end{document}